\documentclass[letterpaper]{article} %
\usepackage{aaai24}  %
\usepackage{times}  %
\usepackage{helvet}  %
\usepackage{courier}  %
\usepackage[hyphens]{url}  %
\usepackage{graphicx} %
\usepackage{natbib}  %
\usepackage{caption} %
\usepackage{algorithm}
\usepackage[noend]{algorithmic}

\usepackage{amsmath}
\usepackage{amsfonts}
\usepackage{cleveref}
\crefname{figure}{Figure}{Figure}
\crefname{table}{Table}{Table}
\crefname{algorithm}{Algorithm}{Algorithm}
\usepackage{multicol}
\usepackage{subcaption}

\usepackage{newfloat}
\usepackage{listings}
\DeclareCaptionStyle{ruled}{labelfont=normalfont,labelsep=colon,strut=off} %
\floatstyle{ruled}
\newfloat{listing}{tb}{lst}{}
\floatname{listing}{Listing}
\title{Symmetric Q-learning: Reducing Skewness of Bellman Error in Online Reinforcement Learning}
\author{
    Motoki Omura\textsuperscript{\rm 1},
    Takayuki Osa\textsuperscript{\rm 1, \rm 2},
    Yusuke Mukuta\textsuperscript{\rm 1, \rm 2},
    Tatsuya Harada\textsuperscript{\rm 1, \rm 2}
}
\affiliations{
    \textsuperscript{\rm 1}The University of Tokyo,\\
    \textsuperscript{\rm 2}RIKEN\\
    \{omura, osa, mukuta, harada\}@mi.t.u-tokyo.ac.jp
}

\usepackage{bibentry}

\begin{document}

\maketitle

\begin{abstract}
In deep reinforcement learning, estimating the value function to evaluate the quality of states and actions is essential. The value function is often trained using the least squares method, which implicitly assumes a Gaussian error distribution. However, a recent study suggested that the error distribution for training the value function is often skewed because of the properties of the Bellman operator, and violates the implicit assumption of normal error distribution in the least squares method. To address this, we proposed a method called Symmetric Q-learning, in which the synthetic noise generated from a zero-mean distribution is added to the target values to generate a Gaussian error distribution. We evaluated the proposed method on continuous control benchmark tasks in MuJoCo. It improved the sample efficiency of a state-of-the-art reinforcement learning method by reducing the skewness of the error distribution.
\end{abstract}

\section{Introduction}
Deep reinforcement learning (RL) has shown remarkable performance in control \cite{haarnoja2018soft} and gameplay \cite{mnih2013dqn}. In RL, it is necessary to estimate a value function to evaluate states and actions. In many cases, the value function is learned through least squares, implicitly assuming that the error distribution is a normal distribution through maximum likelihood estimation. However, during learning, the Bellman operator is used to derive the estimated value for the target. 
Because of the properties of the Bellman operator, the error distribution can become skewed, which depreciates the performance by not meeting the assumption of a normal error distribution while performing least squares. 
This issue was addressed in  \cite{garg2023extreme} by learning the value function using maximum likelihood estimation with a nonnormal error distribution. However, the error distribution can vary depending on the task or learning algorithm, making it unsuitable when the error distribution is unknown.

\begin{figure}[t]
    \centering
    \includegraphics[width=\columnwidth]{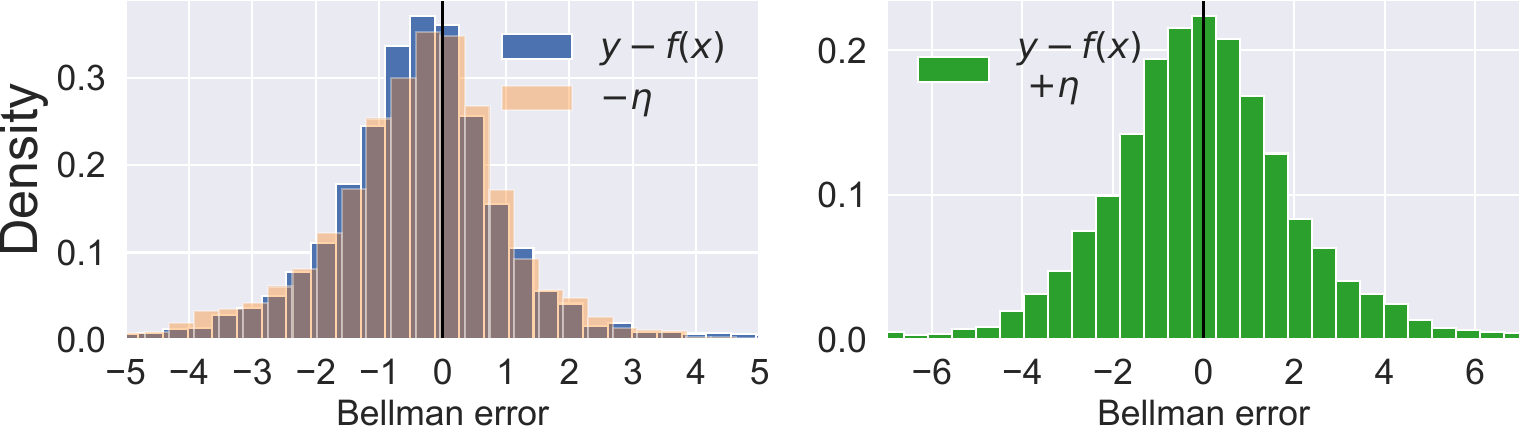}
    \caption{Bellman error, negative values of correction noise, and corrected Bellman error from Symmetric REDQ on Hopper-v2. Left: The blue histogram shows the distribution of Bellman errors. The orange histogram represents the distribution of the negative noise added to reduce skewness. It can be observed that the noise distribution fits well with the negative Bellman errors. Right: The green histogram represents the distribution of Bellman errors after adding correction noise. The skewness decreased compared to the blue distribution.}
    \label{fig:rep_plot}
\end{figure}

Therefore, in this study, we propose a method to reduce the skewness of arbitrary error distributions. We corrected the skewed error distributions to be symmetric by adding noise to the target values. As shown in the left panel of \cref{fig:rep_plot}, we learned the noise $\eta$ distribution which is the inverse of the error distribution $y-f(x)$. This resulted in a more symmetric error distribution with increased symmetry when noise was added to the target, as shown in the right panel of the plot. By applying least squares to this symmetric error distribution, we can better satisfy the assumption of the error distribution and improve performance.
The effectiveness of the proposed method was verified in combination with the Soft Actor-Critic (SAC)  \cite{haarnoja2018soft} and Randomized Ensemble Double Q-learning (REDQ) \cite{chen2021randomized}, which we refer to as Symmetric SAC (SymSAC) and Symmetric REDQ (SymREDQ), respectively. We evaluated our proposed method on five challenging tasks in MuJoCo \cite{mujoco2012,brockman2016openai} and achieved comparable or better sample efficiency than that of state-of-the-art online RL methods. We also confirm that our method corrects skewed error distributions to become symmetric by visualizing the error distributions during the learning process.
The contributions of this study are as follows.

\begin{itemize}
    \item  We propose a method of adding noise to target variables and reducing the skewness of error distributions, which is a problem when using least squares.
    \item We evaluated the proposed method in continuous control benchmark tasks and showed that the proposed method can reduce the skewness of error distribution.
    \item We empirically demonstrated that the proposed method can improve the sample efficiency of REDQ, a state-of-the-art model-free RL algorithm.
\end{itemize}

\section{Background}
\subsection{Reinforcement Learning}
We consider the RL problem under a Markov decision process (MDP) defined by the tuple $(\mathcal{S},\mathcal{A},\mathcal{P},r,\gamma,d)$. $\mathcal{S}$ represents the state space, $s$ represents a state, $\mathcal{A}$ represents the action space, $a$ represents an action, $\mathcal{P}(s_{t+1} \mid s_t,a_t)$ is the transition probability density, $r(s,a)$ is the reward function, $\gamma$ is the discount factor, and $d(s_0)$ is the probability density of the initial state. The policy $\pi(a \mid s)$ is defined as the conditional probability density of a given action. The goal of RL is to identify the policy that maximizes the expected cumulative discounted reward $\mathbb{E}[R_0 \mid \pi]$, where $R_t$ is the return, given by $R_t = \sum^{T}_{k=t} \gamma^{k-t} r(s_k, a_k)$.

\subsection{Bellman Error Distribution}

In RL, the Q-function is estimated based on the fact that it satisfies the Bellman equation:

\begin{equation}
    Q^\pi(s,a) = \mathbb{E}_{s' \sim \mathcal{P}, a'\sim \pi} [ r(s, a) + \gamma Q^{\pi}(s', a') ]
\end{equation}
Based on the Bellman equation, the Bellman operator is often used to estimate the Q-function and the Bellman operator is known to be contractive. Therefore, minimizing the mean squared error is a common approach for learning the Q-function.

\begin{equation}
    \mathcal{L}(\theta) = \mathbb{E}_{s, a \sim \rho}[(y-Q^\pi(s, a; \theta))^2]
\end{equation}
$\rho(s, a)$ is a probability distribution over states $s$ and actions $a$, $\theta$ denotes the parameters of the Q-function approximator, and the target value $y$ is computed as:

\begin{equation}
  y = \mathbb{E}_{s' \sim \mathcal{P}, a'\sim \pi}[r(s, a) + \gamma Q^\pi(s', a'; \theta) \mid s, a] 
\end{equation}
The prediction error between $y$ and the Q-function is known as the Bellman error.

In the least squares method, it is implicitly assumed that the error distribution follows a normal distribution through maximum likelihood estimation. However, in \cite{garg2023extreme}, it is explained based on extreme value theory \cite{fisher_tippett_1928_evt,mood1963introduction_evt} that the distribution of the Bellman error can become skewed when using policies that aim to maximize the Q-value owing to the influence of the maximum operator. They then used a Gumbel regression, which is a maximum likelihood estimation that assumes a Gumbel distribution as the error distribution. 

\begin{figure}[t]
    \centering
    \includegraphics[width=\columnwidth]{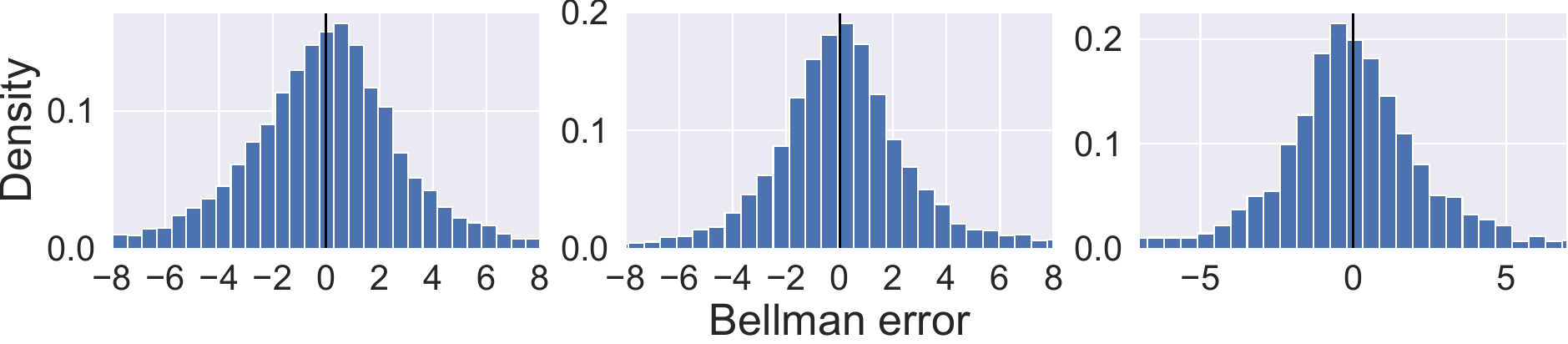}
    \caption{The pre-corrected Bellman error at three different steps when learning Walker2d with SymREDQ.}
    \label{fig:plot3}
\end{figure}

However, that method has several issues. While extreme value theory assumes that the maximum is taken from independently sampled values, in the context of RL, the Q-values in the maximum operator are dependent on state-action pairs. Therefore, it is difficult to satisfy this assumption and obtain an exact extreme value distribution such as the Gumbel distribution. In fact, the Bellman error distribution deviates from the Gumbel distribution during learning, depending on the task \cite{garg2023extreme}. Furthermore, the error distribution is influenced by many factors such as double Q-learning  \cite{hasselt2010doubleq,vanhasselt2015ddqn,fujimoto2018addressing} and the target network \cite{mnih2015dqn}, and may change during the learning process as shown in \cref{fig:plot3}, making it difficult to make accurate assumptions regarding the error distribution. Therefore, this study proposes a flexible approach to assumptions regarding error distribution.

\section{Symmetric Q-learning}
In this section, we introduce symmetric Q-learning, a technique that corrects the error distribution, making it symmetric and close to a normal distribution by adding noise to the target variable.

\subsection{Assumption of Error Distribution}
Our motivation is to approach the error distribution as a normal distribution to satisfy the assumption of the least squares method. Therefore, we first demonstrate that the error distribution is assumed to follow a normal distribution.

Here, we explain that the least squares method is equivalent to maximum likelihood estimation assuming Gaussian errors.
Assuming that we are given a training dataset consisting of pairs of observed values $x$ and the corresponding target values $y$, we consider the problem of predicting the value of $y$ for a new value of $x$. We assume that the target variable $y$ is the sum of a function $f(x;\theta)$ with parameters $\theta$ and Gaussian noise.

\begin{equation}
  y = f(x; \theta) + \epsilon, \; \epsilon \sim \mathcal{N}(0,\,\sigma^{2}) 
\end{equation}
In this case, the probability of $y$ given $x, \theta$, and $\sigma$ follows a normal distribution with mean $f(x; \theta)$.

\begin{equation}
  p(y \mid x, \theta, \sigma) = \mathcal{N}(y \mid f(x; \theta), \sigma^{2}) 
\end{equation}
Therefore, the expected log-likelihood is as follows:

\begin{equation}
\begin{split}
  &\mathbb{E}_{y, x \sim p_{data}}[\log p(y \mid x, \theta, \sigma)] \\
  &= \mathbb{E}_{y, x \sim p_{data}}\left[-\frac{1}{2}  \log{2\pi\sigma^2}– \frac{1}{2\sigma^2}(y - f(x; \theta))^2\right]
\end{split}
\end{equation}
If we consider maximizing the expected log-likelihood, the loss function with respect to $\theta$ can be expressed as follows:

\begin{equation}
    \mathcal{L}(\theta) = \mathbb{E}_{y, x \sim p_{data}}[(y-f(x; \theta))^2]
\end{equation}
This is equivalent to the loss function used in the least squares method.

Thus, in the least squares method, it is implicitly assumed through a maximum likelihood estimation that the error distribution follows a normal distribution.
However, as explained in the previous section, the Bellman error can exhibit a distribution that deviates from the Gaussian distribution.
As seen in the left plot of \cref{fig:rep_plot}, even in REDQ, which has state-of-the-art sample efficiency and uses the least squares method, the pre-corrected Bellman error is also skewed away from a normal distribution.
Thus, in online RL, the error distribution varies depending on the task or algorithm. Therefore, we propose an algorithm that can be applied to any error distribution that changes during the learning process.

\subsection{Normalization of Error Distribution}
As seen in the previous section, it is possible for the Bellman error $\epsilon$ to not follow either a normal or a Gumbel distribution in practice. Therefore, we consider controlling the error distribution by adding zero-mean noise $\eta$ to the target variable to ensure that the error distribution follows the normal distribution. We assume that the noise $\eta$ follows the probabilistic model $q(\eta; \omega)$ with parameter $\omega$. The method used to estimate $\omega$ is described in the following section.

\begin{equation}
    y + \eta = f(x; \theta) + \epsilon + \eta, \; \eta \sim q(\eta; \omega)
\end{equation}
If we assume that the sum of the original error $\epsilon$ and the added noise $\eta$ follows a normal distribution, then the probability of $y+\eta$ follows the normal distribution with mean $f(x;\theta)$.

\begin{equation}
    \epsilon + \eta \sim \mathcal{N}(0,\,\sigma^{2}) 
\end{equation}

\begin{equation}
  p(y+\eta) = \mathcal{N}(f(x; \theta),\,\sigma^{2}) 
\end{equation}
Following the same approach as in the previous section, the expected log likelihood and loss function are given as

\begin{equation}
\begin{split}
  &\mathbb{E}_{y, x \sim p_{data}, \eta \sim q}[\log p(y+\eta \mid x, \theta, \sigma, \omega)] \\ 
  &= \mathbb{E}_{y, x \sim p_{data}, \eta \sim q}\left[-\frac{1}{2}  \log{2\pi\sigma^2}– \frac{1}{2\sigma^2}(y - f(x; \theta) + \eta)^2\right]
\end{split}
\end{equation}

\begin{equation}
    \mathcal{L} (\theta) = \mathbb{E} _{y, x \sim p_{data}, \eta  \sim q }[(y -f(x; \theta) +\eta)^2]
\end{equation}
When the error $y - f(x; \theta) + \eta$ follows the normal distribution, minimizing the mean squared error satisfies the assumption of an error distribution, allowing for more accurate estimates. Therefore, it is necessary to learn the distribution of $\eta$ which causes $y - f(x; \theta) + \eta$ to approach the normal distribution.

Before explaining the learning of the distribution of $\eta$, it should be noted that adding zero-mean noise $\eta$ does not introduce any bias into the learning of $\theta$. This is easily demonstrated through the following manipulations:

\begin{equation}
\begin{split}
    \mathcal{L} (\theta) 
    = & \mathbb{E} _{y, x \sim p_{data}, \eta  \sim q }[(y-f(x; \theta)+\eta)^2]
    \\  = &\mathbb{E} _{y, x \sim p_{data}, \eta  \sim q }[y^2+f(x; \theta)^2+\eta^2 - 2y f(x;\theta) \\&  - 2\eta f(x;\theta) +2y\eta]
    \\ = & \mathbb{E} _{y, x \sim p_{data}, \eta  \sim q }[(y-f(x; \theta))^2 + \eta^2] 
\end{split}
\end{equation}
Because $\eta^2$ does not affect the learning of $\theta$, adding $\eta$ to the target value does not introduce any bias into the estimator or change the expectation of the mean squared error.

However, $\eta$ affects the error distribution shape. Let us consider skewness as an indicator of symmetry. By substituting the pre-corrected error $y-f(x)$ with $\epsilon$, the skewness of the post-corrected error distribution can be expressed as follows:

\begin{equation}
\begin{split}
    Skewness(\epsilon + \eta) 
       & = \frac{\mu_{3}(\epsilon+\eta)}{\sigma^{3}(\epsilon+\eta)}
    \\ &= \frac{\mu_{3}(\epsilon) + \mu_{3}(\eta)}{(\sigma^{2}(\epsilon)+\sigma^{2}(\eta))^{3/2}}
\end{split}
\end{equation}

$\mu_3$ and $\sigma$ represent the third central moment and the variance, respectively.
This equation indicates that the skewness, and hence the symmetry, of the error distribution after correction is influenced by the third moment of $\eta$. Moreover, correcting the distribution can reduce the skewness and increase the symmetry when the third moment of $\eta$ has the opposite sign to that of $\epsilon$. 

It is expected that learning until convergence while adding zero-mean noise will result in no change in the final return because adding noise does not introduce any bias. 
However, online RL is a non-stationary learning process, and it is difficult to always achieve sufficient learning during the learning process. Furthermore, bias accumulates through bootstrapping of target values. Therefore, improving the learning efficiency through distribution correction during the learning process may lead to improved performance, including the final return.
The following section describes the learning of the distribution of $\eta$.

\subsection{Training of Noise Distribution}

Any arbitrary distribution can be used for the probability model $q(\eta ; \omega)$, and the parameter $\omega$ is estimated in a way that $ \epsilon + \eta $ approaches the normal distribution. However, it is difficult to estimate $\omega$ accurately so that it becomes a normal distribution. Therefore, we considered making the distribution symmetric. 

When $\eta$ follows a distribution of $- \epsilon$, that means $\eta \sim  p_{\epsilon}( - \eta)$, $\mu_{3}(\eta)=-\mu_{3}(\epsilon)$ and the skewness is zero. Therefore, the skewness of $\epsilon+\eta$ decreases and approaches a symmetric distribution by making the distribution of $\eta$ closer to a distribution of $- \epsilon$. 
In other words, we bring the probability model $q(\eta; \omega)$ closer to $p_{\epsilon}(-\eta)$. To do this, we estimate the parameters $\omega$ by maximizing the likelihood of $y-f(x; \theta)$ with respect to $p_{\epsilon}$. The specific steps are shown below.

Assuming that $\eta$ follows a distribution of $- \epsilon$, then conversely, $\epsilon$ follows a distribution of $- \eta$:
\begin{equation}
    \epsilon \sim p_{\eta}(-\epsilon)
\end{equation}
Then, the probability of $\epsilon$ can be expressed as follows:

\begin{equation}
    p_{\epsilon}(\epsilon) = p_{\eta}(-\epsilon) \simeq q(-\epsilon; \omega)
\end{equation}

From $\epsilon = y - f(x; \theta)$, the likelihood of the observed sample $y-f(x; \theta)$ becomes $q(-(y-f(x; \theta); \omega)$, and the loss function for maximizing this likelihood can be expressed as follows:

\begin{equation}
\label{eq:loss_eta}
    L(\omega) = \mathbb{E} _{y, x \sim p_{data}}[ - \log q(-(y - f(x;\theta)) ; \omega)]
\end{equation}
By minimizing this loss function, $q(\eta; \omega)$ approaches $p_{\epsilon}(-\eta)$ and $\eta$ becomes noise, which reduces the skewness of $\epsilon$.

As a result of the aforementioned learning, it was demonstrated that the skewness of $\epsilon + \eta$ can approach zero. However, a zero skewness does not necessarily imply that the distribution is symmetric. Nevertheless, it is shown from the following formula that the error distribution becomes symmetric:

\begin{equation}
\begin{split}
    p_{\epsilon + \eta}(x) & = p_{\epsilon}(x) * p_{\eta}(x)  \\
    & = \int_{-\infty}^\infty p_{\epsilon}(u)p_{\eta}(x-u) \,du  \\
    & = \int_{-\infty}^\infty p_{\epsilon}(u)p_{\epsilon}(u-x) \,du  \\
    & = \int_{-\infty}^\infty p_{\epsilon}(v+x)p_{\epsilon}(v) \,dv, (v = u - x)  \\
    & = \int_{-\infty}^\infty p_{\eta}(-x-v)p_{\epsilon}(v) \,dv  \\
    & = p_{\epsilon + \eta}(-x) \\
\end{split}
\end{equation}

$p_{X}(x)$ is the probability density function for the random variable $X$, and $p_{\eta}(x) =  p_{\epsilon}(-x)$ when $\eta$ follows a distribution of $- \epsilon$. From the above formula, it can be seen that the error distribution $p_{\epsilon + \eta}(x)$ is symmetric with respect to $x=0$.

\subsection{Practical Algorithms}

For the distribution $q(\eta; \omega)$, it is preferable to have a distribution that can represent a skewed distribution with sufficiently wide support. Therefore, as a practical algorithm, we approximate $q(\eta; \omega)$ with the Gaussian mixture models (GMM) by employing variational Bayes expectation maximization \cite{Bishop:2006}, which results in a more stable parameter estimation than that of the expectation maximization algorithm based on maximum likelihood estimation. GMM can approximate skewed distributions such as the Gumbel distribution with high accuracy and supports the entire real line. During the training process, we iteratively update the GMM parameters to approach the distribution of negative Bellman errors using the same samples used for Q-function learning. To prevent overfitting the samples, we use a hyperparameter $k$ to update the parameters for every $k$ steps.
Because the Bellman error is a one-dimensional data, estimating the GMM has a relatively low computational cost.

\begin{algorithm}[H]
   \caption{Symmetric REDQ}
   \label{alg:symredq}
    \begin{algorithmic}
       \STATE Initialize policy parameters $\theta$, $N$ Q-function parameters $\phi_i$, $i=1, \ldots , N$, GMM parameters $\omega$, empty replay buffer $\mathcal{D}$. Set target parameters $\phi_{targ,i} \leftarrow \phi_i$, for $i = 1, 2, \ldots , N$
       \REPEAT
       \STATE Take one action $a_t \sim \pi_\theta(\cdot \mid s_t)$. Observe reward $r_t$, new state $s_{t+1}$.
       \STATE Add data to buffer: $\mathcal{D} \leftarrow \mathcal{D} \cup \{(s_t, a_t, r_t, s_{t+1})\}$
       \FOR{$G$ updates}
       \STATE Sample a mini-batch $B = \{(s, a, r, s')\}$ from $\mathcal{D}$
       \STATE Sample a set $\mathcal{M}$ of $M$ distinct indices from $\{1, 2, \ldots, N\}$
       \STATE Compute the Q target $y$ (same for all of the $N$ Q-functions): 
        \begin{align*}
            y =\; & r + \gamma \left( \min_{i \in \mathcal{M}} Q_{\phi_{\text{targ}, i}}(s', \tilde{a}') -  \alpha \log \pi_\theta (\tilde{a}' \mid s') \right) \\
            & ,\tilde{a}' \sim \pi_\theta(\cdot \mid s')
        \end{align*}
       \FOR{$i = 1, \ldots , N$}
       \STATE Compute error $\epsilon_i = y - Q_{\phi_{i}}(s, a)$
       \ENDFOR
       \STATE Every k steps update $\omega$ using $- \epsilon$ by minimizing (\ref{eq:loss_eta}) 
       \FOR{$i = 1, \ldots , N$}
       \STATE Sample $\eta$ from GMM $q(\eta ; \omega)$
       \STATE Update $\phi_i$ with gradient descent using
       $$ \displaystyle \nabla_\phi \frac{1}{|B|} \sum_{(s,a,r,s') \in B} (y- Q_{\phi_i}(s,a)+\eta)^2 $$
       \STATE Update target networks with $$\phi_{targ,i} \leftarrow \rho \phi_{targ,i} + (1 - \rho)\phi_i $$
       \ENDFOR
       \ENDFOR
       \STATE Update policy parameters $\theta$ with gradient ascent using
       \begin{align*}
        \nabla_\phi \frac{1}{|B|} &  \sum_{s \in B} \biggl( \frac{1}{N}\sum^{N}_{i=1} Q_{\phi_i}(s,\tilde{a}_{\theta}(s)) - \\ 
        & \alpha \log \pi_\theta (\tilde{a}_\theta(s) \mid s) \biggr) , \tilde{a}_\theta(s) \sim \pi_\theta (\cdot \mid s) 
       \end{align*}
       \UNTIL{end}
    \end{algorithmic}
\end{algorithm}

To ensure that $\eta$ becomes unbiased noise, the distribution is shifted such that the mean becomes zero each time the parameters of GMM are updated. One problem with adding noise to the error is that it can increase the error variance, which can be reduced by ensembling. Therefore, when combining this method with algorithms such as REDQ, it is recommended to use a larger ensemble to adjust the variance of the Q-function

The pseudocode for the SymREDQ algorithm, which combines the proposed method with REDQ, is presented in \cref{alg:symredq}. 
When the ensemble size $N=2$, in-target minimization parameter $M=2$, and update-to-data (UTD) ratio $G=1$, it becomes SymSAC. When $N=1$, $M=1$, and $G=1$, it becomes SymSAC without an ensemble.
The pseudocode for GMM training is shown in the appendix.

\begin{figure*}[t]
  \centering
  \setlength{\tabcolsep}{0pt}
  \begin{tabular}{ccccc}
    \includegraphics[width=0.41\columnwidth]{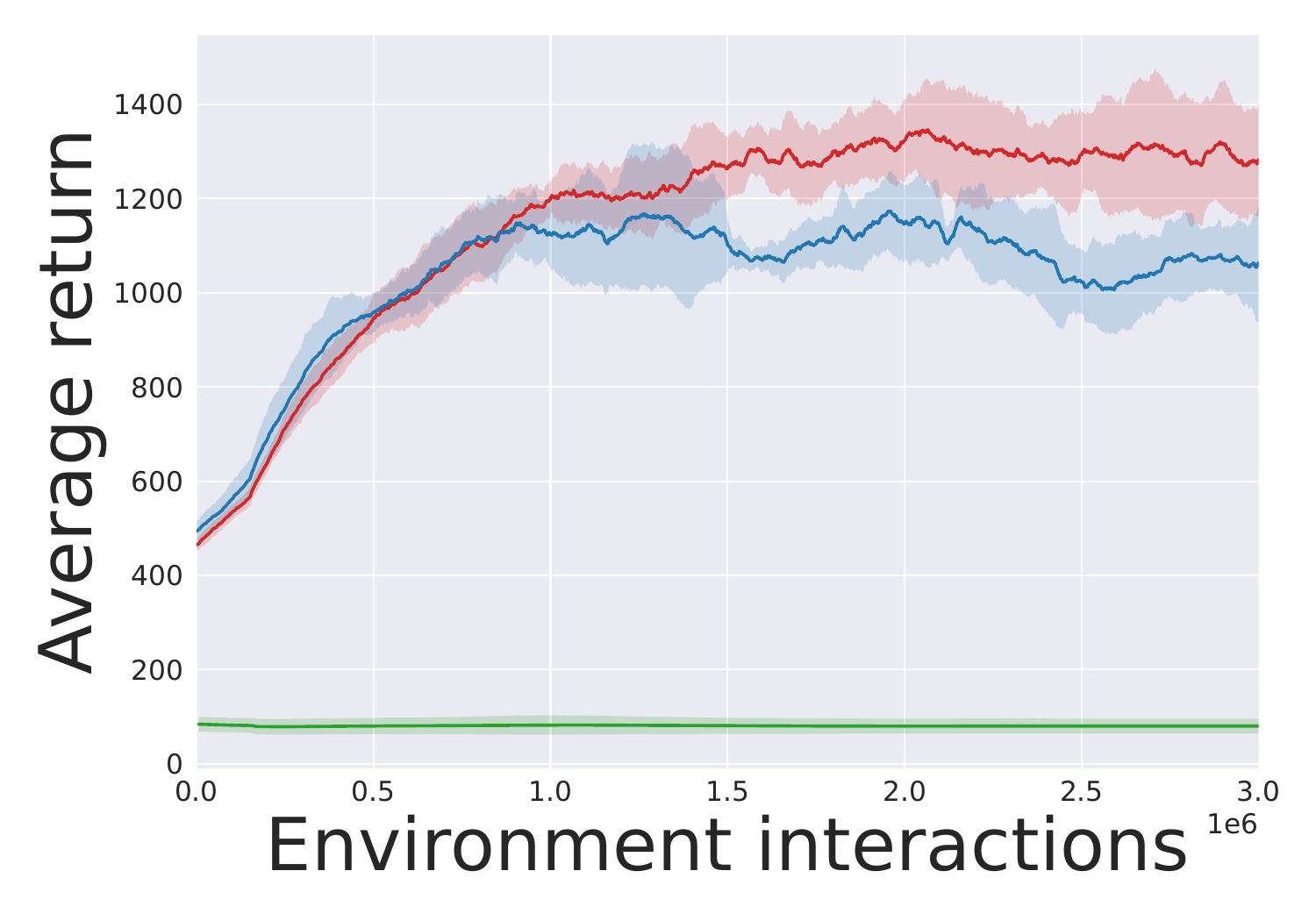} &
    \includegraphics[width=0.41\columnwidth]{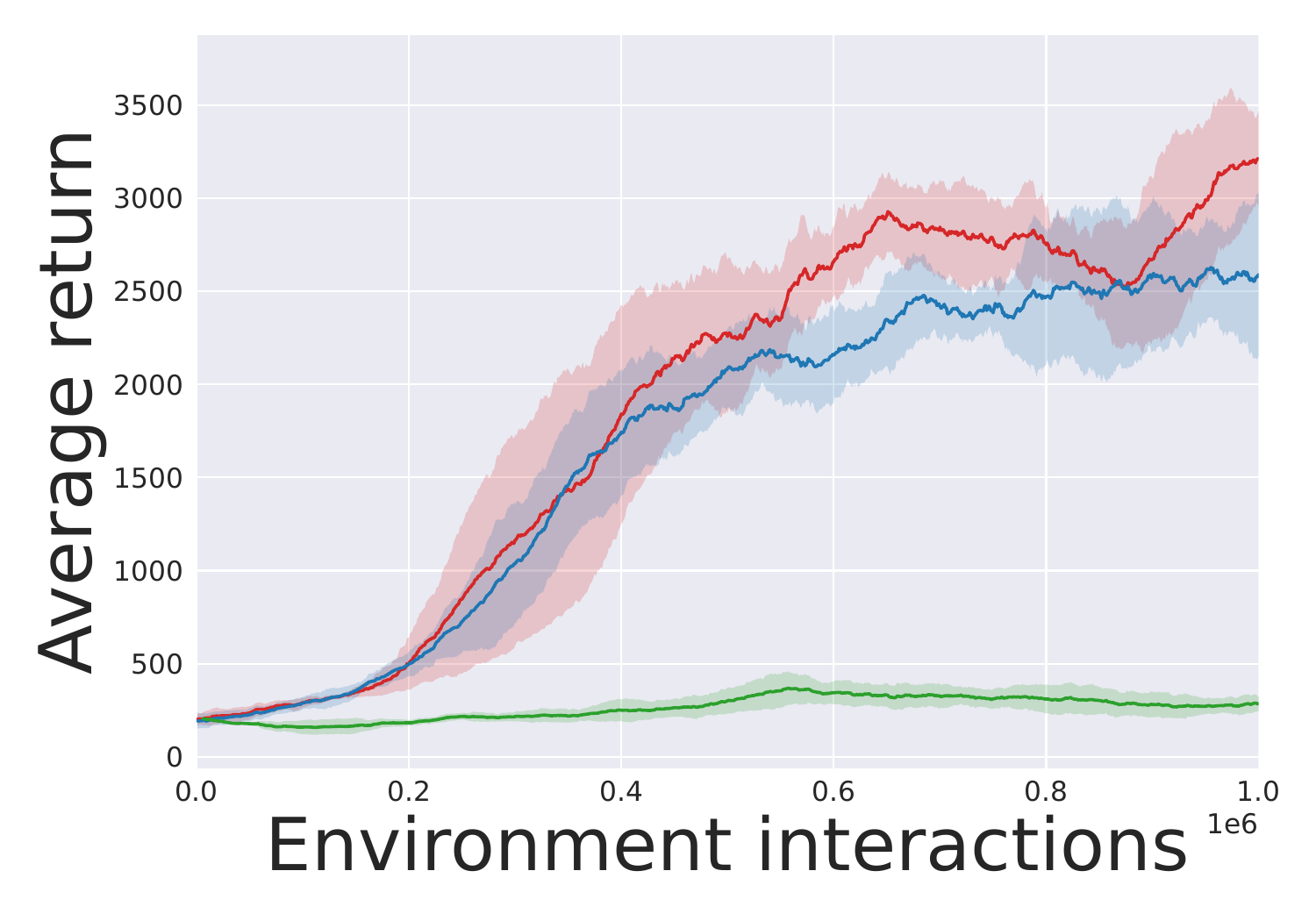} &
    \includegraphics[width=0.41\columnwidth]{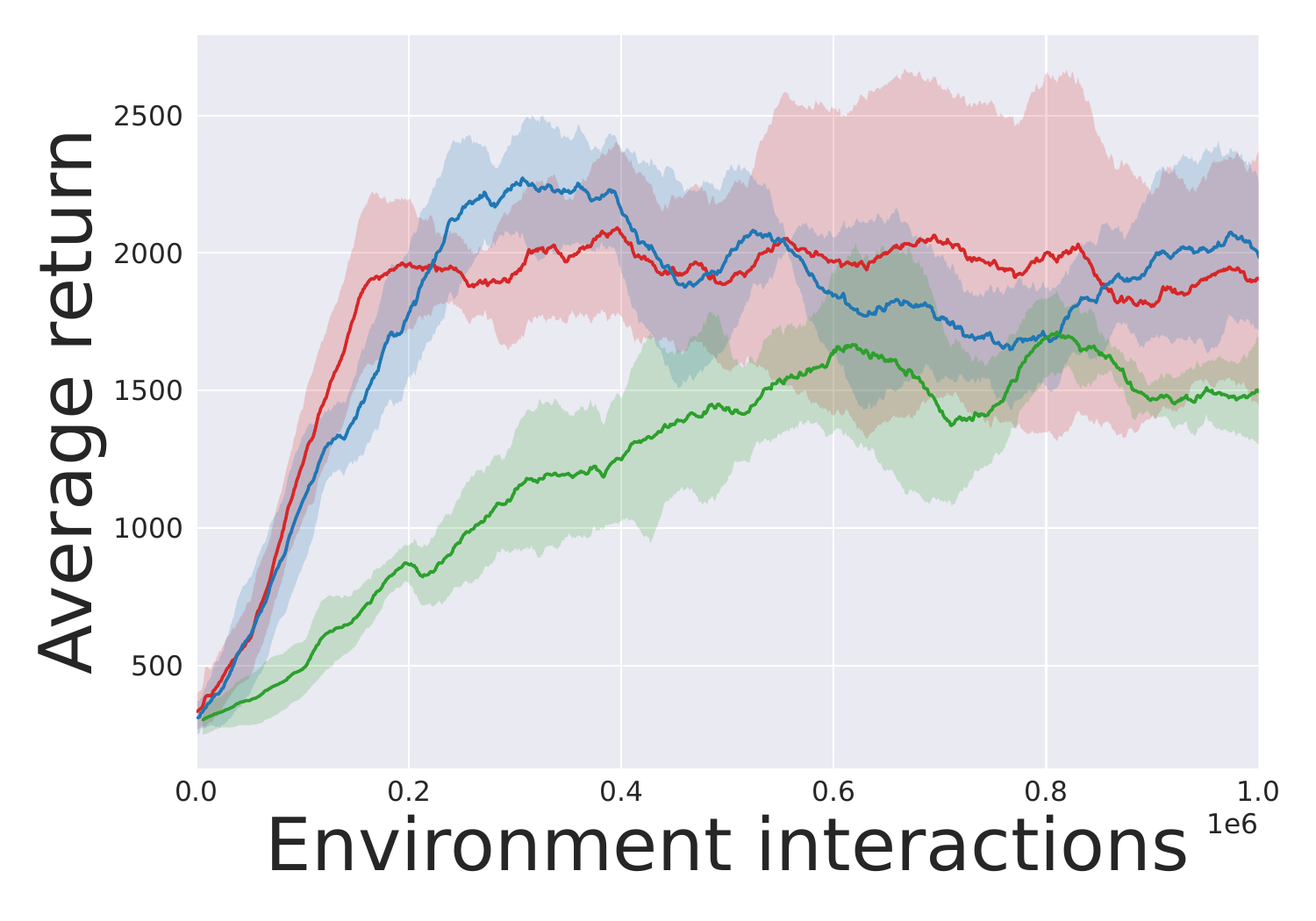} &
    \includegraphics[width=0.41\columnwidth]{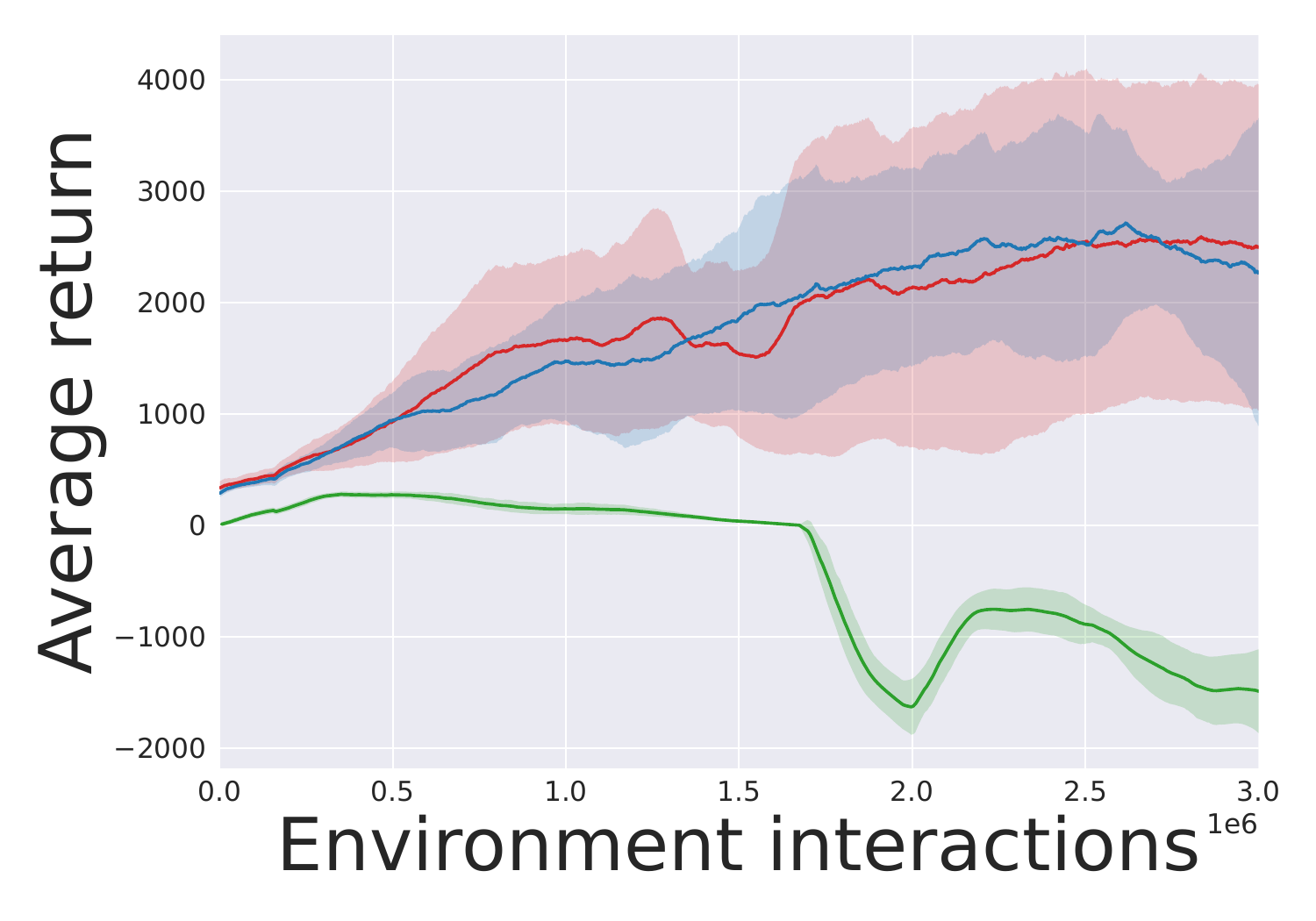} &
    \includegraphics[width=0.41\columnwidth]{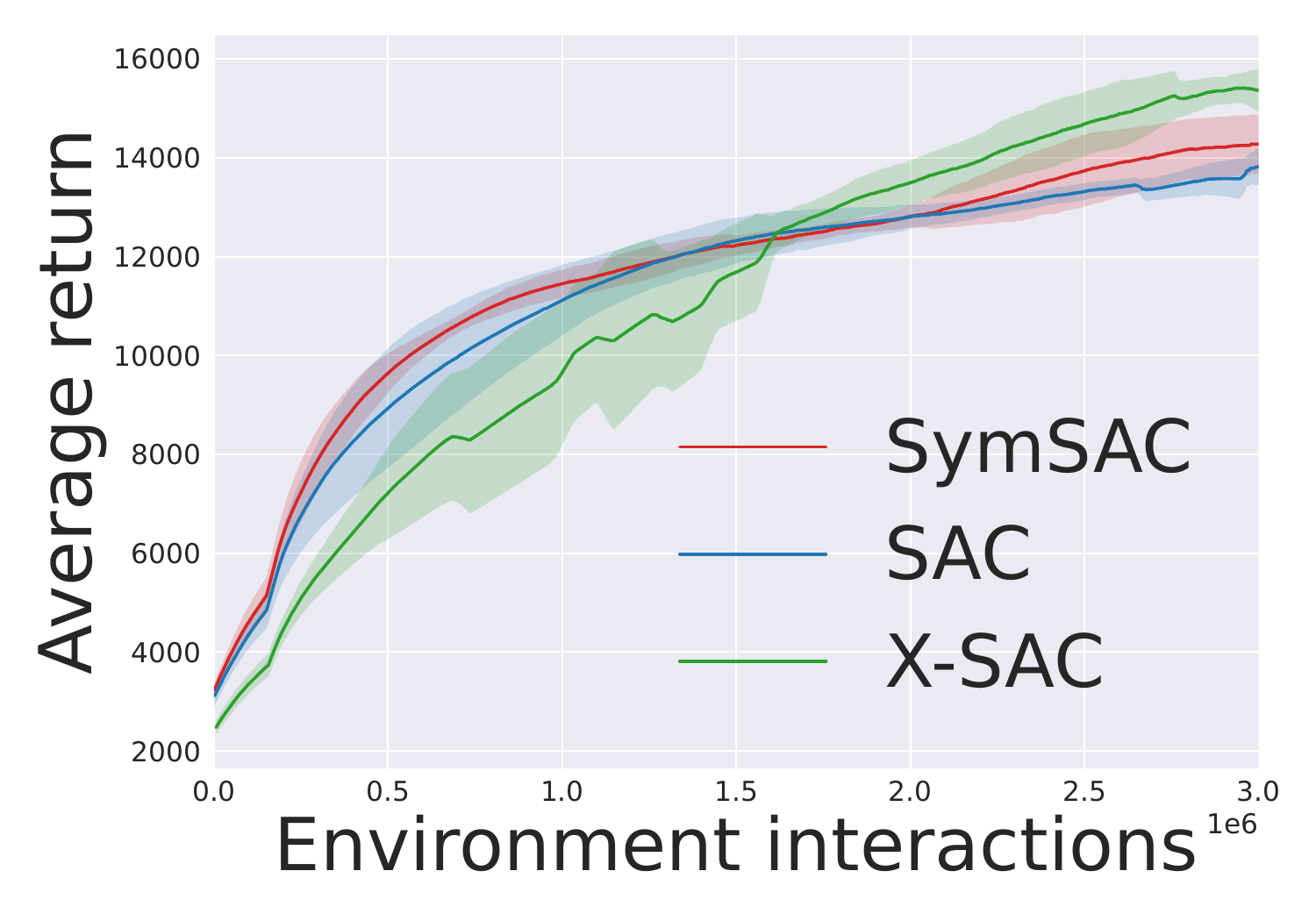} \\
    (a) Humanoid-v2 & (b) Walker2d-v2 & (c) Hopper-v2 & (d) Ant-v2 & (e) HalfCheetah-v2
  \end{tabular}
  \caption{Comparison of SymSAC, SAC and $\mathcal{X}$-SAC without ensembles for UTD=1}
  \label{fig:utd1}
\end{figure*}

\begin{figure*}[h]
  \centering
  \setlength{\tabcolsep}{0pt}
  \begin{tabular}{ccccc}
    \includegraphics[width=0.41\columnwidth]{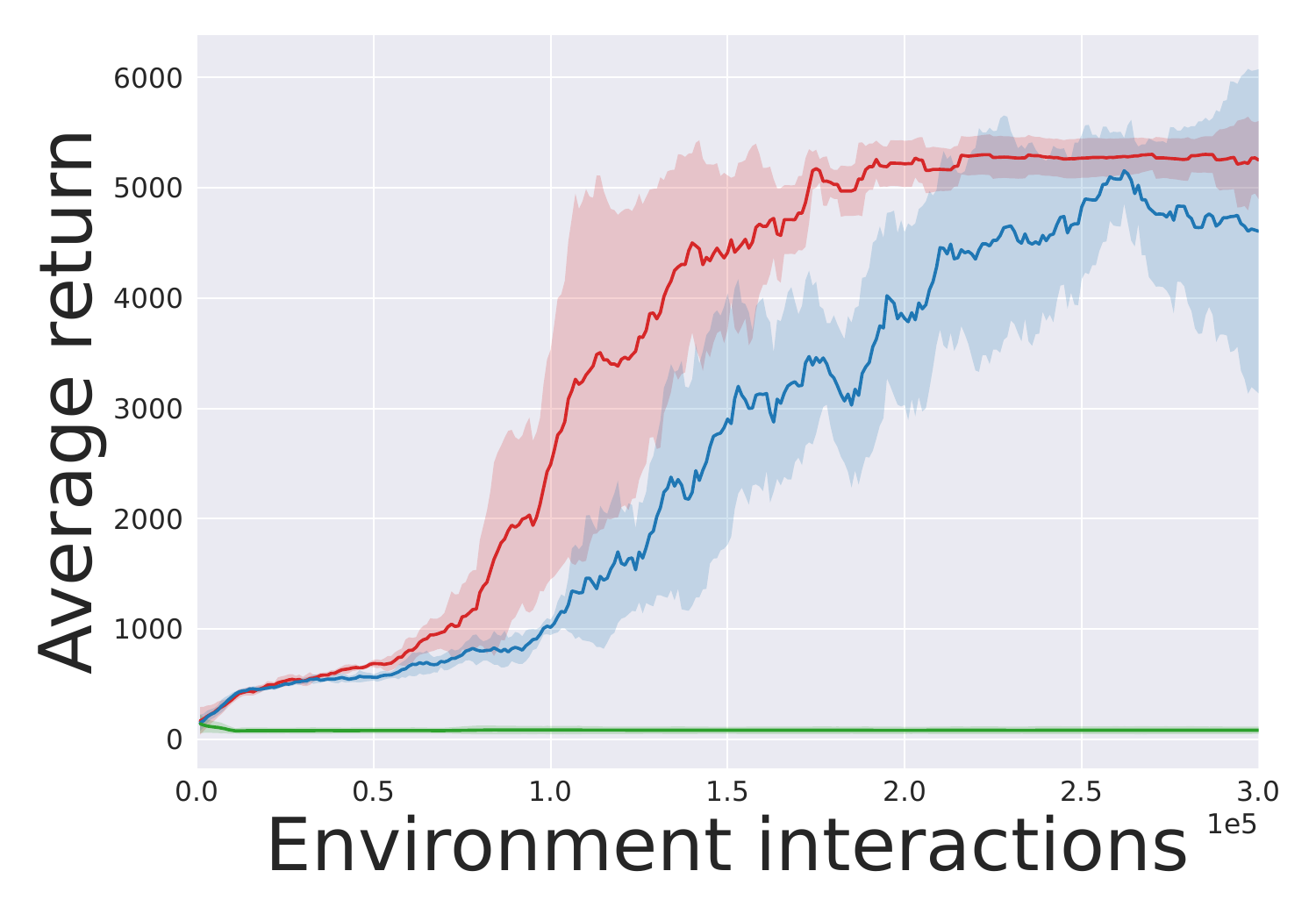} &
    \includegraphics[width=0.41\columnwidth]{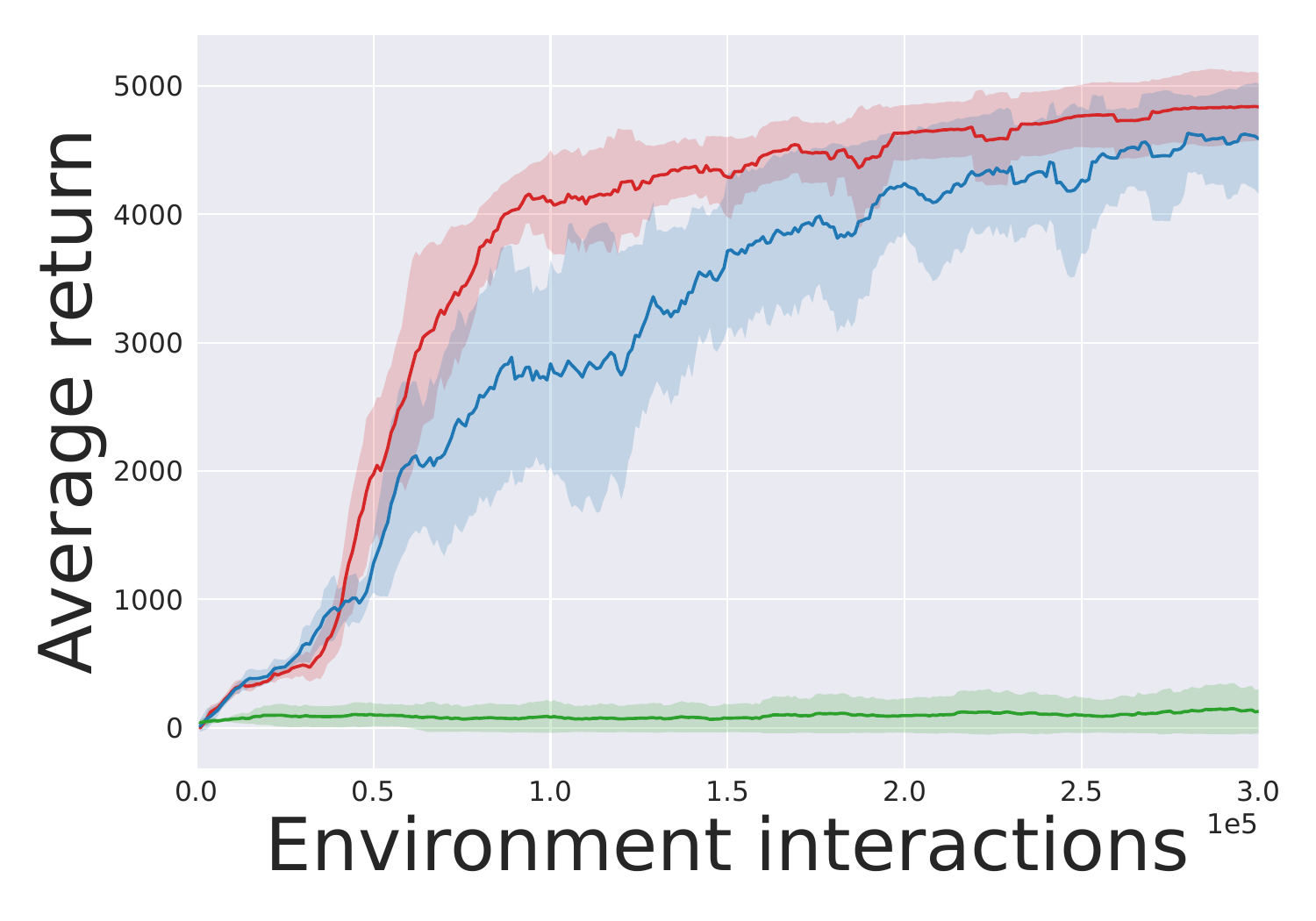} &
    \includegraphics[width=0.41\columnwidth]{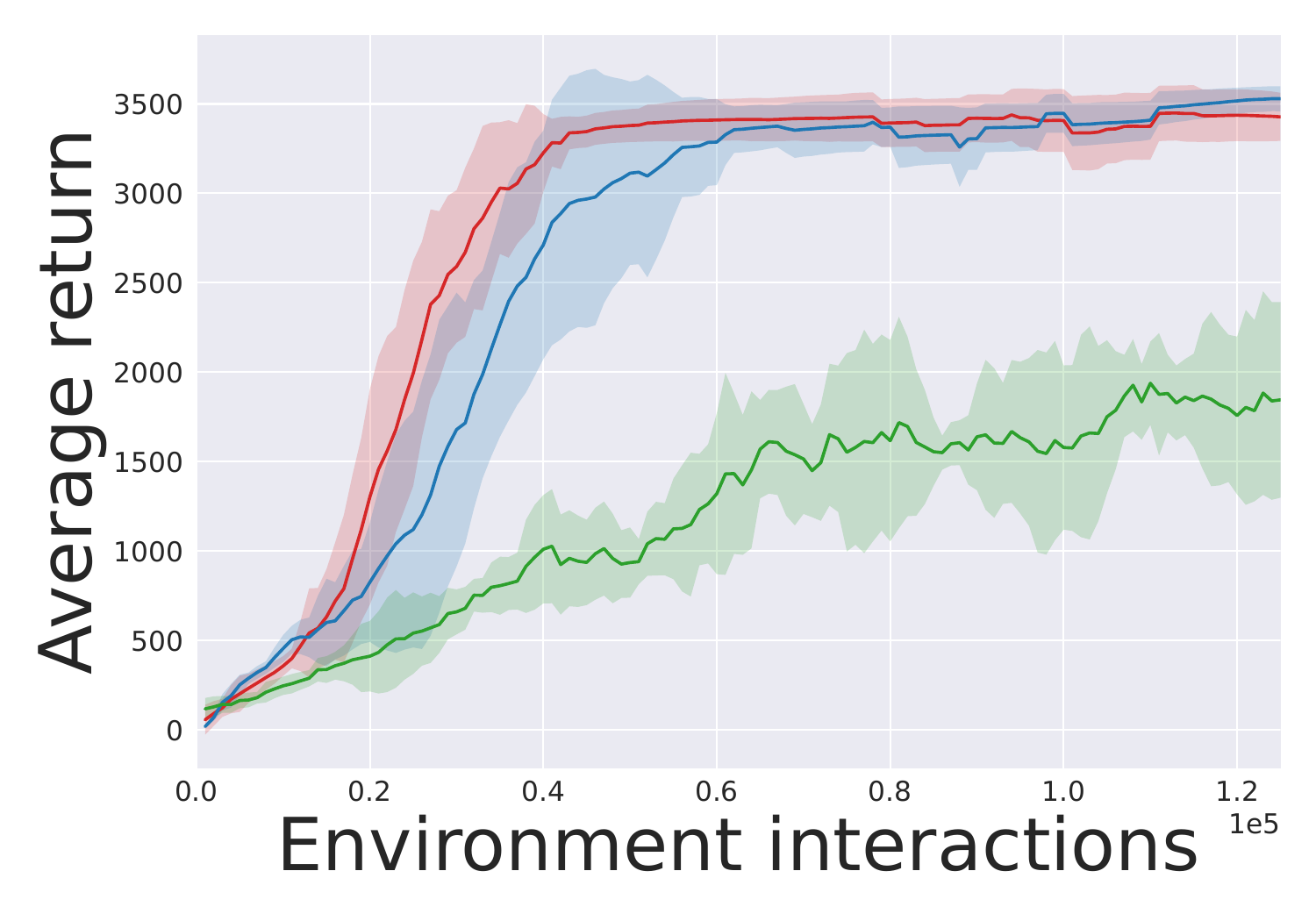} &
    \includegraphics[width=0.41\columnwidth]{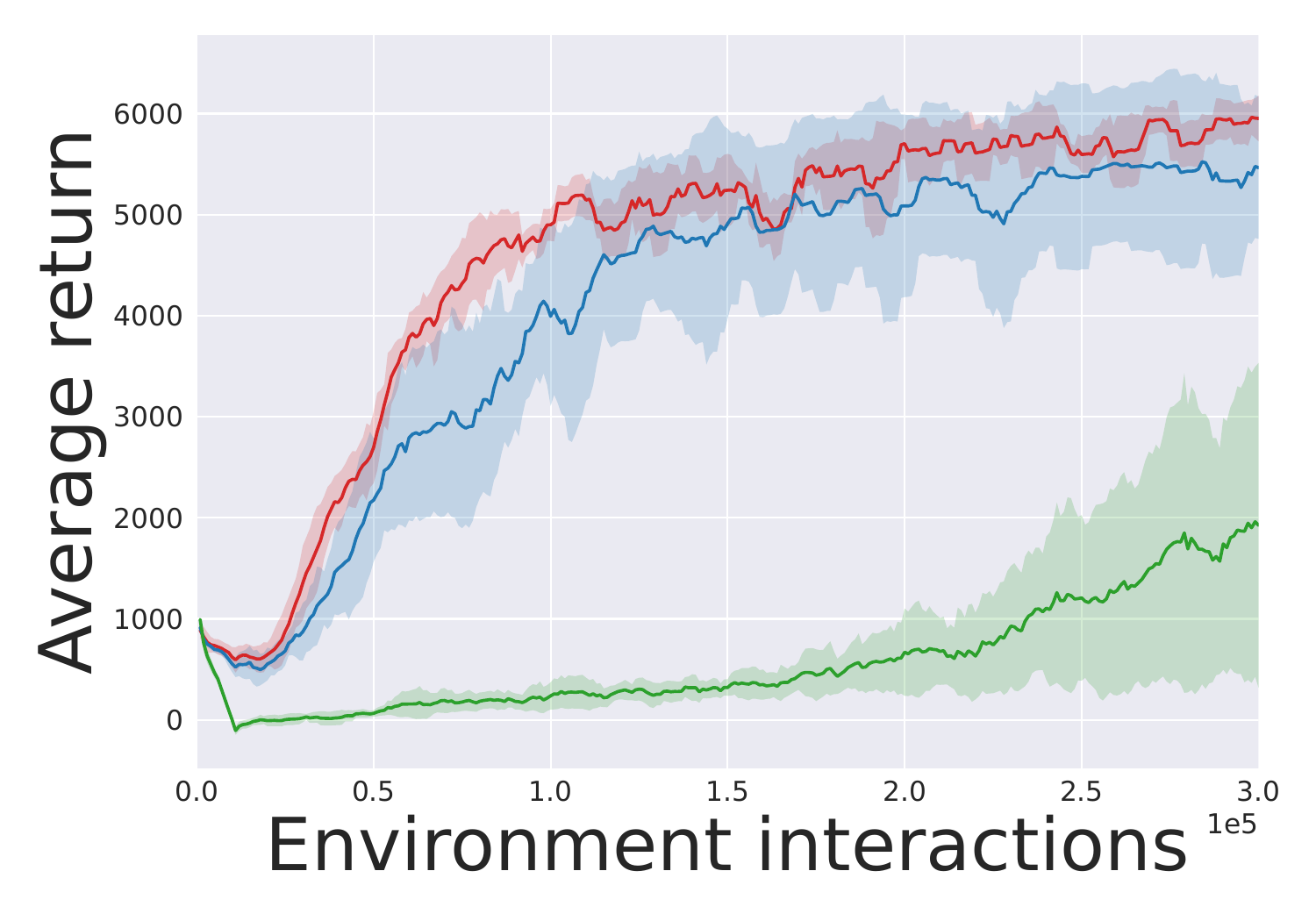} &
    \includegraphics[width=0.41\columnwidth]{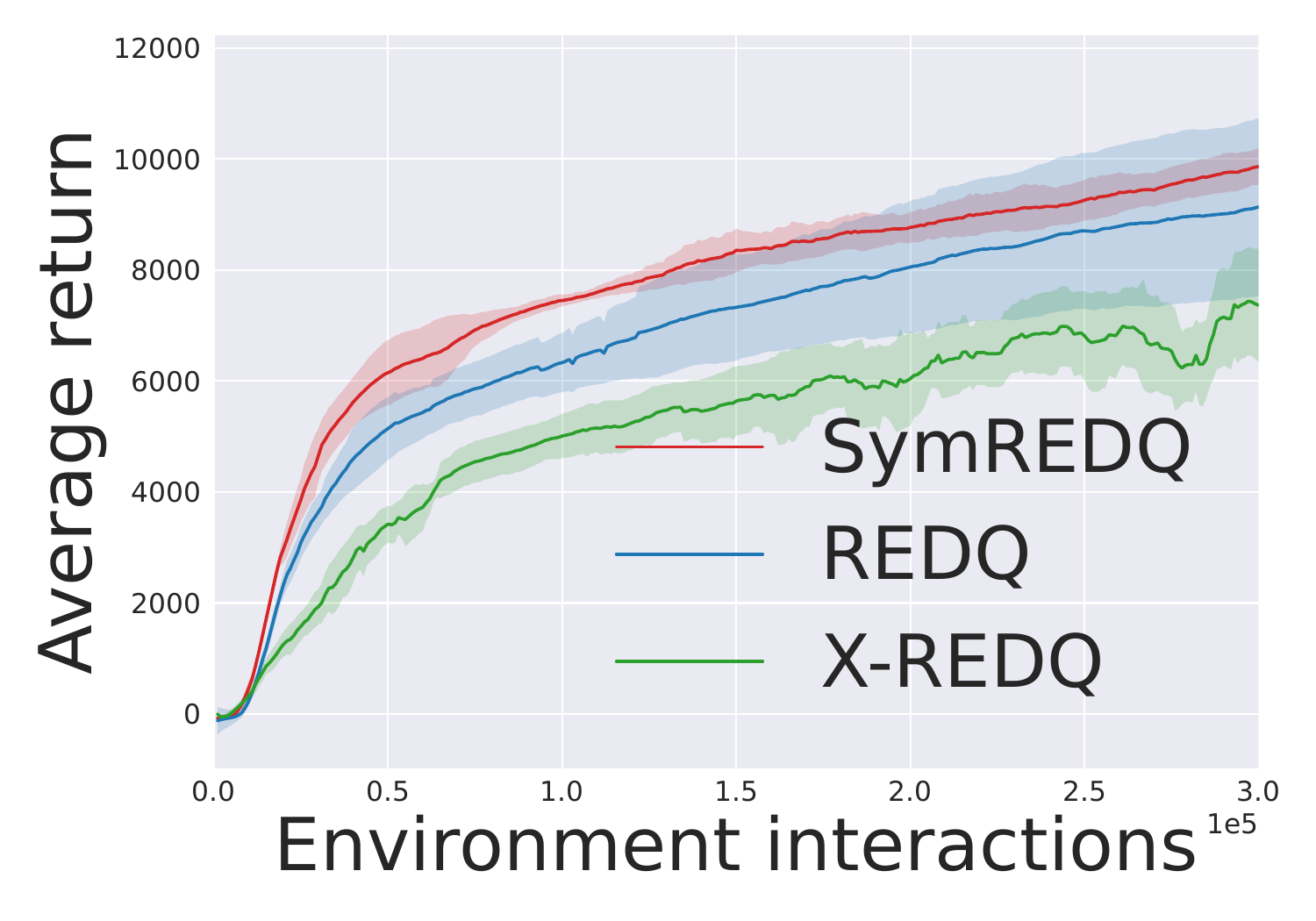} \\
    (a) Humanoid-v2 & (b) Walker2d-v2 & (c) Hopper-v2 & (d) Ant-v2 & (e) HalfCheetah-v2
  \end{tabular}
  \caption{Comparison of SymREDQ, REDQ and $\mathcal{X}$-REDQ for UTD=20.}
  \label{fig:utd20}
\end{figure*}

\section{Experiments}
In our experiments, we investigated the following:

\begin{itemize}
    \item Does distribution correction improve performance?
    \item Can the error distribution be corrected to a symmetric distribution?
\end{itemize}

To answer these questions, we implemented the proposed method in addition to prevalent RL methods and evaluated their performance in continuous control tasks.

\subsection{Setups}

We conducted experiments using five challenging tasks in MuJoCo: Humanoid-v2, Walker2d-v2, Hopper-v2, Ant-v2 and HalfCheetah-v2. The average return in the tasks was evaluated by performing independent trials using five random seeds for each task and algorithm. 
We experimented with two settings: a simple setting without an ensemble and a setting that increased sample efficiency using an ensemble. 

In the former, SymSAC, which combined the proposed method with SAC, was compared with SAC and $\mathcal{X}$-SAC from  \cite{garg2023extreme}.
All algorithms used only one critic network to eliminate the effects of clipped double Q-learning or ensemble. 

In the latter, SymREDQ, which combines the proposed method with REDQ, was compared with REDQ and $\mathcal{X}$-REDQ. $\mathcal{X}$-REDQ is $\mathcal{X}$-SAC with the addition of ensemble and in-target minimization. To verify sample efficiency, experiments were conducted with the UTD ratio set to 20, as in the REDQ paper. Thus 20 updates were performed per iteration. In SymREDQ, the size of the ensemble was set to 20, instead of the default 10 of REDQ, to reduce variance.

The parameter $k$, which represents the parameter updating frequency of the distribution of $\eta$, was set to the value that showed the best performance between 1, 2, and 3. 
Regarding $\mathcal{X}$-SAC and $\mathcal{X}$-REDQ, the temperature parameter $\beta$ was chosen from between 2, 5, 10, and 20 in the same way.
Details of the hyperparameters and performance changes based on $k$, network size and other hyperparameters are provided in the appendix.

\subsection{Comparative Evaluation}

\subsubsection{Comparison at UTD=1 without ensemble}

The average return for the case in which UTD is 1 and there is no ensemble is shown in \cref{fig:utd1}. In Hopper-v2, HalfCheetah-v2 and Ant-v2, our proposed SymSAC method achieved returns similar to those of SAC. For Walker2d-v2 and Humanoid-v2, SymSAC outperformed SAC in terms of returns. 
$\mathcal{X}$-SAC outperformed the other methods only on HalfCheetah-v2, a relatively simple task, but fell far short on the other tasks.

\subsubsection{Comparison at UTD=20 with ensemble}
The average return for the case in which UTD is 20 and the ensemble is used is shown in \cref{fig:utd20}. The final average returns were equivalent to or slightly higher than those of the existing methods. The learning speed was also equivalent to or faster, with clear improvements in certain tasks. The numbers of iterations required to achieve a certain return level are listed in \cref{tab:speed}. In Hopper-v2, HalfCheetah-v2, and Ant-v2, the sample efficiency was significantly improved. In addition, Table 2 shows the final average return and the average return for half of the iterations, and the standard error is described in the appendix. In both cases, SymREDQ achieved equivalent or higher returns compared to REDQ and $\mathcal{X}$-REDQ.

\subsubsection{Discussion}
In SymREDQ, the improvement in learning speed was found to be more significant than that in the final return. This is because the Q-function update amplitude is large in the early stages of learning and the error distribution can change significantly during the learning process. However, using the corrected error distribution with the least squares method leads to stable learning, resulting in an improved learning speed.

We observed that SymSAC and SAC did not show significant improvement, unlike SymREDQ and REDQ. This is likely due to the lack of variance reduction by ensembling in SymREDQ. However, because REDQ is state-of-the-art in terms of sample efficiency and the performance of SymREDQ improved in comparison, the effectiveness of the proposed method is considered sufficient.

In the case of $\mathcal{X}$-SAC and $\mathcal{X}$-REDQ, decent learning was only achieved for Hopper-v2 and HalfCheetah-v2, which are relatively simple tasks. To ensure proper learning, more detailed hyperparameter tuning and algorithm improvement are necessary. 
While extreme Q-learning-based methods were also developed to address the problem of the Bellman error distribution, the experimental results demonstrated that the proposed method was more stable than extreme Q-learning-based methods.

\begin{table}[t]
\centering
\begin{tabular}{l|ccc}
Score & SymREDQ & REDQ & $\mathcal{X}$-REDQ\\
\hline
Humanoid at 5000 & \bfseries172K & 255K & - \\
Walker2d at 3500 & \bfseries77K & 141K & -\\
Hopper at 3000 & \bfseries34K & 46K & -\\
Ant at 5000 & \bfseries101K & 153K & -\\
HalfCheetah at 8000 & \bfseries131K & 196K & -\\
\end{tabular}
\caption{Sample efficiency comparison of SymREDQ, REDQ and $\mathcal{X}$-REDQ. The numbers indicate the amount of data collected until the specified performance level is reached.}
\label{tab:speed}
\end{table}

\begin{table}[t]
\centering
\begin{tabular}{l|ccc}
Amount of data & SymREDQ & REDQ & $\mathcal{X}$-REDQ\\
\hline
Humanoid at 150K & \bfseries4414 & 2904 & 81\\
Humanoid at 300K & \bfseries5252 & 4606 &  82\\
Walker2d at 150K & \bfseries4289 & 3715 & 75\\
Walker2d at 300K & \bfseries4836 & 4587 & 127\\
Hopper at 62K & \bfseries3413 & 3356 & 1433\\
Hopper at 125K & 3428 & \bfseries3529 & 1845\\
Ant at 150K & \bfseries5241 & 4910 & 320\\
Ant at 300K & \bfseries5951 & 5464 & 1924\\
HalfCheetah at 150K & \bfseries8358 & 7328 & 5639 \\
HalfCheetah at 300K & \bfseries9865 & 9138 & 7365 \\
\end{tabular}
\caption{Performance comparison of SymREDQ, REDQ and $\mathcal{X}$-REDQ. This shows the final performance and the performance with half the amount of data collected. Mainly SymREDQ showed significant improvement in performance with half the amount of data collected.}
\label{tab:score}
\end{table}

\subsection{Distribution Plot}

\begin{figure*}[h]
  \centering
  \setlength{\tabcolsep}{0pt}
  \begin{tabular}{ccccc}
    \includegraphics[width=0.41\columnwidth]{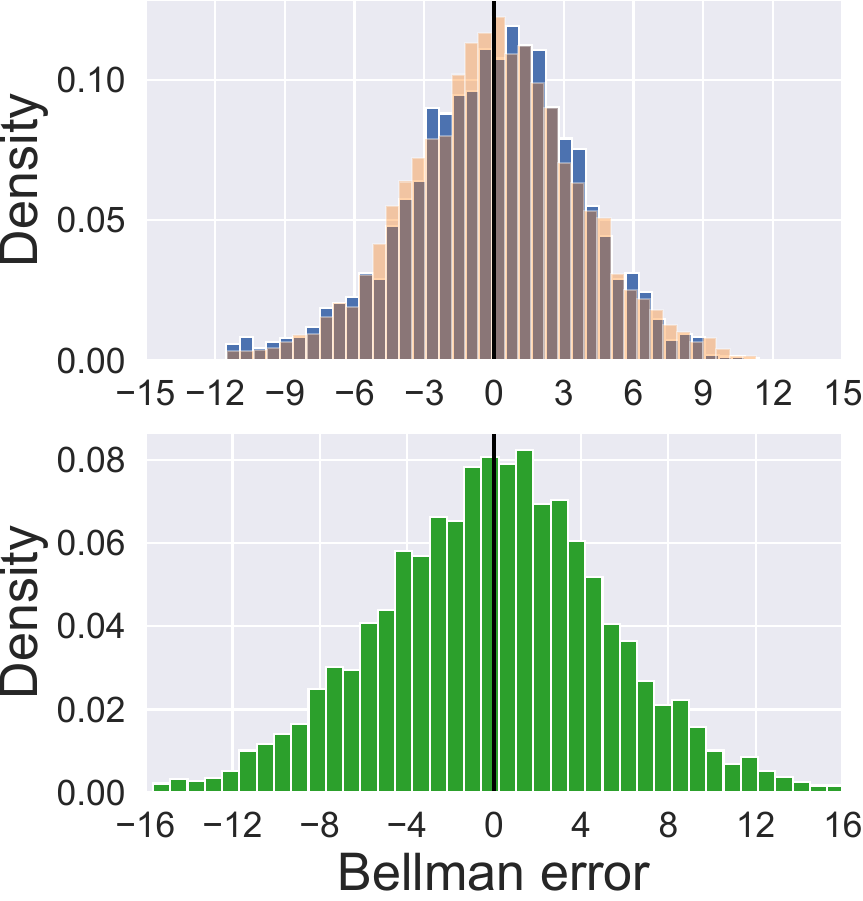} &
    \includegraphics[width=0.41\columnwidth]{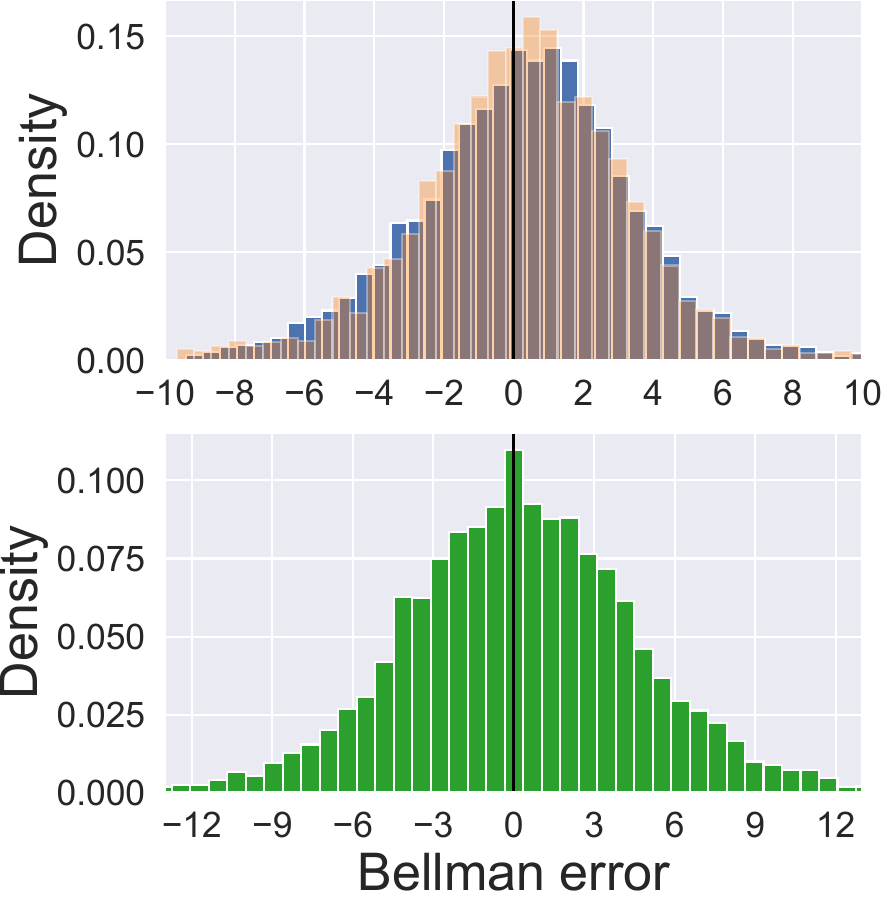} &
    \includegraphics[width=0.41\columnwidth]{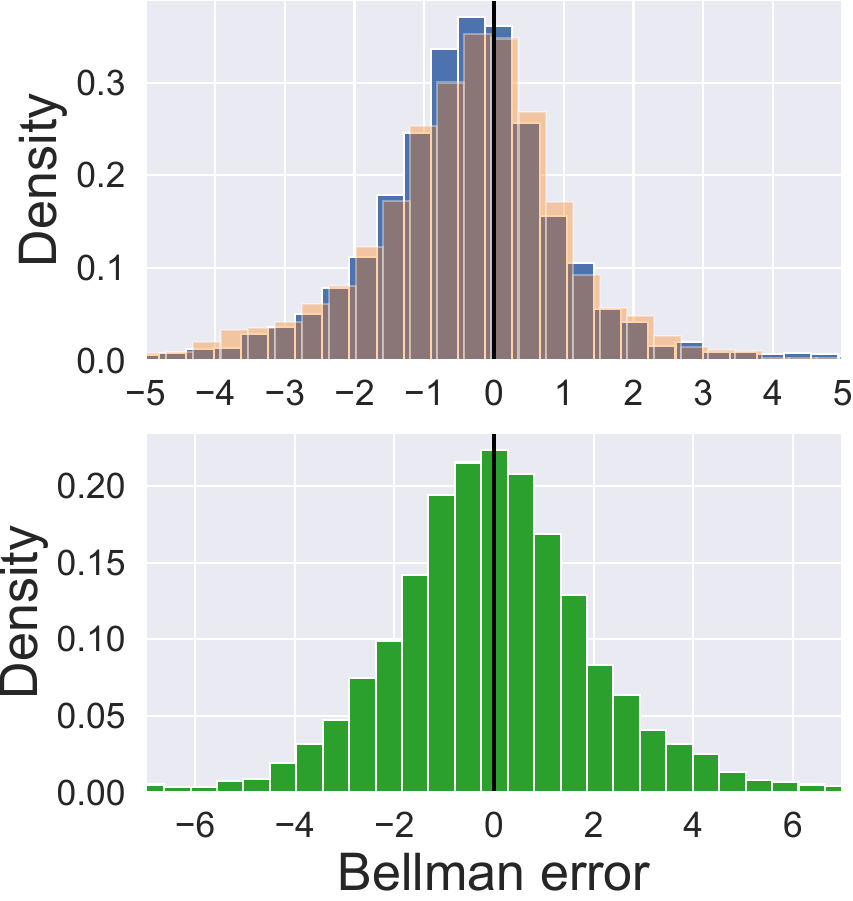} &
    \includegraphics[width=0.41\columnwidth]{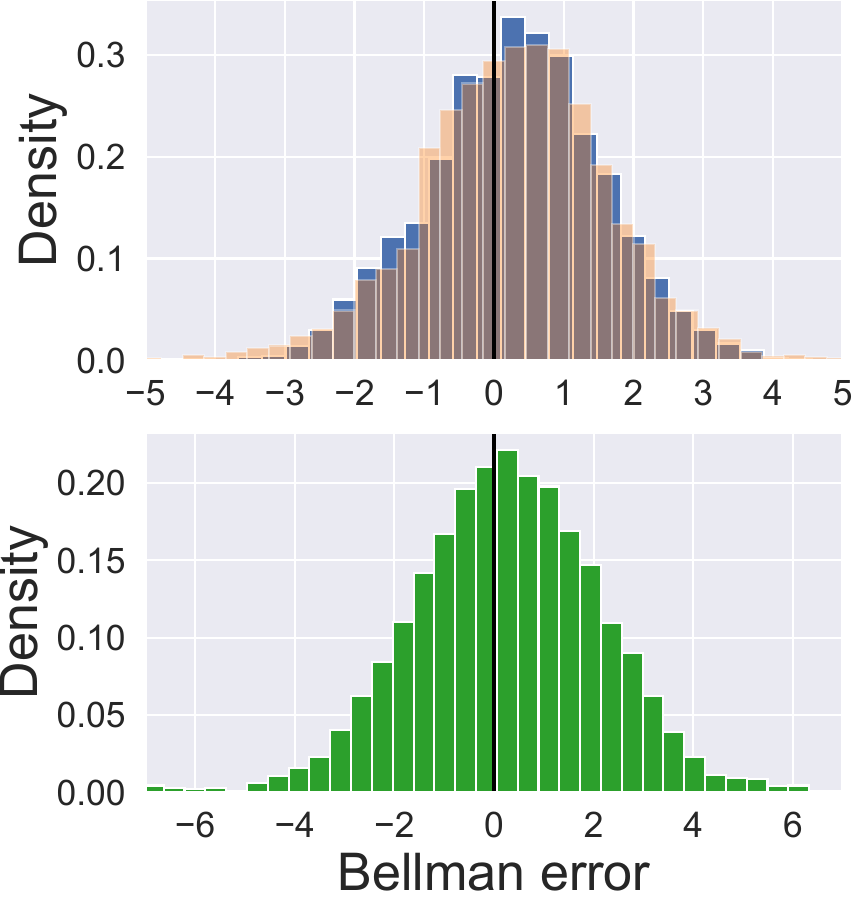} &
    \includegraphics[width=0.41\columnwidth]{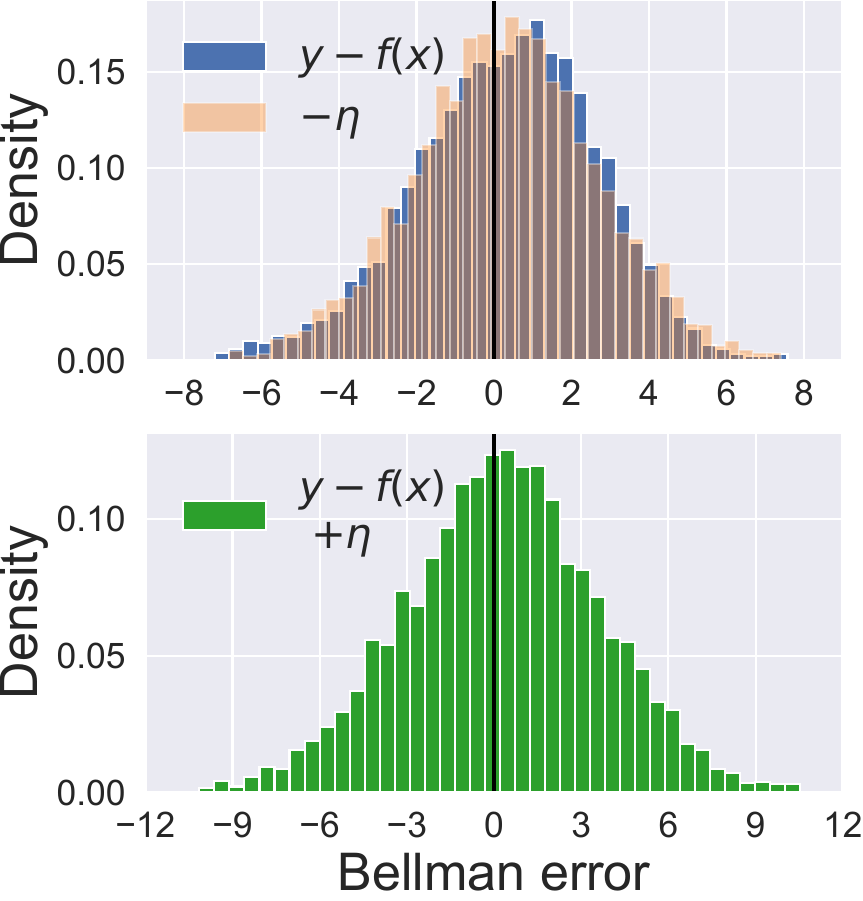} \\
    (a) Humanoid-v2 & (b) Walker2d-v2 & (c) Hopper-v2 & (d) Ant-v2 & (e) HalfCheetah-v2
  \end{tabular}
  \caption{The top figure illustrates the density of pre-corrected Bellman error (blue) and negative values of noise used for correction (orange). It shows how closely the distribution of $\eta$ approaches the distribution of $- \epsilon$. The bottom figure shows the density of the post-corrected error (green), which is the sum of pre-corrected error and noise. This demonstrates the extent to which the distribution approached a symmetric distribution, and the corrected distribution (green) is more symmetric than the pre-corrected distribution (blue).}
  \label{fig:dist}
\end{figure*}

Using SymREDQ, we visualized the actual pre-corrected Bellman error, noise, and post-corrected Bellman error during training. In \cref{fig:dist}, we show how errors were corrected in the steps in which the pre-corrected Bellman error was distorted for each task. During the training process, we ensured that the distribution of noise $\eta$ had a mean value of zero. However, for visualization purposes, we shifted the mean of the noise distribution to be equal to the mean of the Bellman error in order to assess the fitting performance to the Bellman error.

In the figure above, the blue histogram shows the distribution of the skewed Bellman errors. The orange histogram represents the distribution of negative noise added to reduce skewness.
The distribution of noise $\eta$ closely approximates the inverse distribution of the pre-corrected Bellman error $y-f(x)$, meaning that the distribution of $\eta$ approaches the distribution of $-\epsilon$.
In the figure below, the green histogram represents the distribution of the Bellman errors after adding the correction noise $\eta$. The skewness decreased and the symmetry of the error increased compared with the blue distribution. 
The Gumbel distribution assumed in extreme Q-learning was a right-skewed distribution. However, the pre-corrected error distribution (blue) in Humanoid and HalfCheetah showed a left-skewed distribution, and not meeting the assumption may be leading to a decrease in performance.
The figures for other steps are provided in the appendix.

\section{Related Work}
In our study, we addressed the problem of the skewed error distribution in online RL. Therefore, we introduce several representative approaches for online RL and approaches related to the error distribution.

\subsection{Online Reinforcement Learning}

In online RL, the goal is to maximize rewards through iterations with the environment. While table-based RL with methods such as Q-learning \cite{watkins1992qlearning} and SARSA \cite{sarsa} has shown promising results, the combination of deep learning with Q-learning in DQN \cite{mnih2013dqn} has achieved performance surpassing human-level capabilities in game playing. For continuous action tasks such as robotics, policy gradient-based methods such as TRPO \cite{schulman2017trust} and PPO \cite{schulman2017proximal}, as well as actor-critic-based methods such as SAC \cite{haarnoja2018soft} and TD3 \cite{fujimoto2018addressing}, have been used with good performance.
All of these existing methods that use function approximation employ least squares to estimate the value function, and can be combined with the proposed method.

When iteration with the environment has high cost, such as in robotics, sample efficiency becomes important in RL. 
To increase sample efficiency, it is efficient to set UTD $\gg 1$. However, this leads to accumulation of overestimation bias, making learning difficult. To address this issue, REDQ combines ensemble and in-target minimization across a random subset of Q functions to adjust variance and bias suppression, enabling learning even with UTD $\gg 1$. By combining REDQ with SAC, it achieved a sample efficiency equal to or greater than that of state-of-the-art model-based algorithms for MuJoCo benchmark. 
In DroQ \cite{hiraoka2022dropout} and MEPG \cite{he2022mepg}, the ensemble-related computational efficiency was improved using dropout \cite{dropout2014} as a substitute for the ensemble. 
It can also be used for reducing computational cost of SymREDQ.

\subsection{Error Distribution}
In maximum likelihood estimation, a certain distribution is assumed to be the error distribution, and the estimation can be inaccurate when the actual distribution differs from the assumption. The least squares method can be regarded as a maximum likelihood estimation, assuming a normal distribution as the error distribution. It has been experimentally demonstrated that applying the least squares method to errors with nonnormal distributions leads to inaccurate estimation results \cite{ls_nongauss1988,ls_nongauss2017}. Therefore, when using the least squares method in machine learning, learning is sometimes performed by transforming the distribution of the data into normal distribution. Many transformations focus on skewness, and it is well-known that logarithmic transformation is used for right-skewed distributions. The Box-Cox transformation \cite{box1964analysis} and Yeo-Johnson \cite{yeo} transformation can adjust the degree of skewness by parameters, and flexible transformations can be performed by estimating the parameters with maximum likelihood estimation, such that the transformed distribution becomes closer to normal distribution. Applying these methods to reinforcement learning is a task for future research.

The least squares method is also used in RL, where normal distribution is required as an error distribution. However, based on extreme value theory,  \cite{garg2023extreme} showed that the error distribution follows the Gumbel distribution, which is an extreme value distribution because the Bellman equation contains a maximum operator. They addressed this problem using the Gumbel regression, which is a maximum likelihood estimation assuming that the error distribution follows the Gumbel distribution. Gumbel regression can directly estimate optimal soft-value functions without sampling from policies and avoids evaluating out-of-distribution actions, thus significantly improving performance in offline RL settings. However, in online RL, not much improvement is observed. First, Gumbel regression, which assumes a Gumbel distribution as an error distribution, is unstable. It assumes an asymmetric error distribution and has an exponential term in the loss function, making the gradient prone to being large. Second, the actual error distribution may differ from the Gumbel distribution because the state-action pairs taking the maximum are not independent, violating the premises of extreme value theory. Furthermore, because the properties of algorithms such as double Q-learning and target networks can affect the actual error distribution, it is difficult to assume a single static error distribution for maximum likelihood estimation. Therefore, the error distribution can vary depending on the task and the algorithm, highlighting the need for a method that can be applied to arbitrary error distributions.

The Gauss-Markov theorem states that under the assumptions of homoscedasticity and uncorrelated errors, the linear unbiased estimator estimated by the least squares method has the smallest variance and is the best linear unbiased estimator (BLUE). In RL, a small variance is not necessarily good, and the use of linear parameters is infrequent, but there is room to verify the impact of meeting these assumptions.

\section{Conclusion}

This study addresses the issue of the error distribution being far from the assumed normal distribution in online RL methods that use the least squares method. The proposed method adds noise that cancels out the distortion in the error, making it closer to a normal distribution. Experiments conducted on the MuJoCo benchmark demonstrate that the proposed method can symmetrically correct the error distribution, leading to improved sample efficiency and performance equivalent to or better than that of the state-of-the-art RL method. 
This is because stable learning with an error distribution close to the normal distribution can be achieved in the early stages of learning, resulting in improved efficiency.

In RL using the estimated values as the target, the error distribution is influenced by various factors, such as the Bellman operator and bias-suppression algorithms, which can pose problems during learning. 
However, there is limited research focusing on error distribution in RL, indicating substantial potential for future advancements in this field.

\section{Acknowledgments}
We would like to thank Jen-Yen Chang and Thomas Westfechtel for their valuable proofreading of this manuscript. This work was partially supported by JST Moonshot R\&D Grant Number JPMJPS2011, CREST Grant Number JPMJCR2015 and Basic Research Grant (Super AI) of Institute for AI and Beyond of the University of Tokyo. T.O. was partially supported by JSPS KAKENHI Grant Number JP23K18476.

\onecolumn
\appendix

\section{Appendix}
\date{}

\section{Hyperparameters and performance changes.}

The list of hyperparameters is shown in \cref{tab:hp}, \cref{tab:beta} and \cref{tab:interval}. The implementation is based on REDQ, and almost the same hyperparameters as REDQ are used. REDQ is based on SAC and inherits its hyperparameters, so similar hyperparameters are used for SAC. Unlike SAC \cite{haarnoja2018soft}, the number of critics is set to 1 instead of 2 to eliminate the influence of ensembles.
As for $\mathcal{X}$-SAC and $\mathcal{X}$-REDQ, we experimented based on the official implementation of extreme Q-learning \cite{garg2023extreme}.
To suppress the increase in variance of Q-values due to the addition of noise, the ensemble size for SymREDQ was increased to 20. 
For SymREDQ, REDQ and $\mathcal{X}$-REDQ experiments, UTD=20 was used, while for SAC experiments, UTD=1 was used.
The GMM cluster numbers are both set to 10 for SymREDQ and SymSAC. The values in \cref{tab:hp} are used universally for all tasks.

\begin{table}[h]
\centering
\caption{Hyperparameters}
\label{tab:hp}
\begin{tabular}{l|cccc|cc}
Hyperparameter & SymSAC & SAC & SymREDQ & REDQ & $\mathcal{X}$-SAC & $\mathcal{X}$-REDQ \\
\hline
Optimizer & \multicolumn{4}{c|}{Adam \cite{kingma2014adam}} & \multicolumn{2}{c}{Adam} \\
Learning rate & \multicolumn{4}{c|}{$3\cdot10^{-4}$} & \multicolumn{2}{c}{$1\cdot10^{-4}$}\\
Discount $(\gamma)$ & \multicolumn{4}{c|}{0.99} & \multicolumn{2}{c}{0.99} \\
Target smoothing coefficient $(\rho)$ & \multicolumn{4}{c|}{0.005} & \multicolumn{2}{c}{0.005} \\
Replay buffer size & \multicolumn{4}{c|}{$10^6$} & \multicolumn{2}{c}{$10^6$}\\
Number of hidden layers & \multicolumn{4}{c|}{2} & \multicolumn{2}{c}{2}\\
Number of hidden units per layer & \multicolumn{4}{c|}{256} & \multicolumn{2}{c}{1024}\\
Mini-batch size & \multicolumn{4}{c|}{256} & \multicolumn{2}{c}{1024}\\ 
Nonlinearity & \multicolumn{4}{c|}{ReLU} & \multicolumn{2}{c}{ReLU}\\ 
Random starting data & \multicolumn{4}{c|}{5000} & \multicolumn{2}{c}{5000} \\
\hline
Number of critics & 1 & 1 & 20 & 10 & 1 & 10 \\
In-target minimization parameter & 1 & 1 & 2 & 2 & 1 & 2 \\
UTD ratio & 1 & 1 & 20 & 20 & 1 & 20\\
Number of clusters in GMM & 10 & - & 10 & - & - & -\\
\end{tabular}
\end{table}

\subsection{GMM update interval $k$}
\begin{figure}[]
  \begin{minipage}[b]{0.32\columnwidth}
    \centering
    \includegraphics[width=\columnwidth]{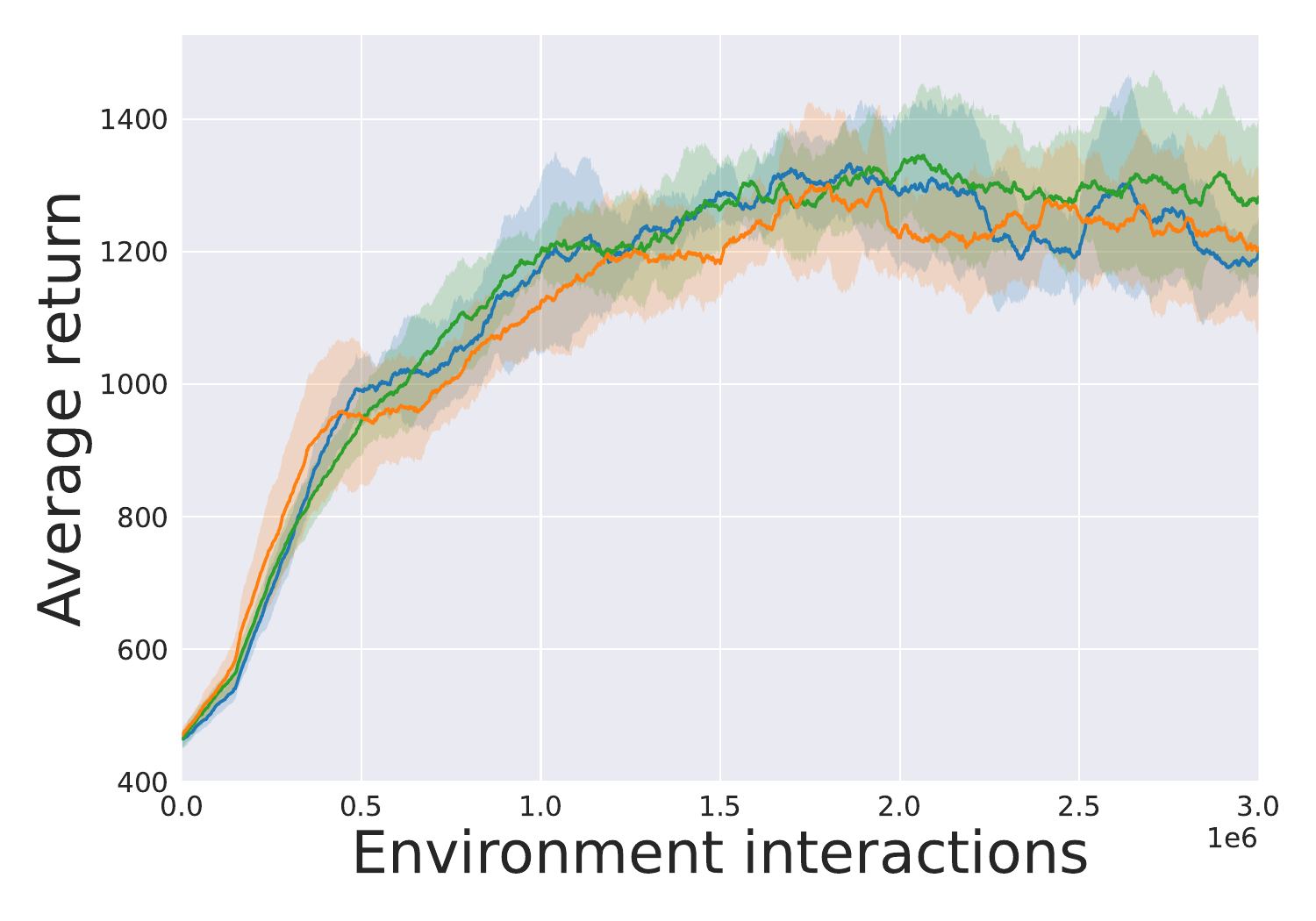}
    \subcaption{Humanoid-v2}
  \end{minipage}
 \begin{minipage}[b]{0.32\columnwidth}
    \centering
    \includegraphics[width=\columnwidth]{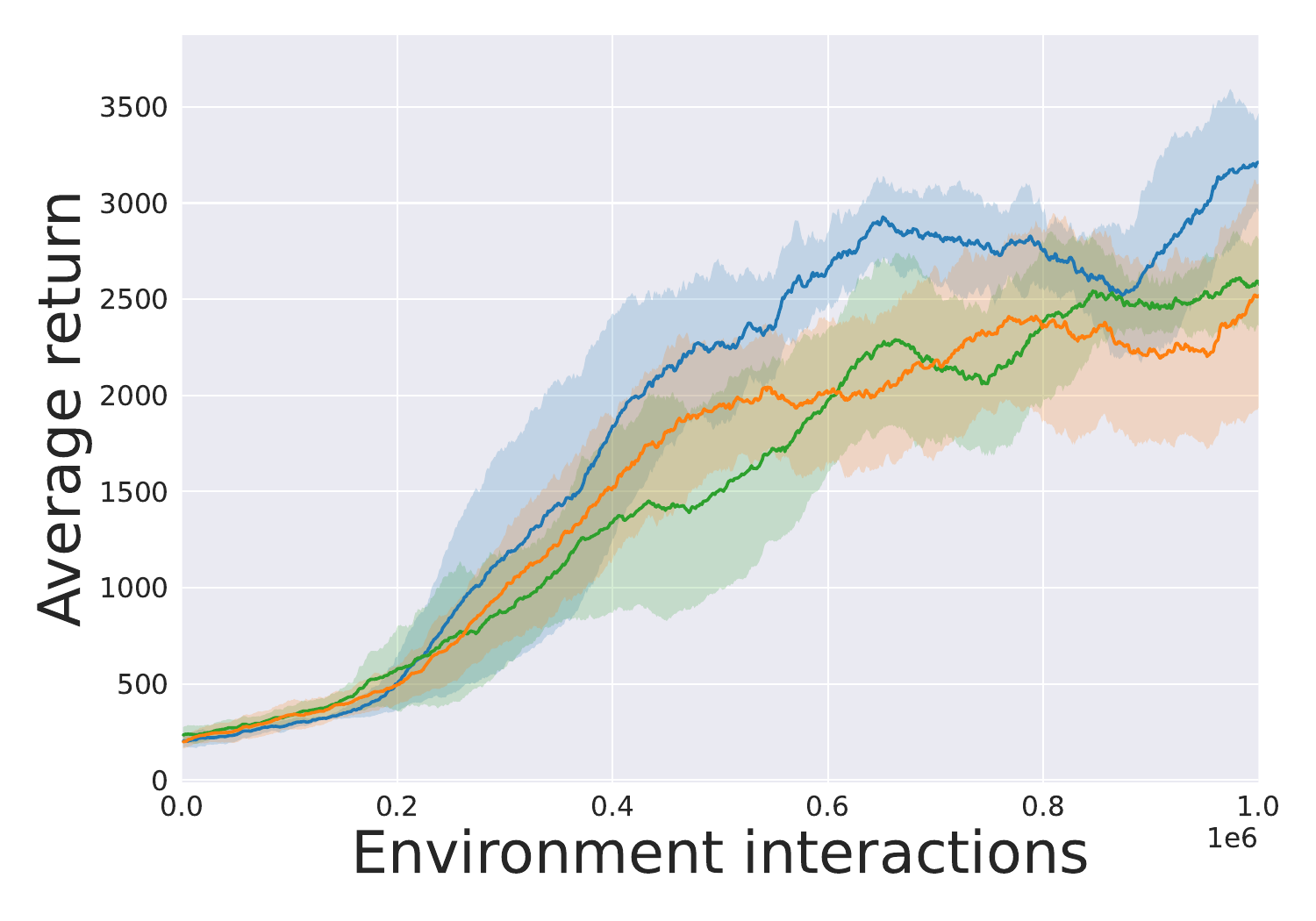}
    \subcaption{Walker2d-v2}
  \end{minipage}
  \begin{minipage}[b]{0.32\columnwidth}
    \centering
    \includegraphics[width=\columnwidth]{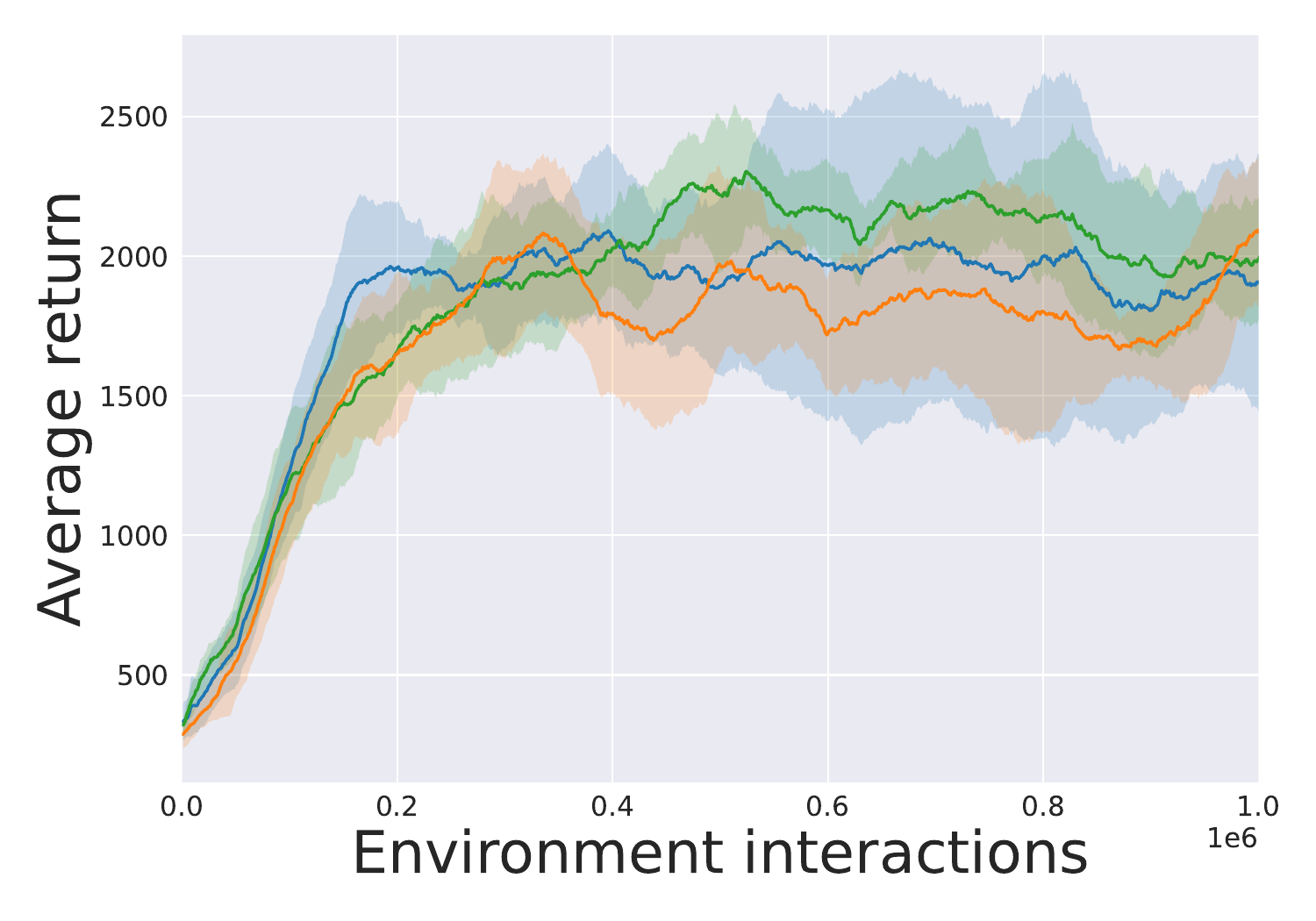}
    \subcaption{Hopper-v2}
  \end{minipage}
  \begin{minipage}[b]{0.32\columnwidth}
    \centering
    \includegraphics[width=\columnwidth]{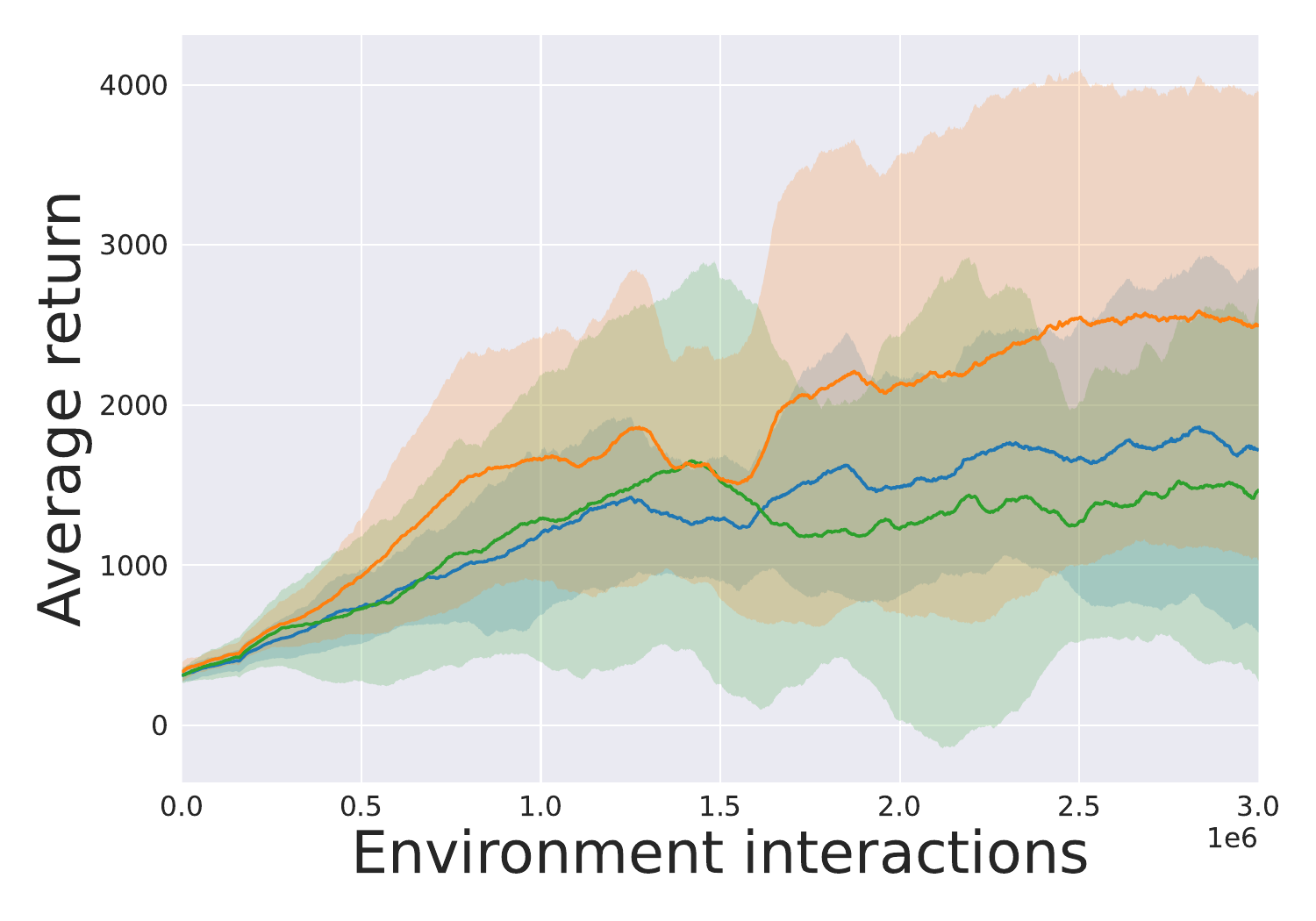}
    \subcaption{Ant-v2}
  \end{minipage}
  \begin{minipage}[b]{0.32\columnwidth}
    \centering
    \includegraphics[width=\columnwidth]{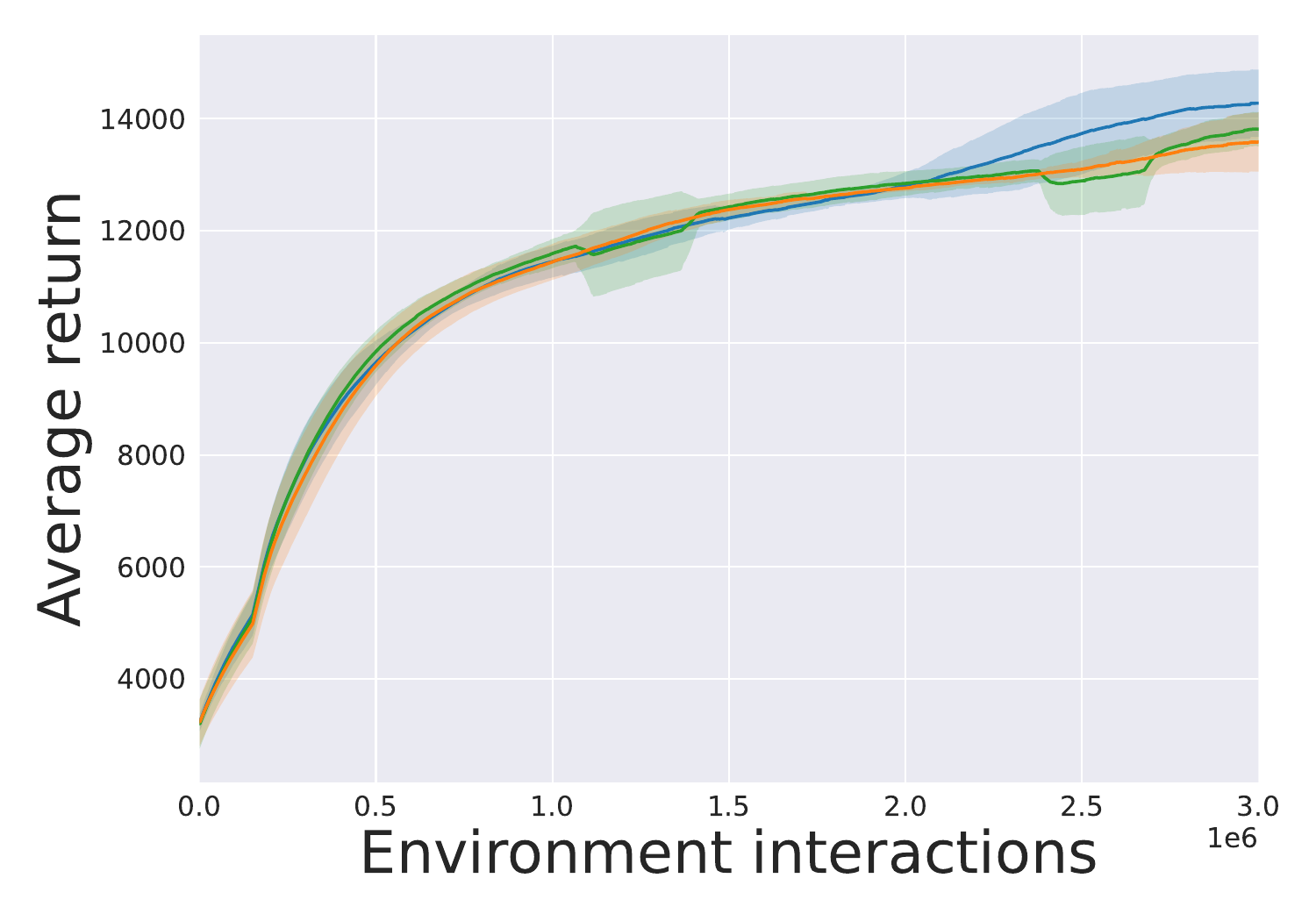}
    \subcaption{HalfCheetah-v2}
  \end{minipage}
  \hspace{1cm}
  \begin{minipage}[b]{0.20\columnwidth}
    \centering
    \includegraphics[width=\columnwidth]{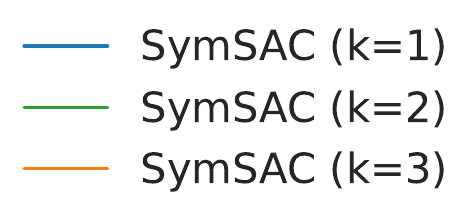}
    \vspace{0.3cm}
    \subcaption*{}
  \end{minipage}
  \caption{Performance changes based on the GMM update interval $k$ in SymSAC}
  \label{fig:k_symsac}
\end{figure}

\begin{figure}[]
  \begin{minipage}[b]{0.32\columnwidth}
    \centering
    \includegraphics[width=\columnwidth]{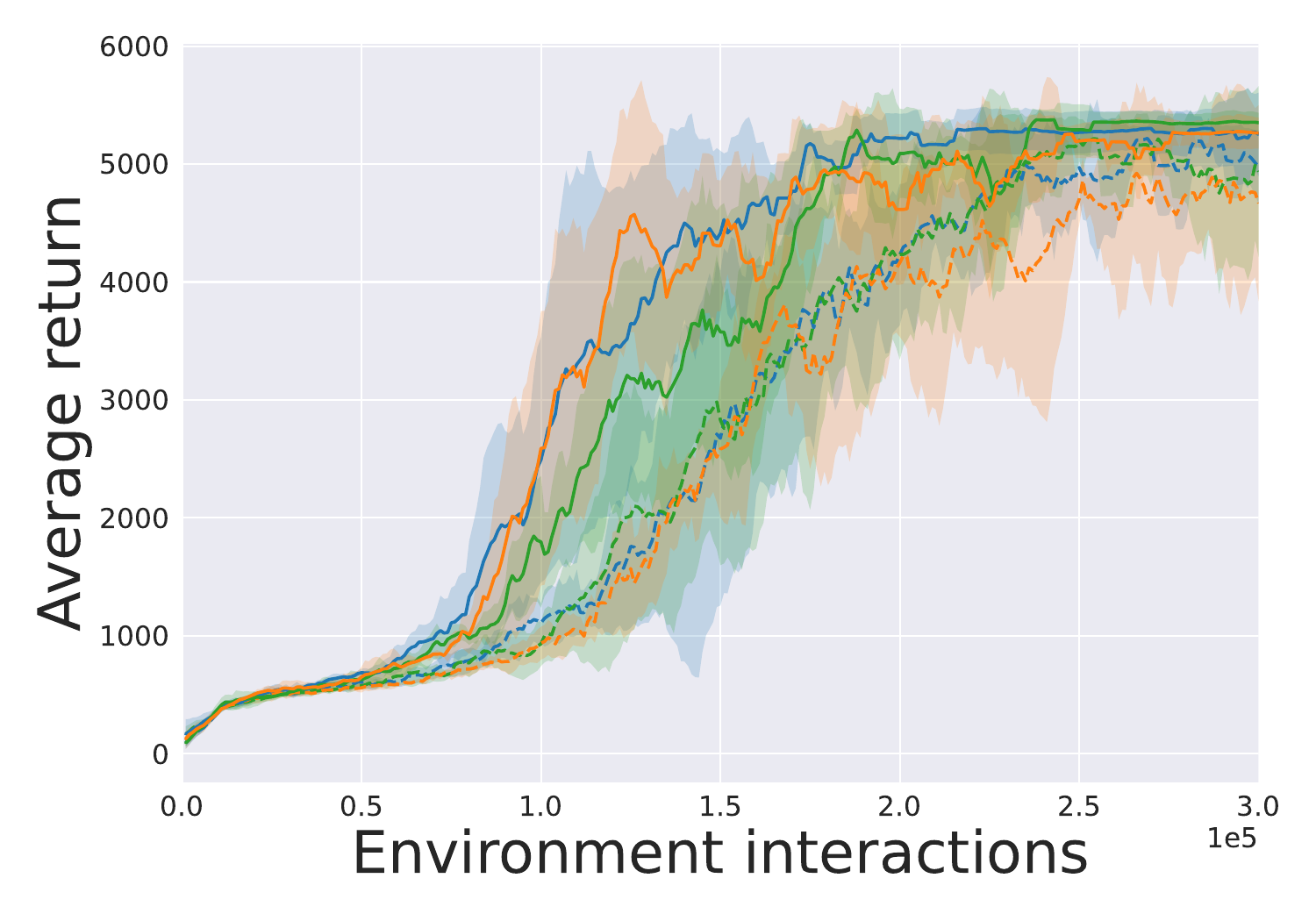}
    \subcaption{Humanoid-v2}
  \end{minipage}
 \begin{minipage}[b]{0.32\columnwidth}
    \centering
    \includegraphics[width=\columnwidth]{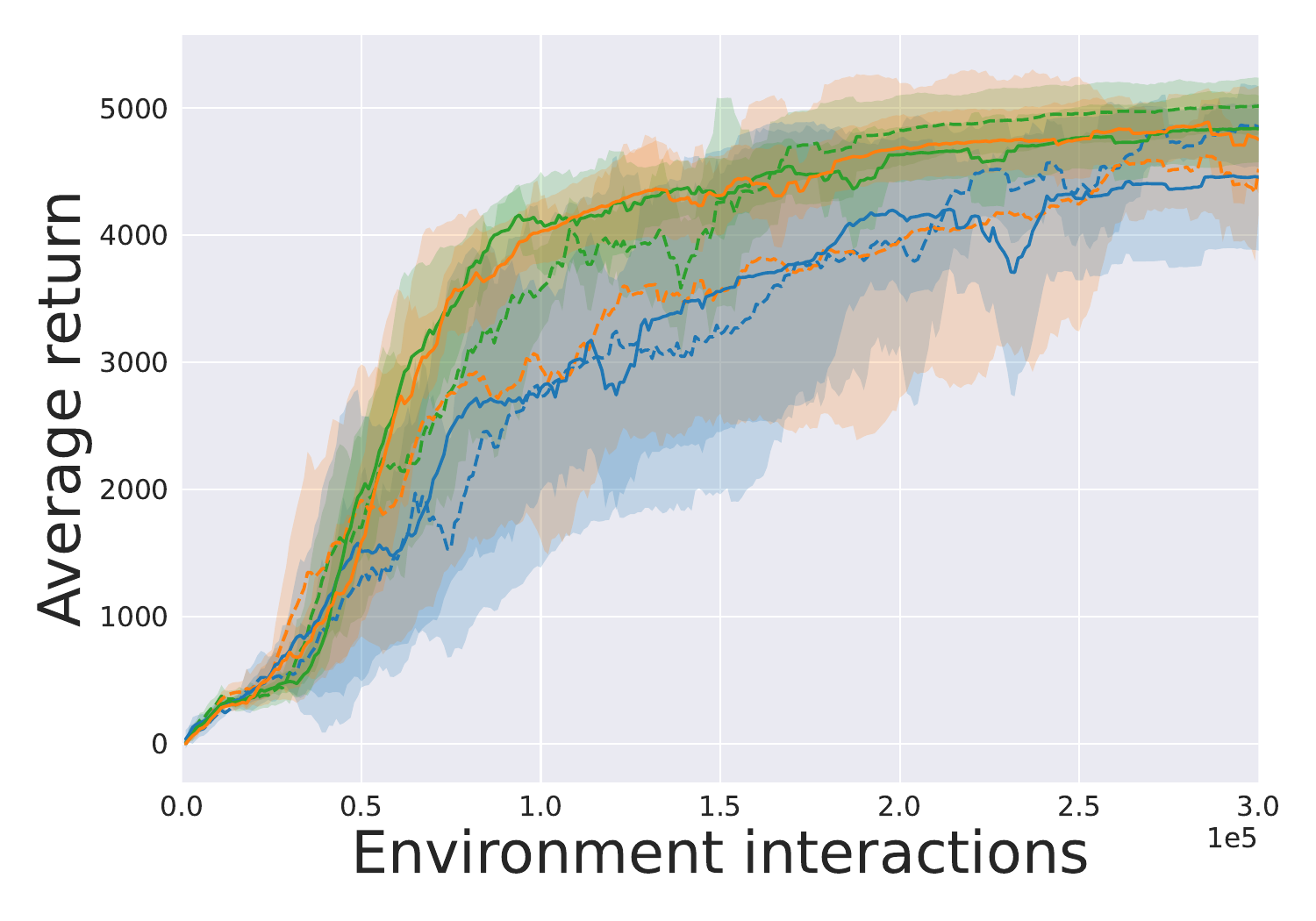}
    \subcaption{Walker2d-v2}
  \end{minipage}
  \begin{minipage}[b]{0.32\columnwidth}
    \centering
    \includegraphics[width=\columnwidth]{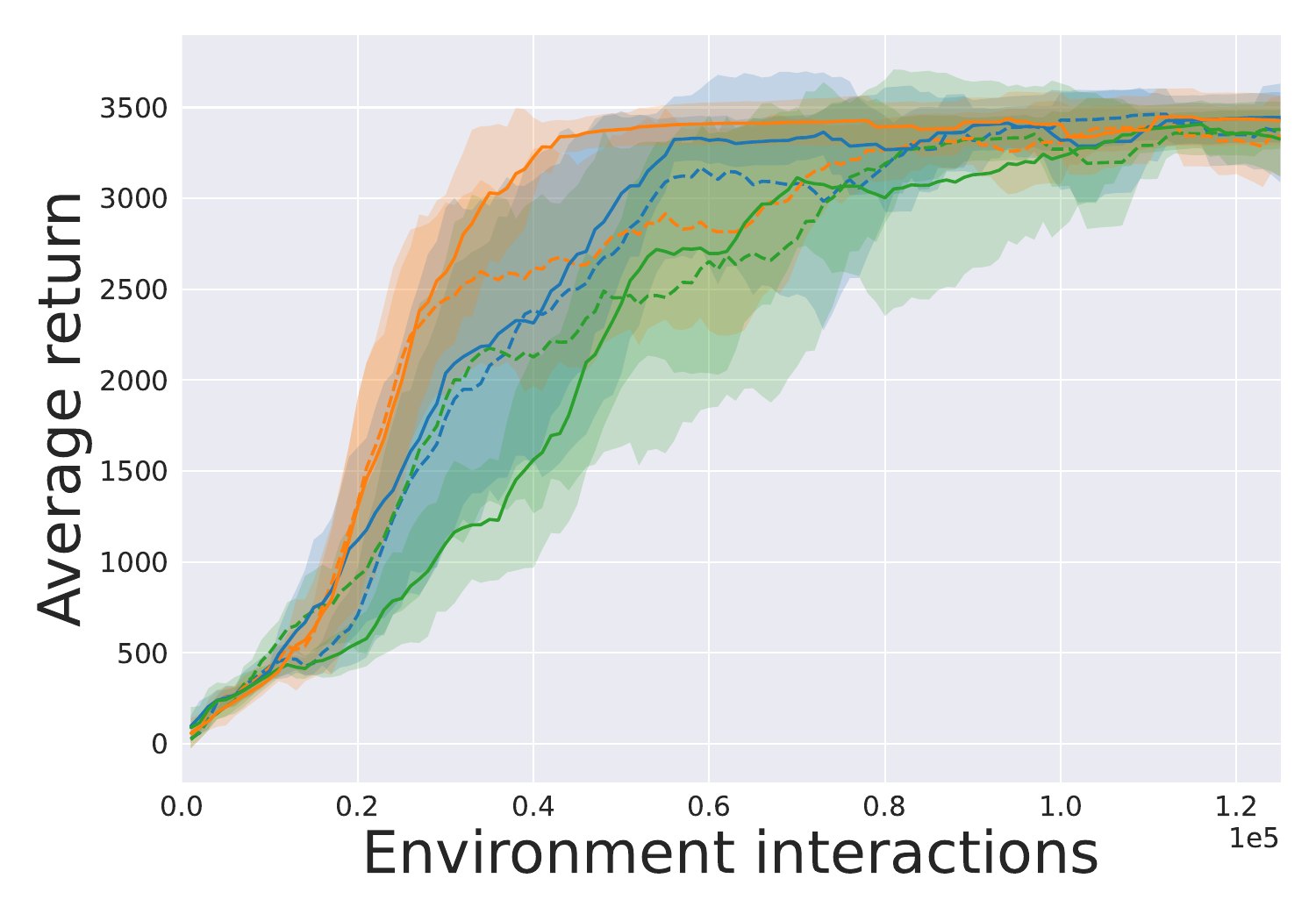}
    \subcaption{Hopper-v2}
  \end{minipage}
  \begin{minipage}[b]{0.32\columnwidth}
    \centering
    \includegraphics[width=\columnwidth]{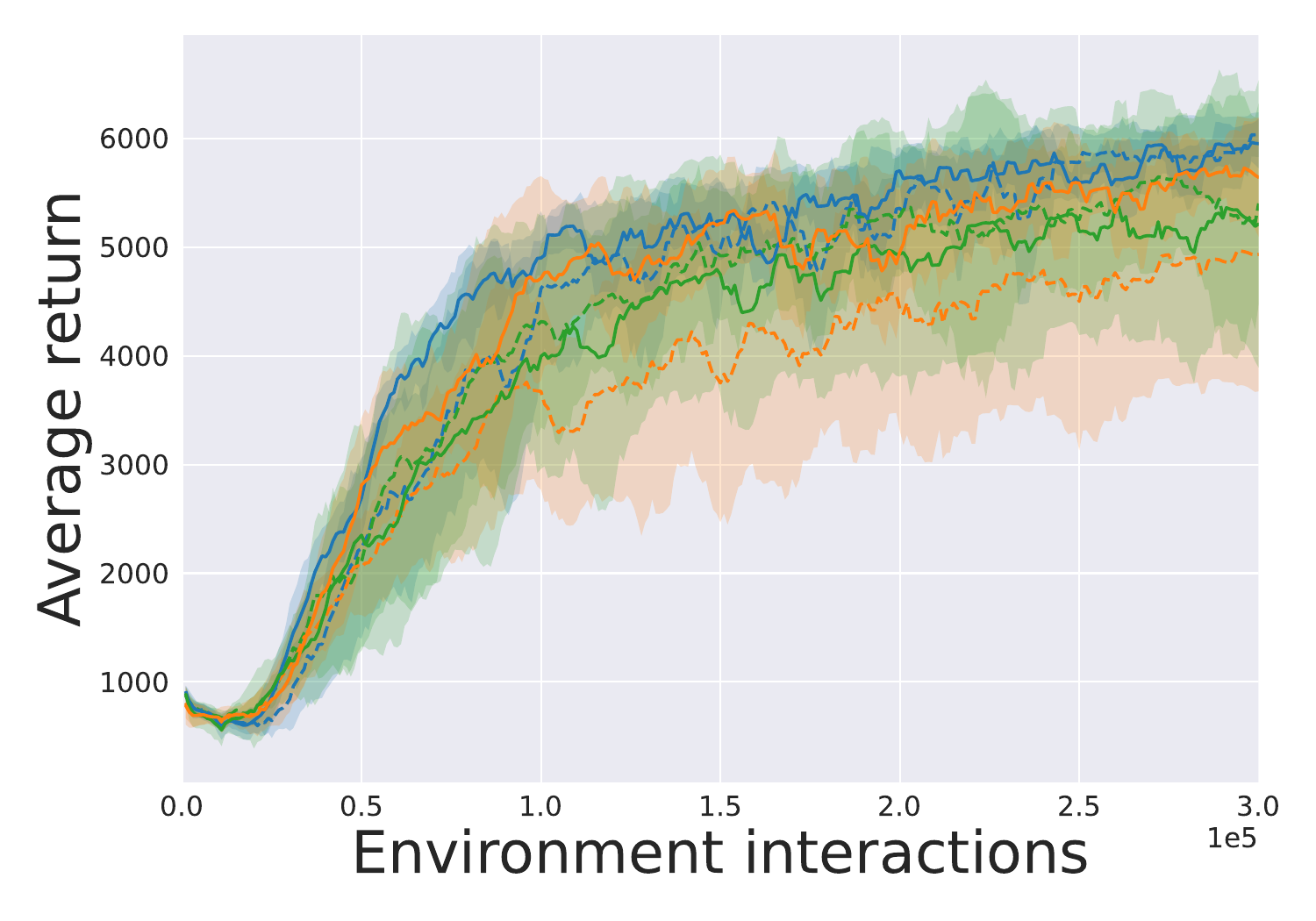}
    \subcaption{Ant-v2}
  \end{minipage}
  \begin{minipage}[b]{0.32\columnwidth}
    \centering
    \includegraphics[width=\columnwidth]{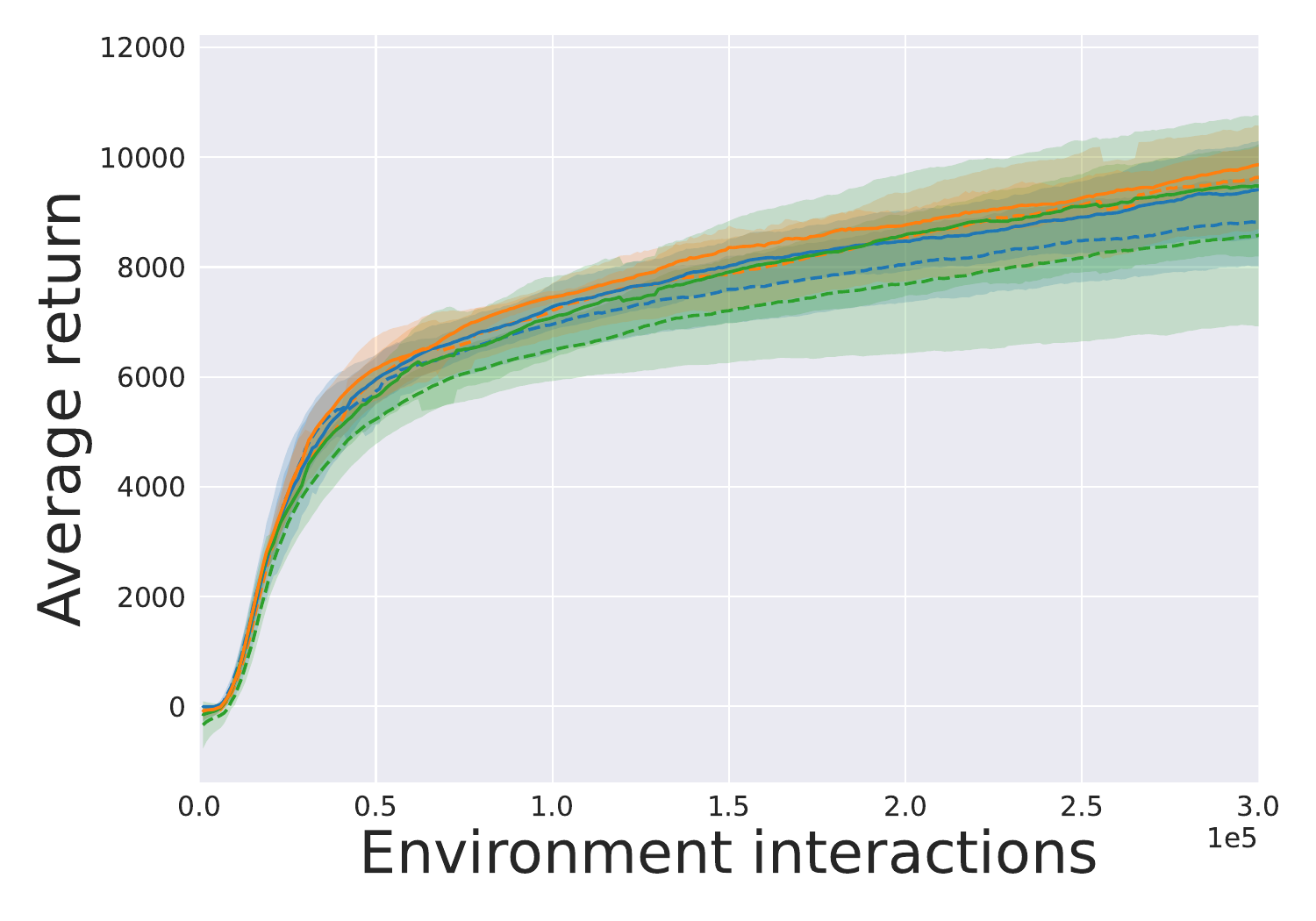}
    \subcaption{HalfCheetah-v2}
  \end{minipage}
  \hspace{1cm}
  \begin{minipage}[b]{0.20\columnwidth}
    \centering
    \includegraphics[width=\columnwidth]{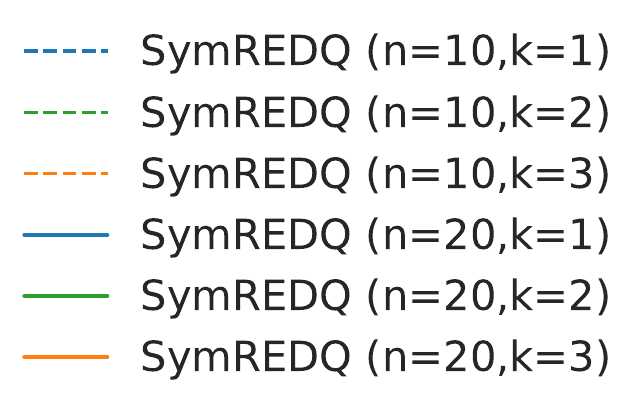}
    \vspace{0.3cm}
    \subcaption*{}
  \end{minipage} 
  \caption{Performance changes based on the GMM update interval $k$ and the number of critics $n$ in SymREDQ}
  \label{fig:kn_symredq}
\end{figure}

The update interval of GMM is adjusted for each task as shown in \cref{tab:interval}. This is because the distribution of the Bellman error during the learning process varies depending on the task and algorithm, requiring appropriate tuning.
The performance changes for SymSAC and SymREDQ based on the GMM update interval $k$ are shown in \cref{fig:k_symsac} and \cref{fig:kn_symredq}, respectively. The performance of online RL is sensitive to hyperparameters, prone to local optima, and has a large variance. Therefore, the performance decreased in some settings. However, the performance in many tasks remained stable regardless of $k$. Notably, SymSAC on Humanoid-v2 and SymREDQ on HalfCheetah-v2, where the proposed method showed significant improvement, have become much less sensitive to hyperparameters.

\subsection{Number of critics}
For SymREDQ, the number of critics was set to 20 to reduce variance by adding noise, while for REDQ and $\mathcal{X}$-REDQ, it was set to the REDQ's default 10. To see the effect of this difference, we verified the performance with 10 and 20 critics. For SymREDQ, performance generally improved for all tasks when $n=20$. For REDQ, performance improved for Humanoid and Walker2d when $n=20$, but for other tasks, it was equivalent or worse than when $n=10$. This suggests that excessive variance reduction can lead to performance degradation. Also, for Humanoid and Walker2d, using a larger number of critics for SymREDQ may improve performance. For $\mathcal{X}$-REDQ, performance improved for Ant when $n=20$, but there was not much change for other tasks.

\begin{table}[htbp]
  \centering
  \begin{minipage}{.5\textwidth}
    \centering
    \begin{tabular}{l|cc}
    Environment & SymSAC & SymREDQ\\
    \hline
    Hopper-v2 & 1 & 3\\
    Walker2d-v2 & 1 & 2\\
    HalfCheetah-v2 & 1 & 3\\
    Ant-v2 & 3 & 1 \\
    Humanoid-v2 & 2 & 1 \\
    \end{tabular}
    \caption{GMM update interval $k$}
    \label{tab:interval}
  \end{minipage}%
  \begin{minipage}{.5\textwidth}
    \centering
    \begin{tabular}{l|cc}
    Environment & $\mathcal{X}$-SAC & $\mathcal{X}$-REDQ \\
    \hline
    Hopper-v2 & 20 & 20\\
    Walker2d-v2 & 20 & 20\\
    HalfCheetah-v2 & 20 & 5\\
    Ant-v2 & 20 & 20 \\
    Humanoid-v2 & 2 & 20 \\
    \end{tabular}
    \caption{Temperature $\beta$ for Gumbel regression}
    \label{tab:beta}
  \end{minipage}
\end{table}

\begin{table*}[]
\centering
\caption{Performance comparison of SymREDQ, REDQ and $\mathcal{X}$-REDQ. This shows the final performance and the performance with half the amount of data collected along with their respective standard errors.}
\label{tab:score_sde}
\begin{tabular}{l|ccc}
Amount of data & SymREDQ & REDQ & $\mathcal{X}$-REDQ\\
\hline
Humanoid at 150K & 4414$\pm$349 & 2904$\pm$570 & 81$\pm$16\\
Humanoid at 300K & 5252$\pm$177 & 4606$\pm$734 & 82$\pm$16\\
Walker2d at 150K & 4289$\pm$156 & 3715$\pm$299 & 75$\pm$54\\
Walker2d at 300K & 4836$\pm$131 & 4587$\pm$215 & 127$\pm$86\\
Hopper at 62K & 3413$\pm$58 & 3356$\pm$65 & 1433$\pm$224\\
Hopper at 125K & 3428$\pm$67 & 3529$\pm$35 & 1845$\pm$273\\
Ant at 150K & 5241$\pm$180 & 4910$\pm$471 & 320$\pm$41\\
Ant at 300K & 5951$\pm$113 & 5464$\pm$352 & 1924$\pm$805\\
HalfCheetah at 150K & 8358$\pm$197 & 7328$\pm$480 & 5639$\pm$313 \\
HalfCheetah at 300K & 9865$\pm$163 & 9138$\pm$802 & 7365$\pm$513 \\
\end{tabular}
\end{table*}

\begin{figure}[]
  \begin{minipage}[b]{0.32\columnwidth}
    \centering
    \includegraphics[width=\columnwidth]{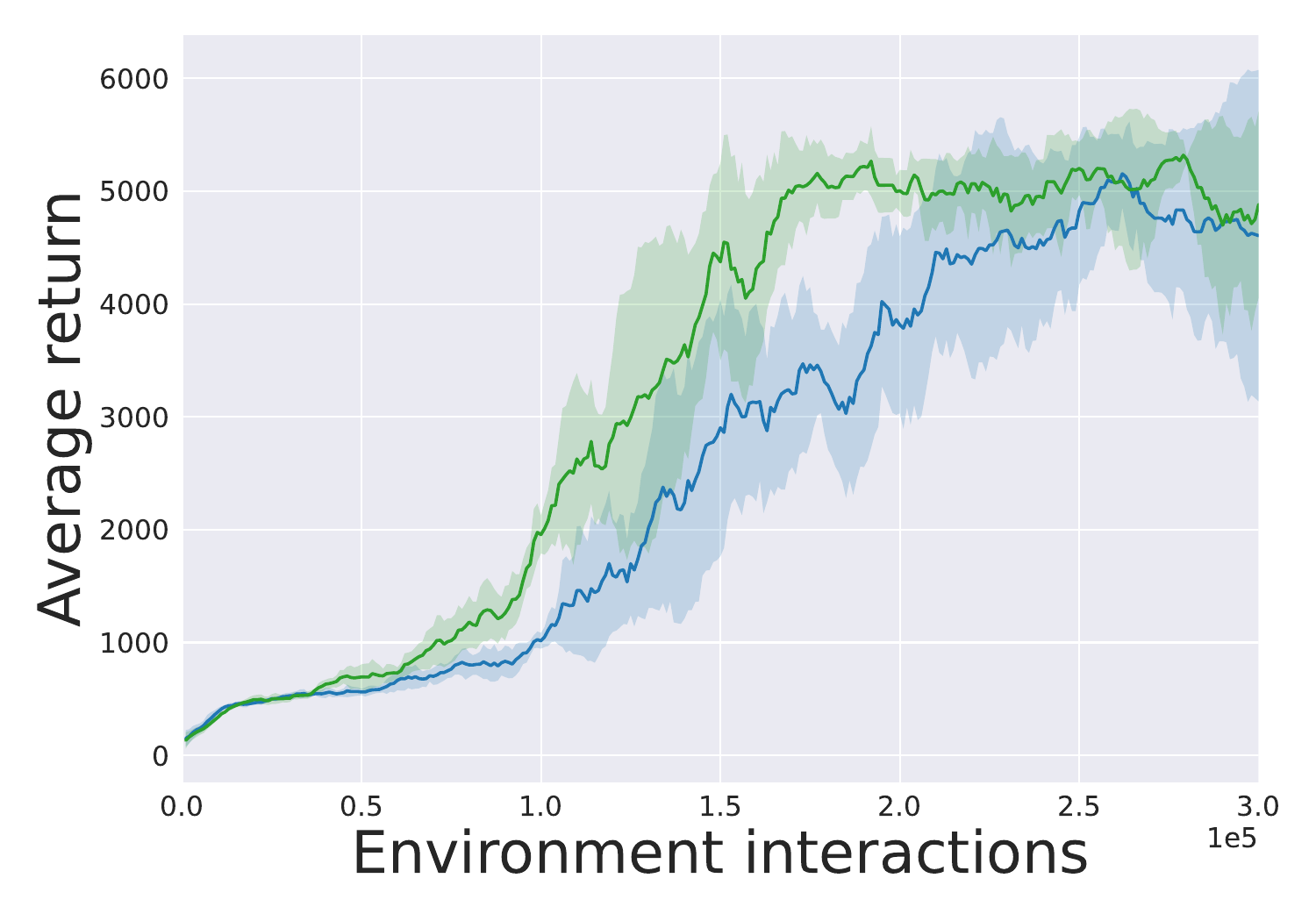}
    \subcaption{Humanoid-v2}
  \end{minipage}
 \begin{minipage}[b]{0.32\columnwidth}
    \centering
    \includegraphics[width=\columnwidth]{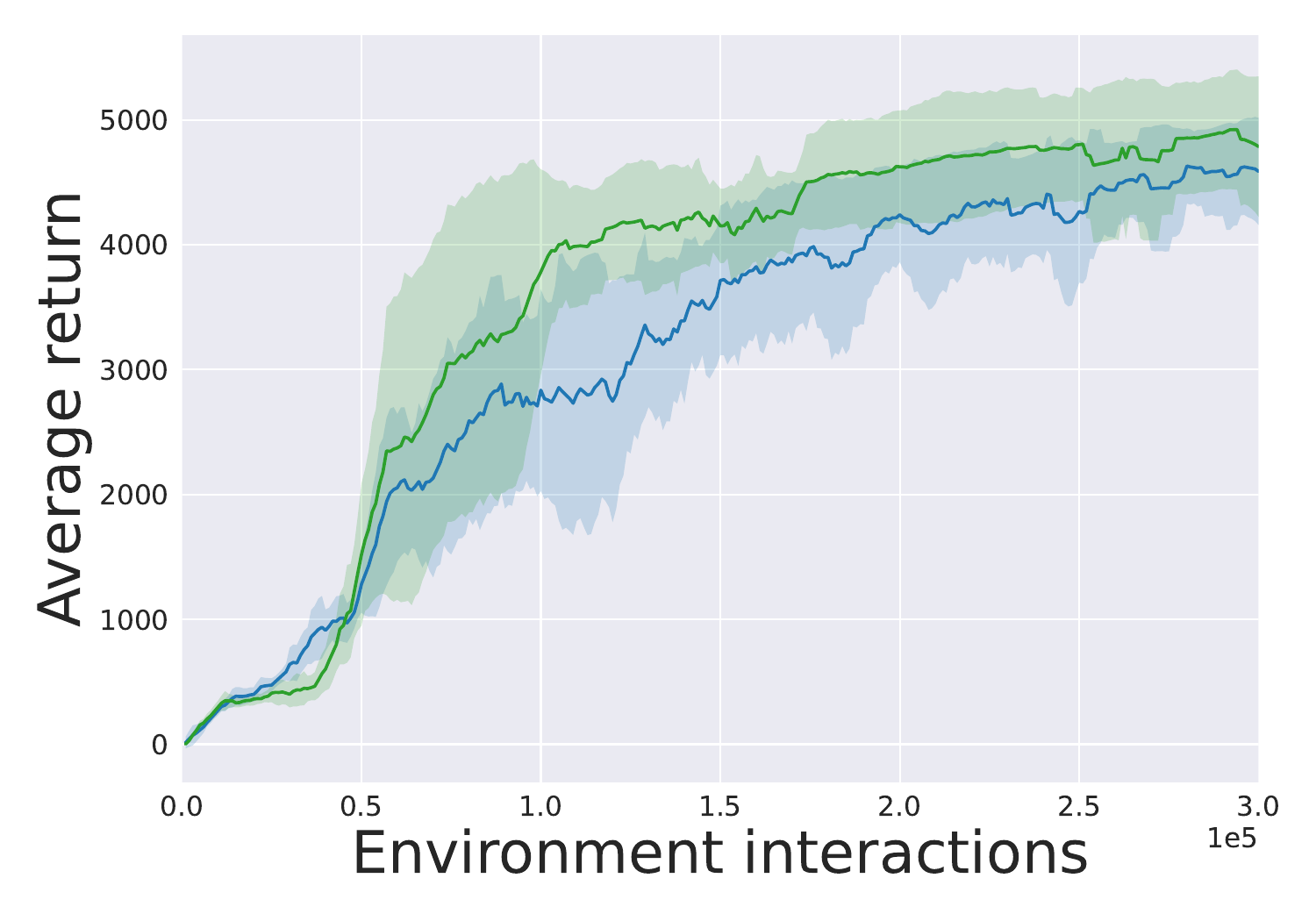}
    \subcaption{Walker2d-v2}
  \end{minipage}
  \begin{minipage}[b]{0.32\columnwidth}
    \centering
    \includegraphics[width=\columnwidth]{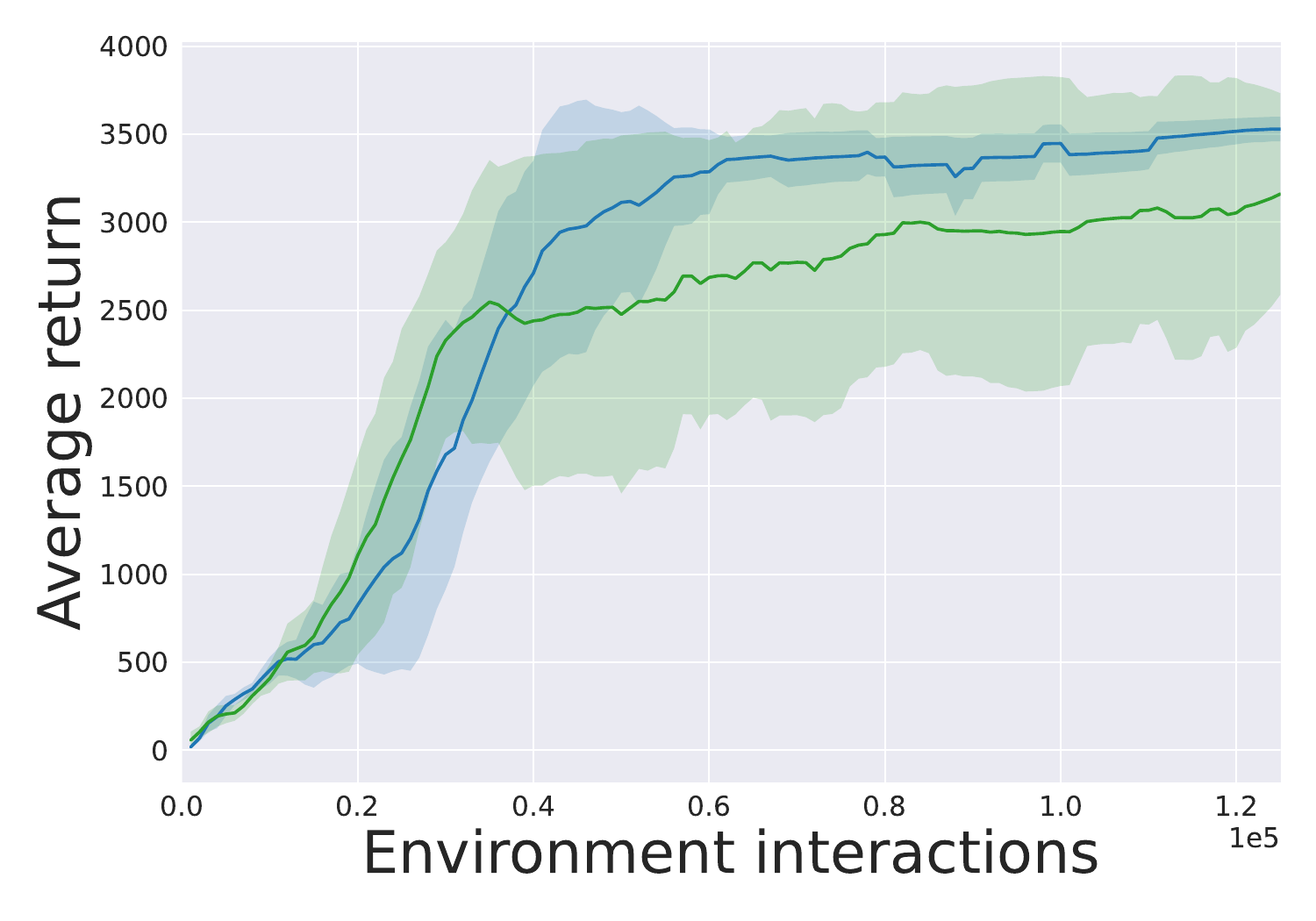}
    \subcaption{Hopper-v2}
  \end{minipage}
  \begin{minipage}[b]{0.32\columnwidth}
    \centering
    \includegraphics[width=\columnwidth]{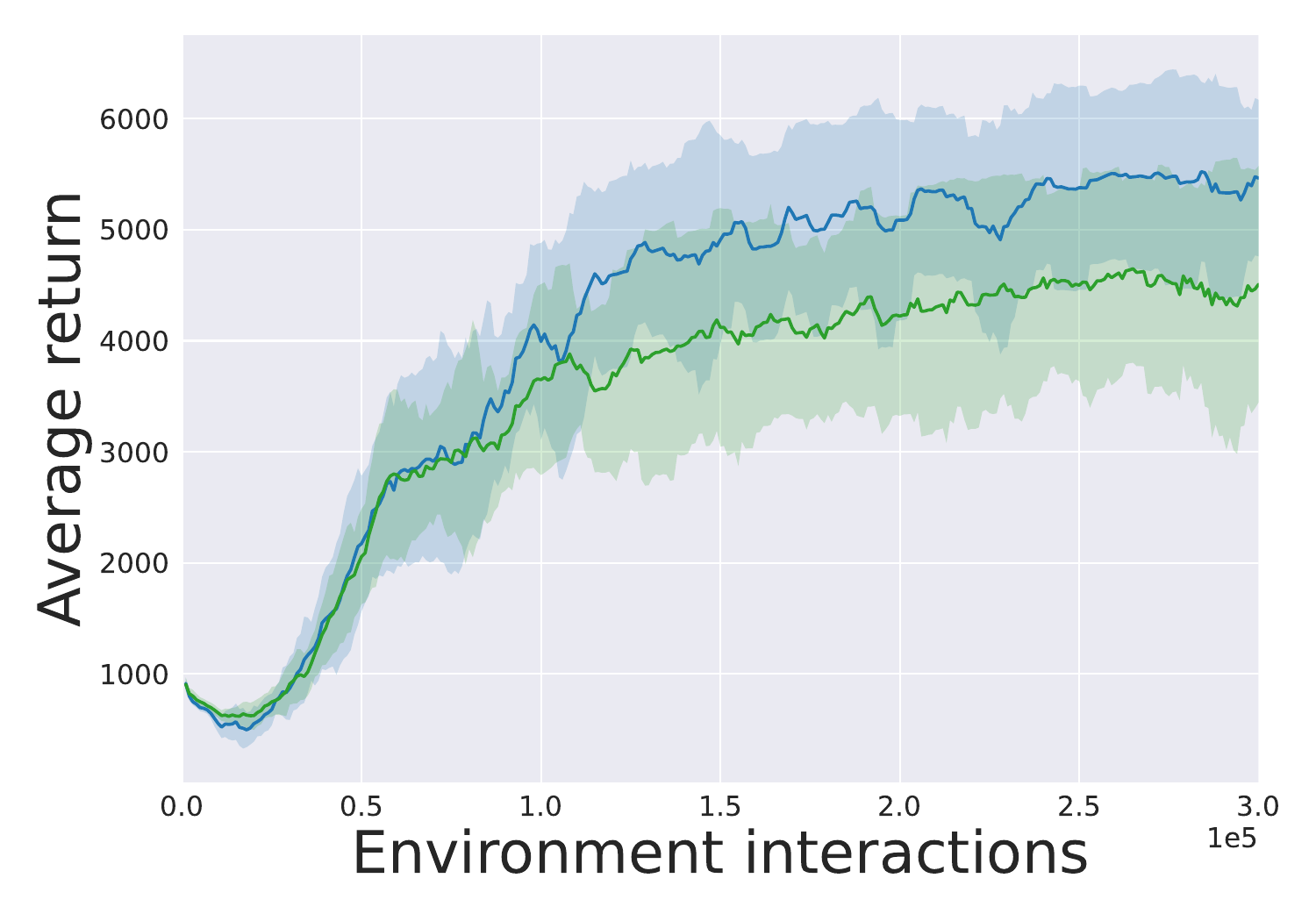}
    \subcaption{Ant-v2}
  \end{minipage}
  \begin{minipage}[b]{0.32\columnwidth}
    \centering
    \includegraphics[width=\columnwidth]{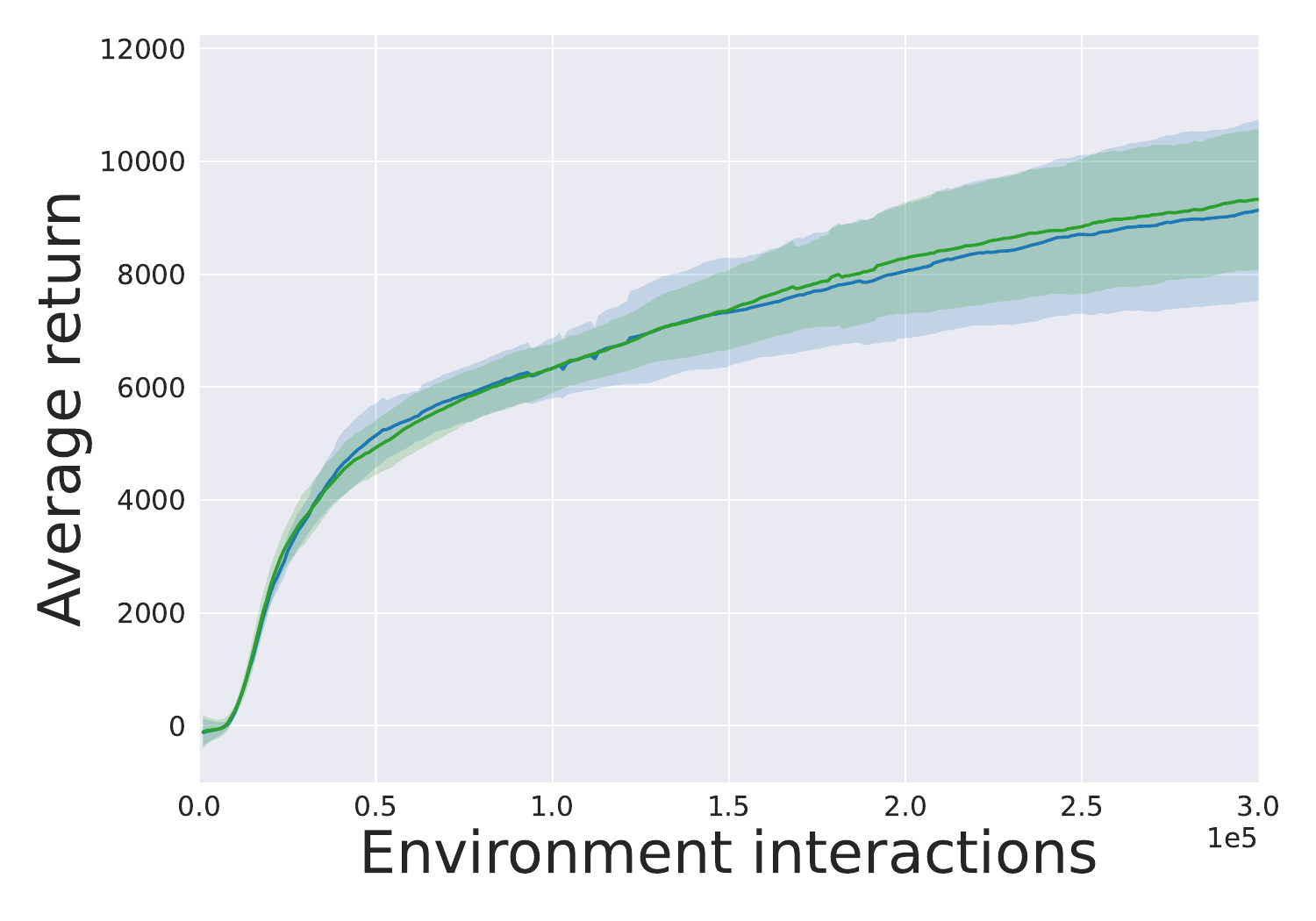}
    \subcaption{HalfCheetah-v2}
  \end{minipage}
  \hspace{1cm}
  \begin{minipage}[b]{0.20\columnwidth}
    \centering
    \includegraphics[width=\columnwidth]{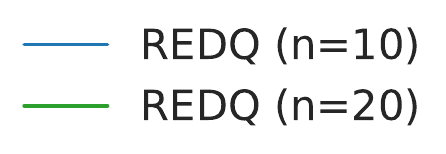}
    \vspace{0.3cm}
    \subcaption*{}
  \end{minipage} 
  \caption{Performance changes based on the number of critics $n$ in REDQ}
  \label{fig:n_redq}
\end{figure}

\subsection{Temperature $\beta$ for Gumbel regression}

\begin{figure}[]
  \begin{minipage}[b]{0.32\columnwidth}
    \centering
    \includegraphics[width=\columnwidth]{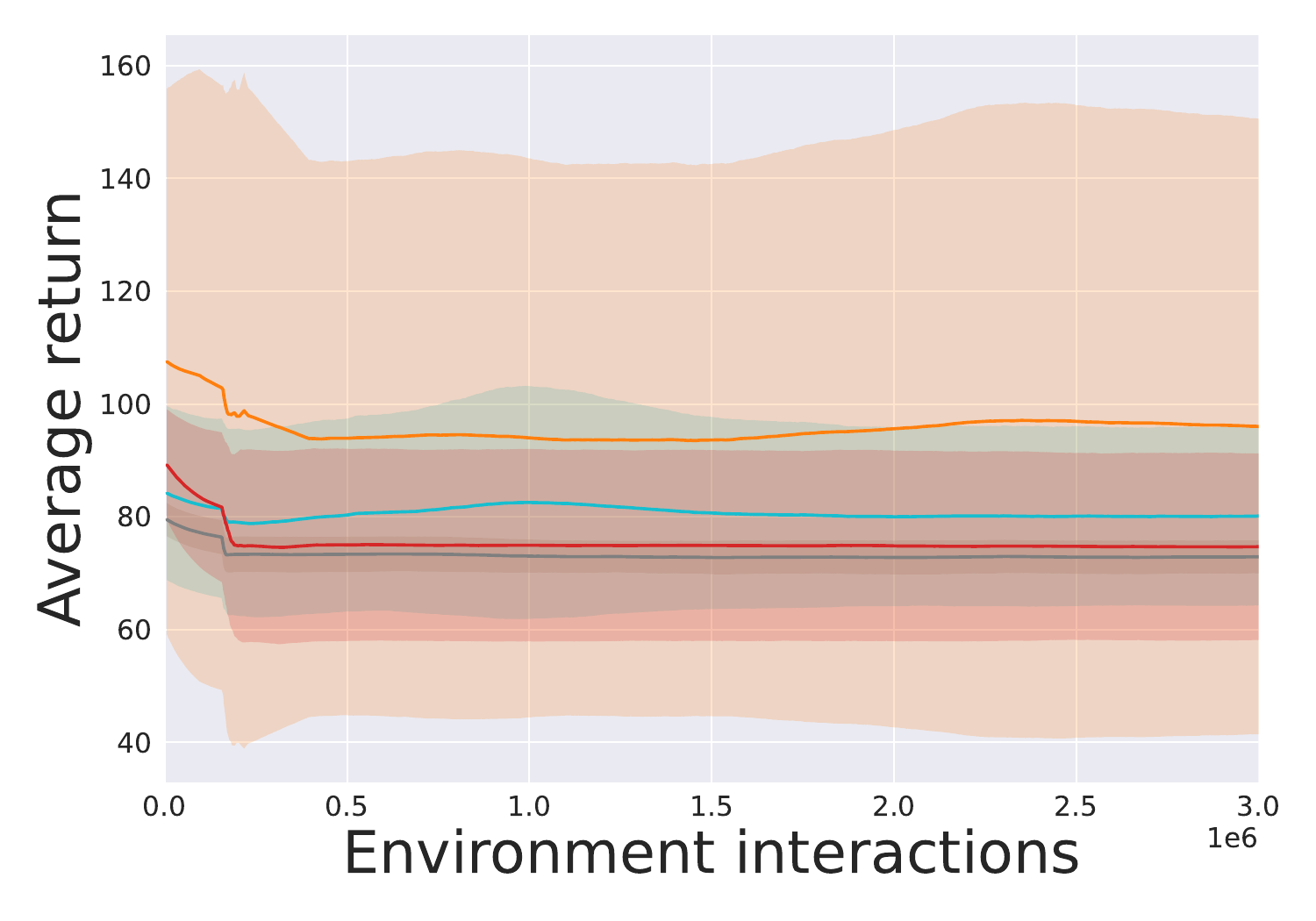}
    \subcaption{Humanoid-v2}
  \end{minipage}
 \begin{minipage}[b]{0.32\columnwidth}
    \centering
    \includegraphics[width=\columnwidth]{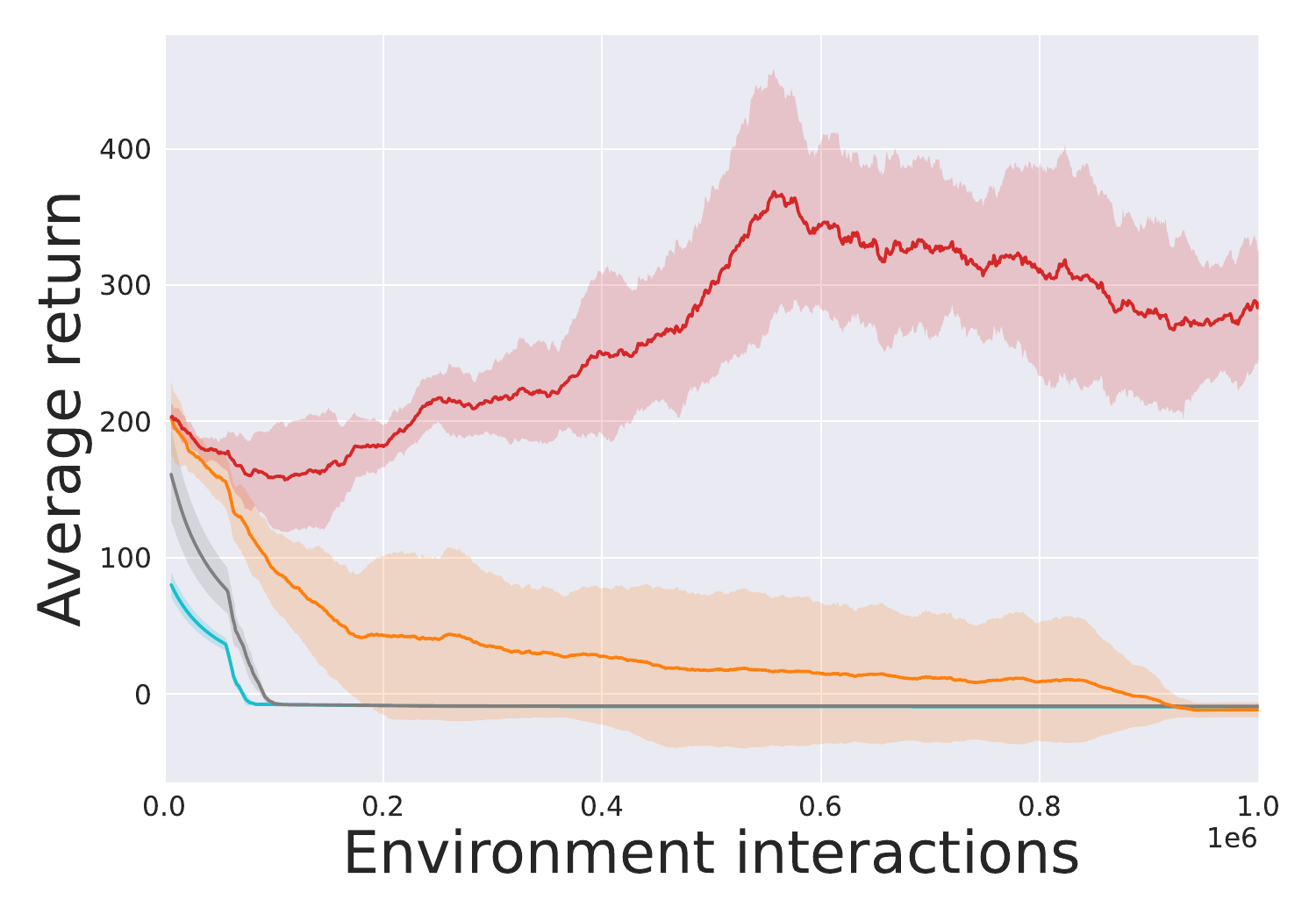}
    \subcaption{Walker2d-v2}
  \end{minipage}
  \begin{minipage}[b]{0.32\columnwidth}
    \centering
    \includegraphics[width=\columnwidth]{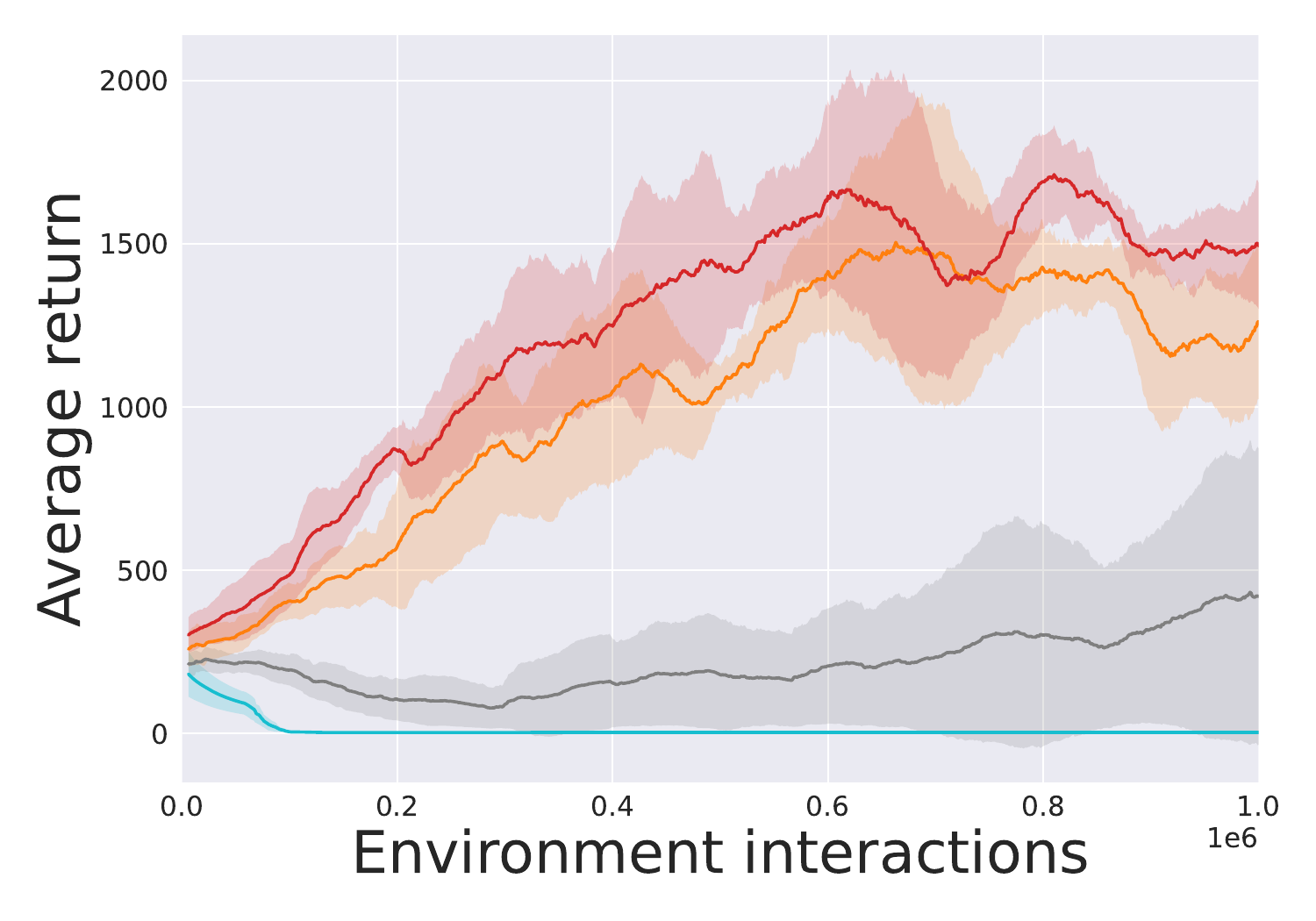}
    \subcaption{Hopper-v2}
  \end{minipage}
  \begin{minipage}[b]{0.32\columnwidth}
    \centering
    \includegraphics[width=\columnwidth]{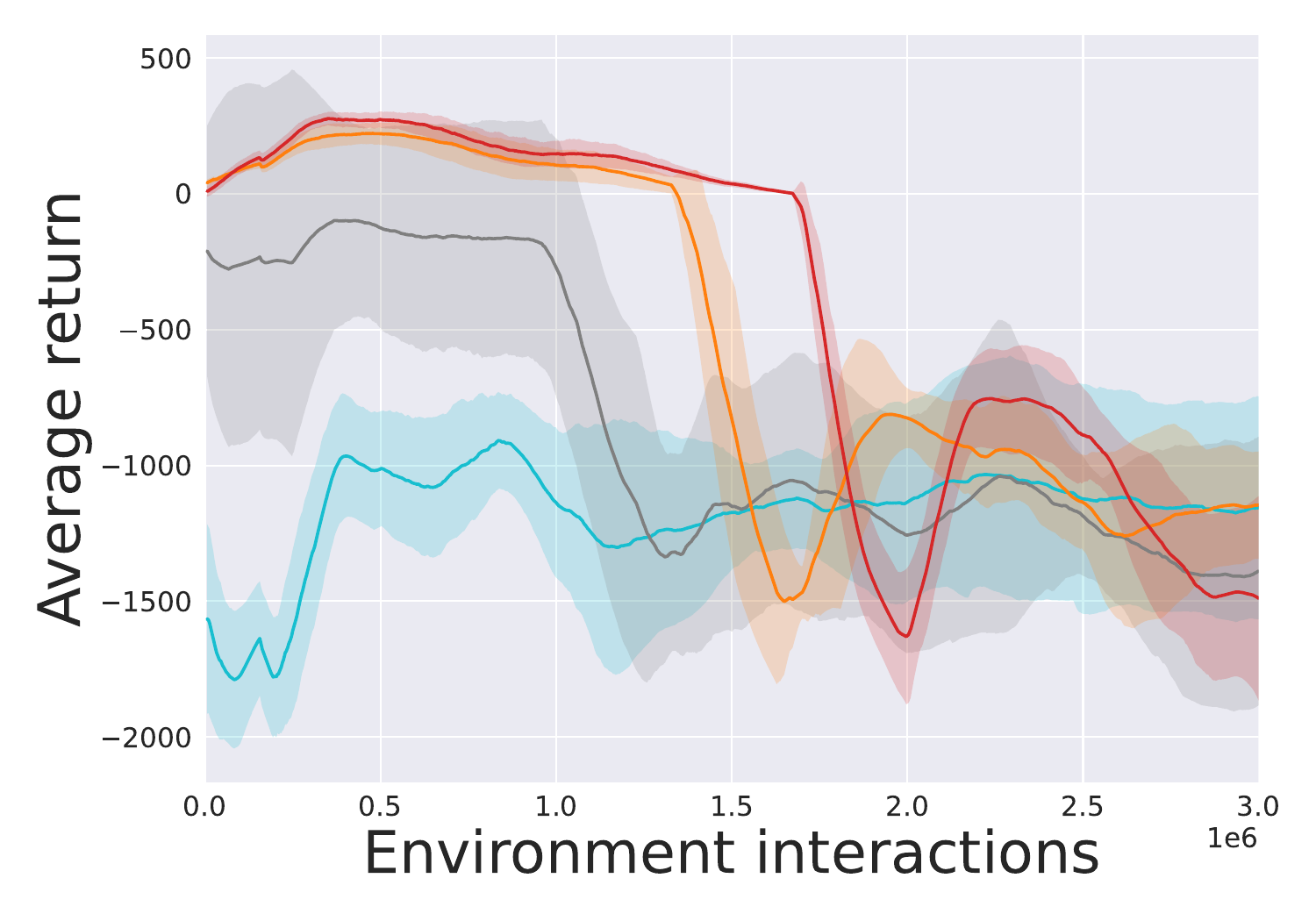}
    \subcaption{Ant-v2}
  \end{minipage}
  \begin{minipage}[b]{0.32\columnwidth}
    \centering
    \includegraphics[width=\columnwidth]{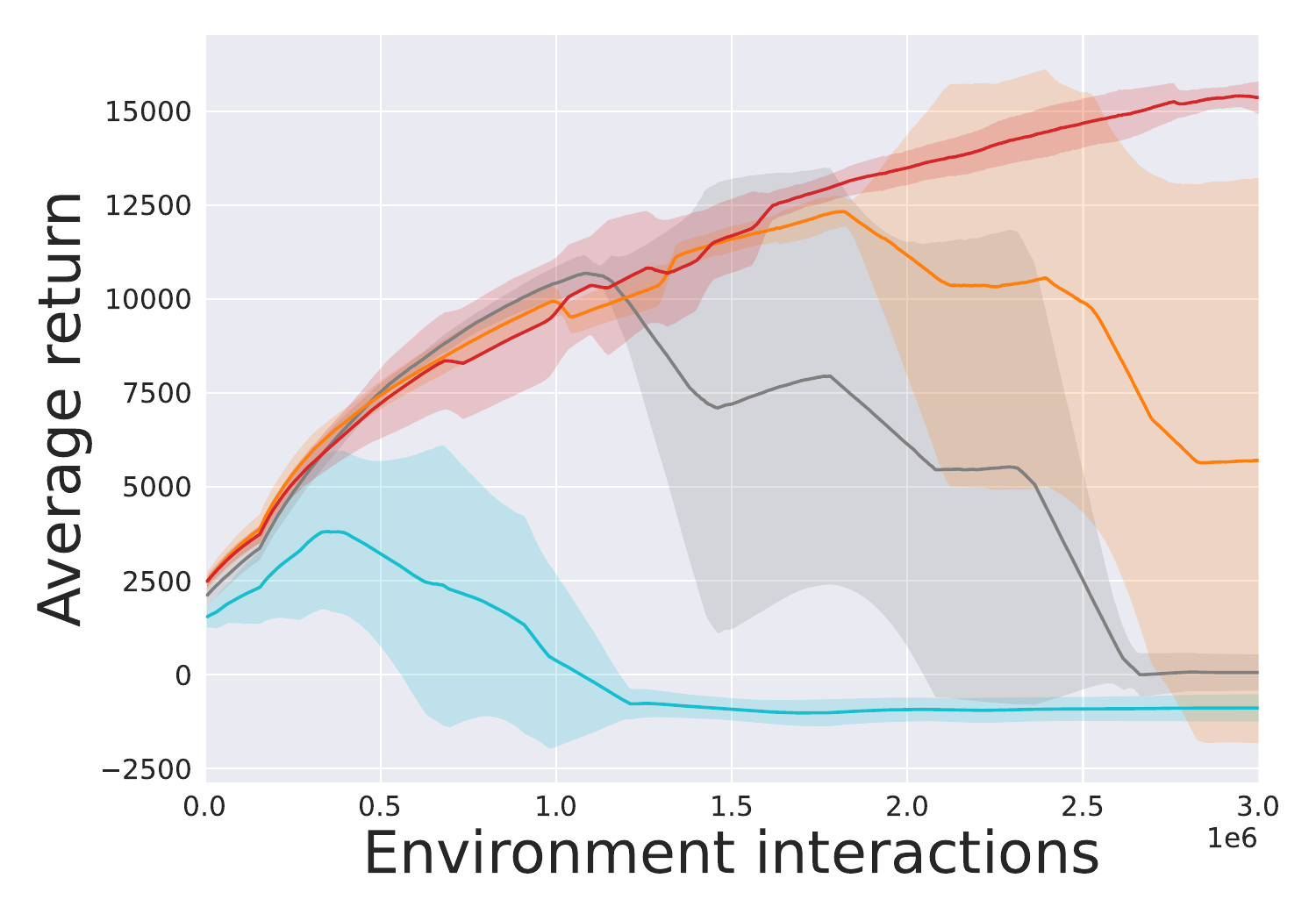}
    \subcaption{HalfCheetah-v2}
  \end{minipage}
  \hspace{1cm}
  \begin{minipage}[b]{0.20\columnwidth}
    \centering
    \includegraphics[width=\columnwidth]{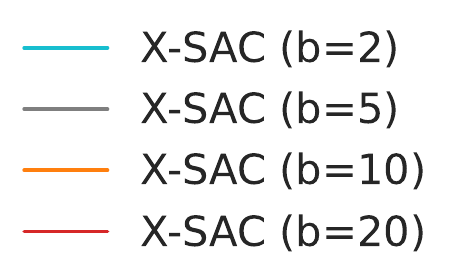}
    \vspace{0.3cm}
    \subcaption*{}
  \end{minipage} 
  \caption{Performance changes based on the temperature $\beta$ in $\mathcal{X}$-SAC}
  \label{fig:b_xsac}
\end{figure}

\begin{figure}[]
  \begin{minipage}[b]{0.32\columnwidth}
    \centering
    \includegraphics[width=\columnwidth]{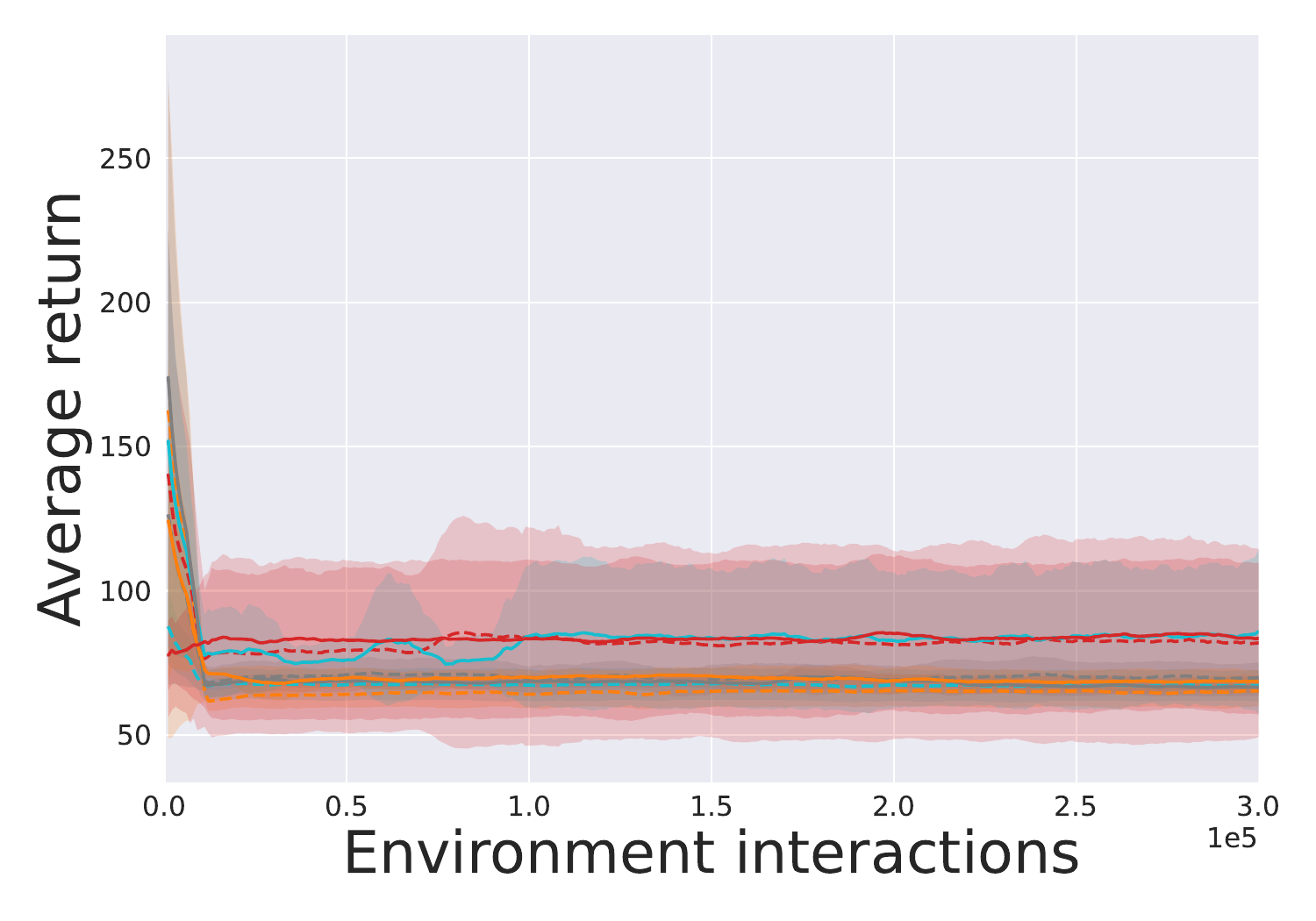}
    \subcaption{Humanoid-v2}
  \end{minipage}
 \begin{minipage}[b]{0.32\columnwidth}
    \centering
    \includegraphics[width=\columnwidth]{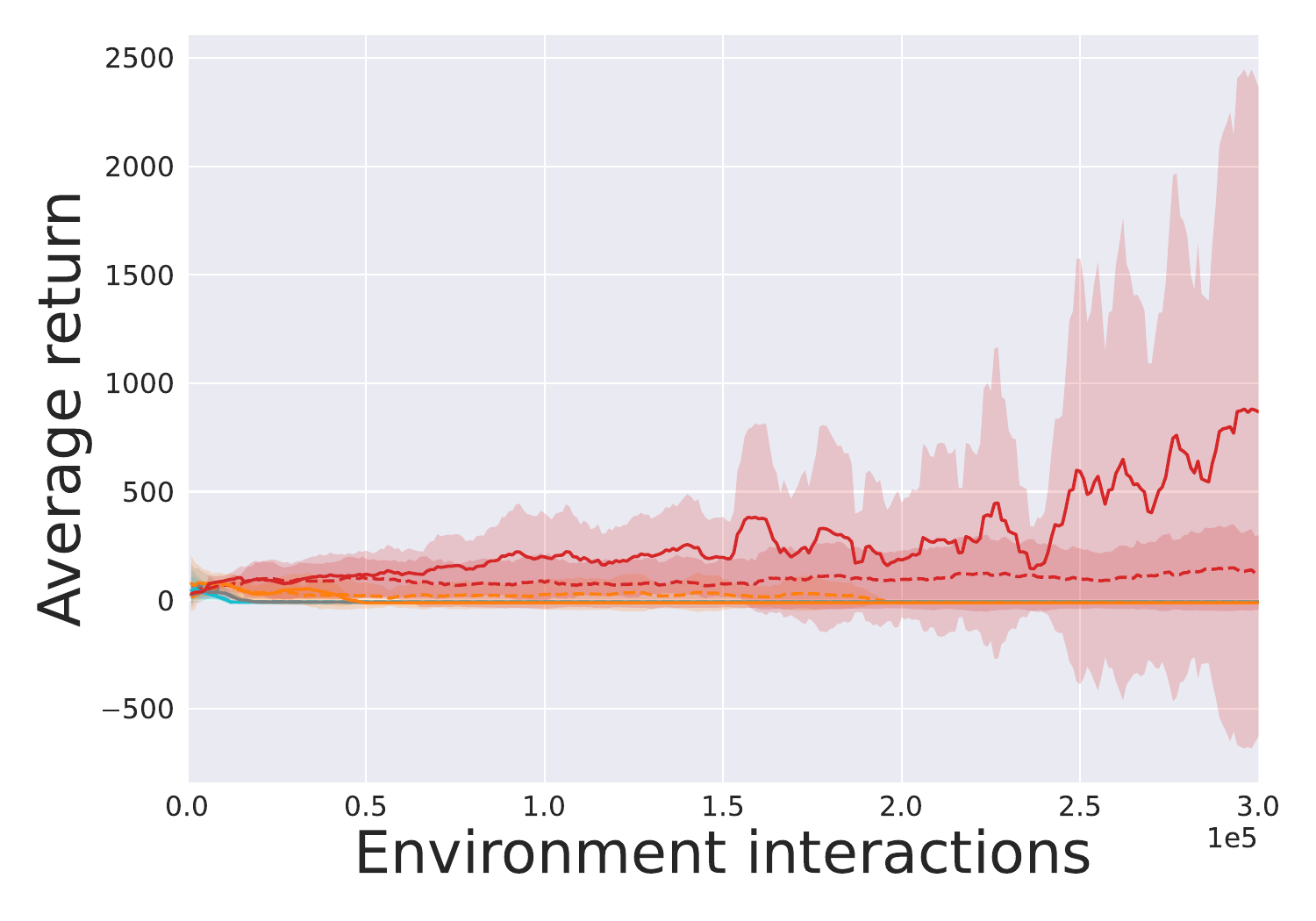}
    \subcaption{Walker2d-v2}
  \end{minipage}
  \begin{minipage}[b]{0.32\columnwidth}
    \centering
    \includegraphics[width=\columnwidth]{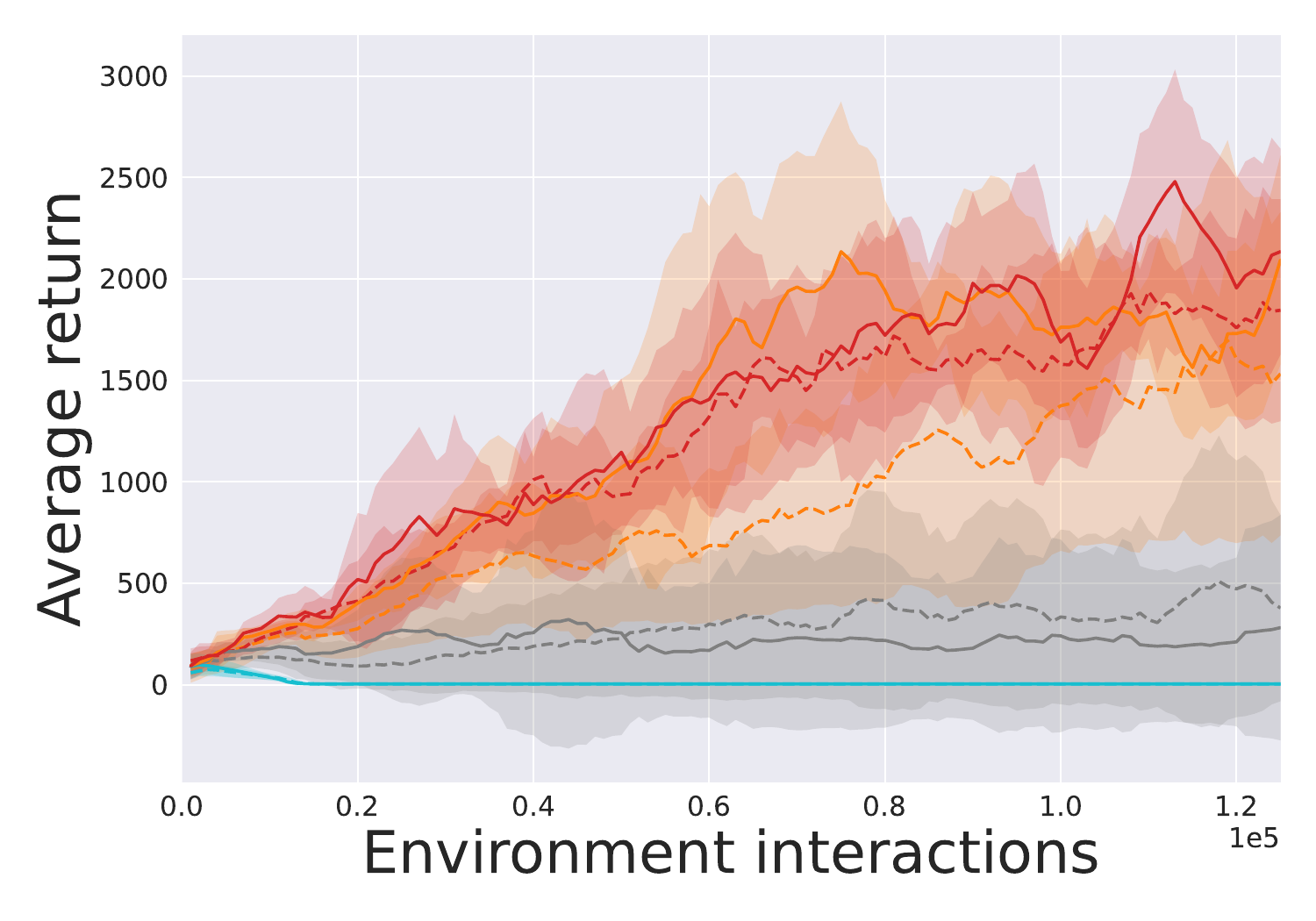}
    \subcaption{Hopper-v2}
  \end{minipage}
  \begin{minipage}[b]{0.32\columnwidth}
    \centering
    \includegraphics[width=\columnwidth]{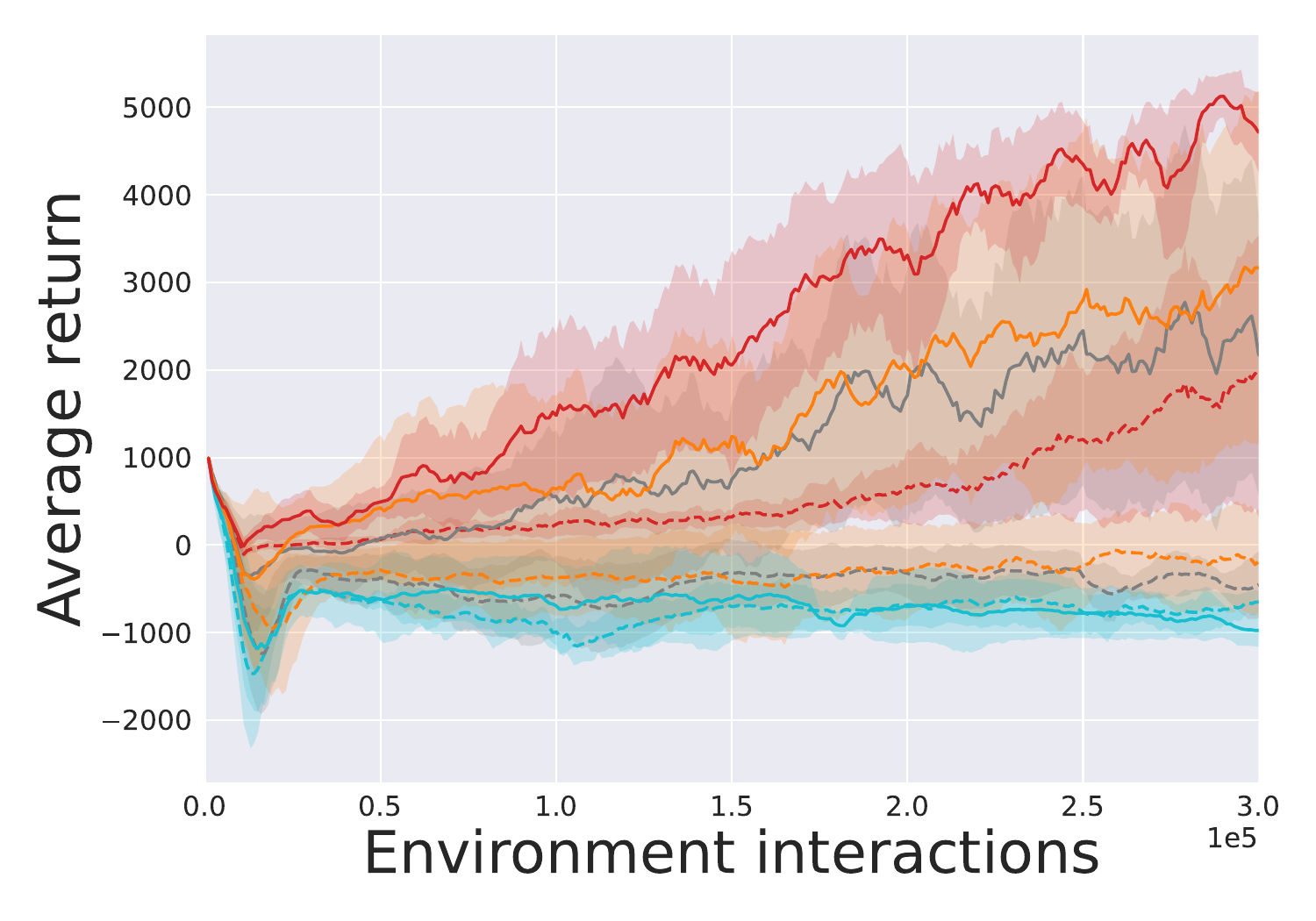}
    \subcaption{Ant-v2}
  \end{minipage}
  \begin{minipage}[b]{0.32\columnwidth}
    \centering
    \includegraphics[width=\columnwidth]{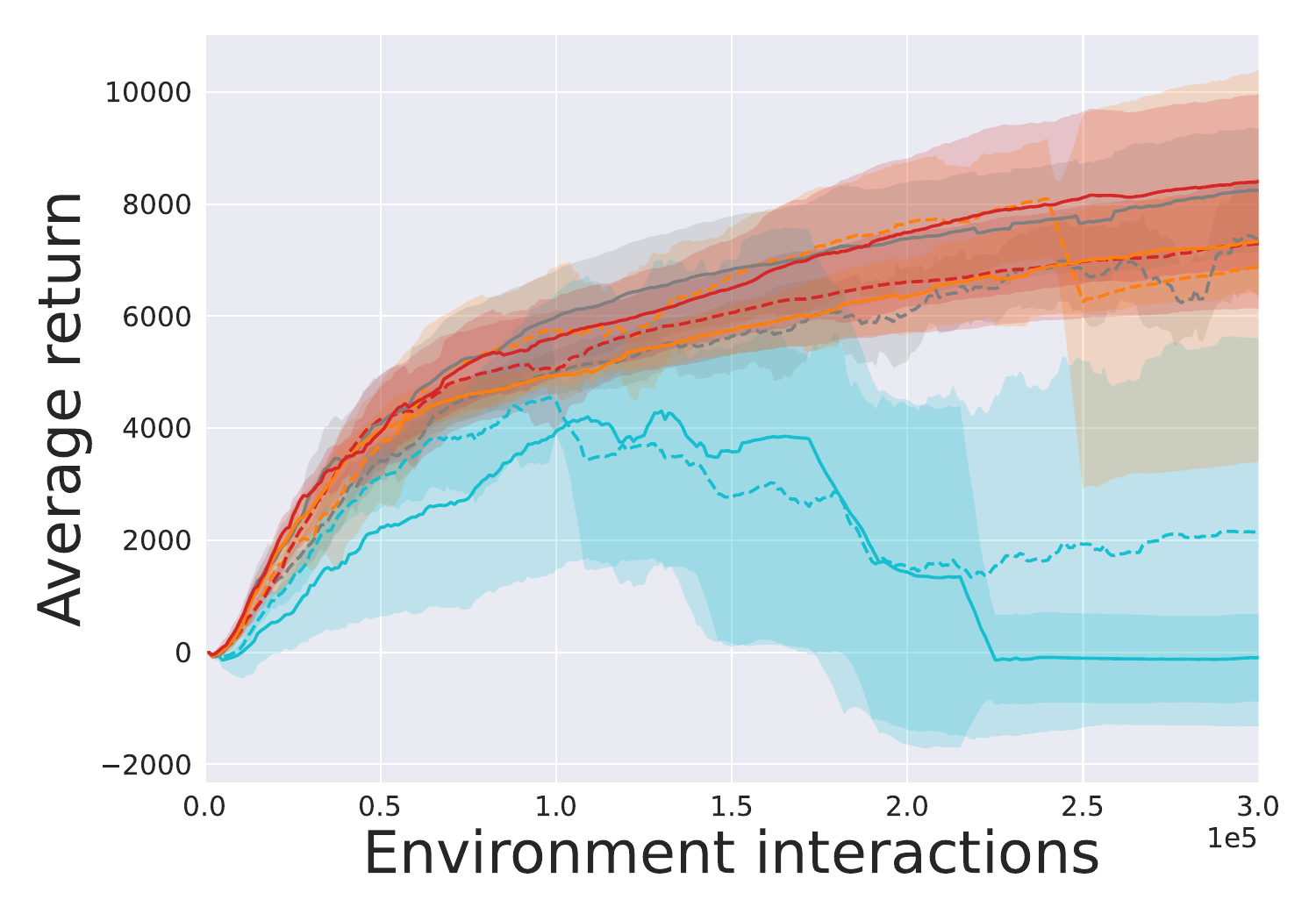}
    \subcaption{HalfCheetah-v2}
  \end{minipage}
  \hspace{1cm}
  \begin{minipage}[b]{0.20\columnwidth}
    \centering
    \includegraphics[width=\columnwidth]{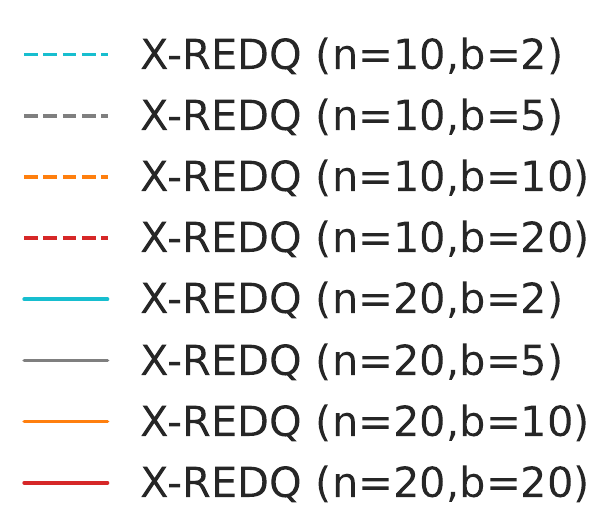}
    \vspace{0.3cm}
    \subcaption*{}
  \end{minipage} 
  \caption{Performance changes based on the temperature $\beta$ and the number of critics $n$ in $\mathcal{X}$-REDQ}
  \label{fig:b_xredq}
\end{figure}

In extreme Q-learning, the temperature $\beta$ adjusts regularization in the max entropy framework. In this study, we adopted the best performing value from [2, 5, 10, 20], which is almost the same as the value verified by \cite{garg2023extreme}. Here, we verify the change in performance due to the difference in $\beta$. The results of $\mathcal{X}$-SAC and $\mathcal{X}$-REDQ are \cref{fig:b_xsac} and \cref{fig:b_xredq}, respectively. In both cases, stable learning was not possible when $\beta$ was small. In the case of $\beta = 20$, there were cases where stable learning was possible, and it was found that a larger $\beta$ needs to be set for difficult tasks like these five tasks. However, compared to SAC and REDQ, the performance is significantly inferior. Therefore, changes other than $\beta$ tuning are necessary.

\begin{algorithm}[]
   \caption{Variational Inference for Gaussian Mixture Models} 
   \label{alg:gmm}
    \begin{algorithmic}
    \STATE \textbf{INITIALIZE}
    \begin{ALC@g}
    \STATE Initialize $\alpha, \beta, \nu, \mathbf{m},  \mathbf W$ for each cluster $k$ for 
    \begin{equation*}
    \begin{split}
    & p(\mathbf{x}) = \Sigma^{K}_{k=1}\pi_k \mathcal{N}(\mathbf{x} \mid \boldsymbol\mu_k, \boldsymbol\Lambda^{-1}_{k}) \\
     & p(\pi_k) = Dir(\pi_k \mid \alpha_k) \\
     & q(\boldsymbol\mu_k, \boldsymbol\Lambda_k) = \mathcal{N}(\boldsymbol\mu_k \mid \mathbf{m}_k, (\beta_k \boldsymbol\Lambda_k)^{-1})\mathcal{W}(\boldsymbol\Lambda \mid \mathbf{W}_k, \nu_k)
    \end{split}
    \end{equation*}
    \end{ALC@g}

    \STATE \textbf{UPDATE}
    \begin{ALC@g}

    \STATE \textbf{Input:} data $\{\mathbf{x}_1,..., \mathbf{x}_N\}$, paramters $(\alpha_k, \beta_k, \nu_k, \mathbf{m}_k,  \mathbf {W}_k)$ for each cluster k
    \STATE \textbf{Output:} paramters $(\alpha_{k}', \beta_{k}', \nu_{k}', \mathbf{m}_{k}',  \mathbf {W}_{k}')$ for each cluster k
    \STATE Compute the resposibilities $r_{nk}$ for each data $\mathbf{x}_n$ and cluster $k$:
    \begin{align*}
    & r_{nk} = \frac{\rho_{nk}}{\Sigma^{K}_{j=1}\rho_{nj}} \\
    & where \notag \\
    & \rho_{nk} = \tilde{\pi}_{k} \tilde{\boldsymbol\Lambda}^{1/2}_{k} \exp\{{-\frac{D}{2\beta_k} - \frac{\nu_k}{2}(\mathbf{x}_n - \mathbf{m}_k)^T \mathbf{W}_k (\mathbf{x}_n - \mathbf{m}_k)}\}  \\
    & \ln\tilde{\pi}_k = \psi(\alpha_k) - \psi(\Sigma^{K}_{k=1}\alpha_k) \\
    & \ln \tilde{\boldsymbol{\Lambda}}_k = \Sigma^{D}_{i=1} \psi(\frac{\nu_k +1 -i}{2}) + D \ln 2 + \ln | \mathbf{W}_k| 
    \end{align*}

    \STATE Update parameters for each cluster $k$: 
    \begin{align*}
    & \alpha_{k}' = \alpha_k + N_k \\
    & \beta_{k}' = \beta_k + N_k  \\
    & \nu_{k}' = \nu_k + N_k \\
    &  \mathbf{m}_{k}' = \frac{1}{\beta_{k}'}(\beta_k  \mathbf{m}_k + N_k \bar{\mathbf{x}}_k) \\
    &  (\mathbf{W}_{k}')^{-1} =  \mathbf{W}^{-1}_{k} + N_k \mathbf{S}_k + \frac{\beta_k N_k}{\beta_k + N_k} (\bar{\mathbf{x}}_k - \mathbf{m}_k)(\bar{\mathbf{x}}_k -  \mathbf{m}_k)^T \\
    & where \notag \\
    & N_k = \Sigma^{N}_{n=1} r_{nk} \\
    & \bar{\mathbf{x}}_k = \frac{1}{N_k} \Sigma^{N}_{n=1} r_{nk} x_n \\
    & \mathbf{S}_k = \frac{1}{N_k} \Sigma^{N}_{n=1} r_{nk} (x_n - \bar{\mathbf{x}}_k)(x_n-\bar{\mathbf{x}}_k)^T 
    \end{align*}

    \STATE Adjust $\mathbf{m}_k$ so that the overall average is zero:
    $$ \mathbf{m}_k = \mathbf{m}_k - \Sigma^K_{j=1} \pi_j  \mathbf{m}_j $$

    \end{ALC@g}

    \end{algorithmic}
\end{algorithm}

\section{Pseudocode of GMM estimation}

In the practical algorithm of our proposed method, the distribution of the noise used to correct the Bellman error is defined by a GMM, and its parameters are estimated by variational inference. Before the start of the learning phase, the parameters of the GMM are initialized as shown in the INITIALIZE function of \cref{alg:gmm} and then updated in an online manner. The operation performed in each update is shown in the UPDATE function of \cref{alg:gmm}, where the input is the Bellman error before correction, denoted by $x$, and the parameters are updated. To eliminate bias caused by adding noise, the mean is adjusted to be zero at each update. Since the Bellman error is one-dimensional data, estimating the GMM has relatively low computational cost. The estimated GMM is used to sample the noise $\eta$. Details about the implementation can be found in \cite{scikit-learn}.

\section{Additional plot of Bellman error}
In the experiment section, we presented examples of correcting the Bellman error distribution for each task. 
In this section, we give further examples for different steps. \cref{fig:error} shows several distributions of the pre-corrected Bellman error, the correction noise, and the post-corrected Bellman error for each task. For each task, we provide eight examples showing the distributions for every 2,000 steps starting from 10,000 steps. While the Bellman error is symmetric in some cases before correction, in most cases, the symmetry of the distribution increases after correction. Figure 2 displays the pre-corrected error at steps 130,000, 235,000, and 265,000 in Walker2d.

\begin{figure}[H]
    \centering
    \includegraphics[width=0.65\columnwidth]{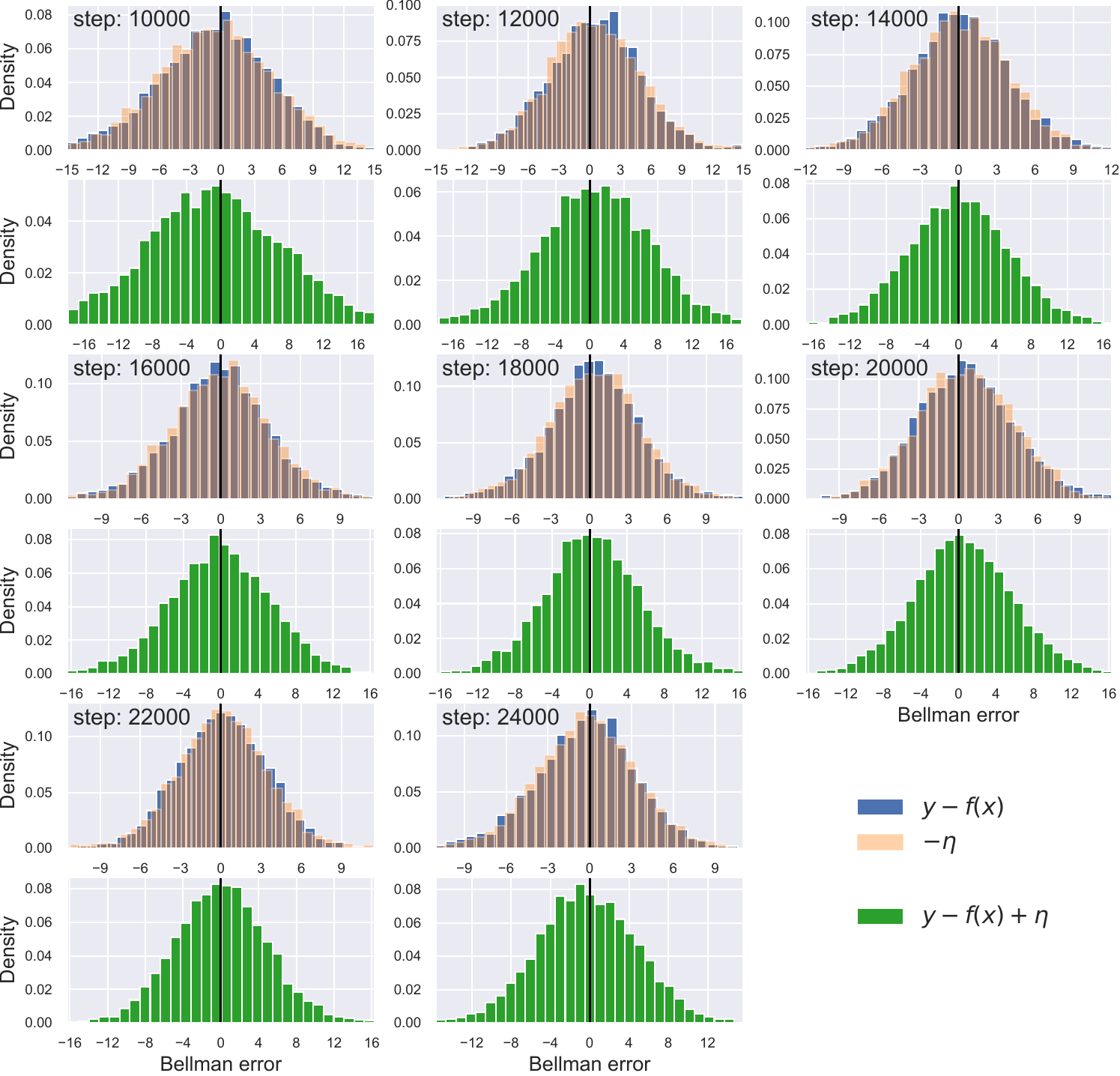}
    \caption*{(e) Humanoid-v2}
\end{figure}
\begin{figure}[H]
    \centering
    \includegraphics[width=0.65\columnwidth]{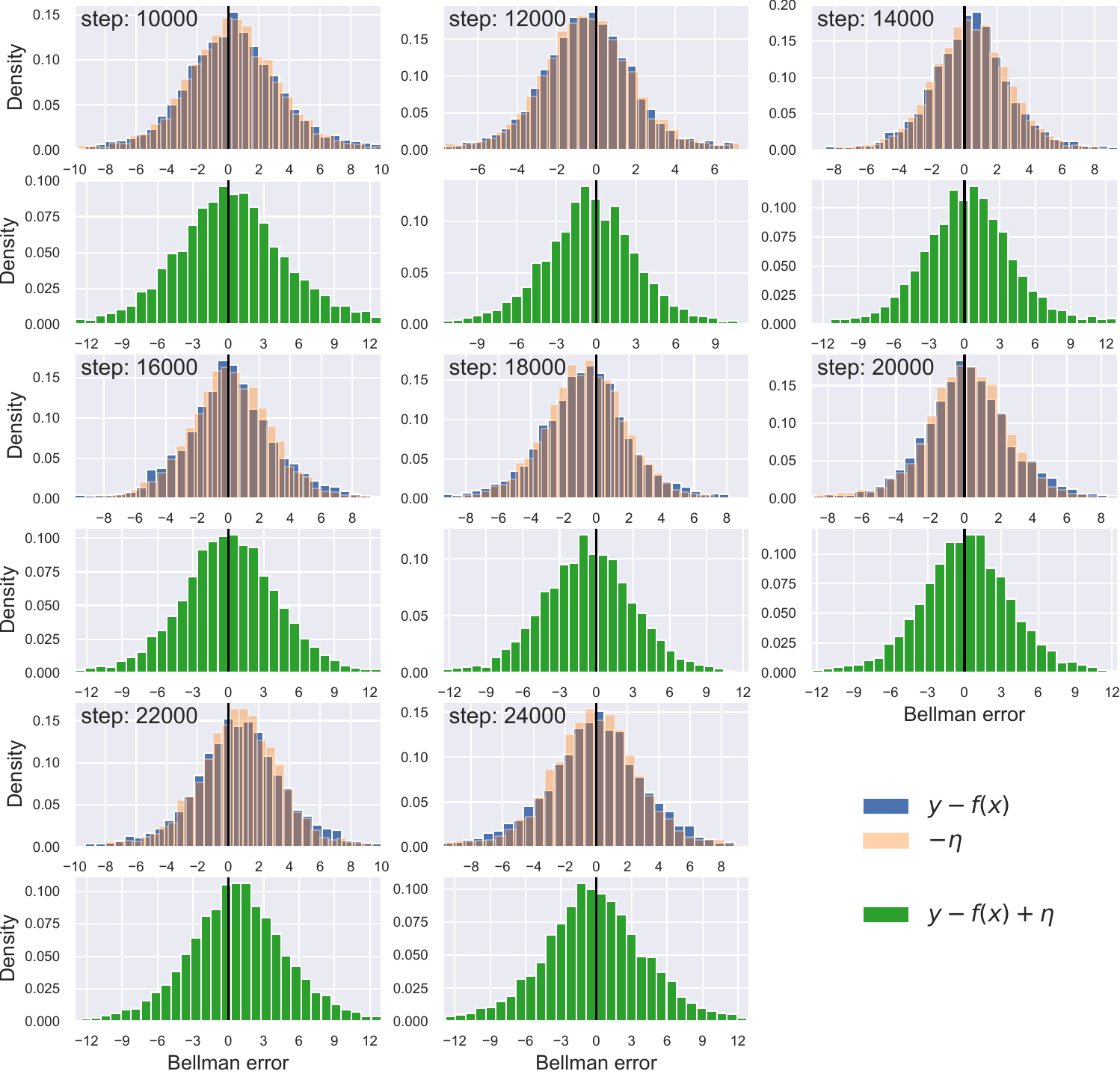}
    \caption*{(b) Walker2d-v2}
\end{figure}
\begin{figure}[H]
    \centering
    \includegraphics[width=0.65\columnwidth]{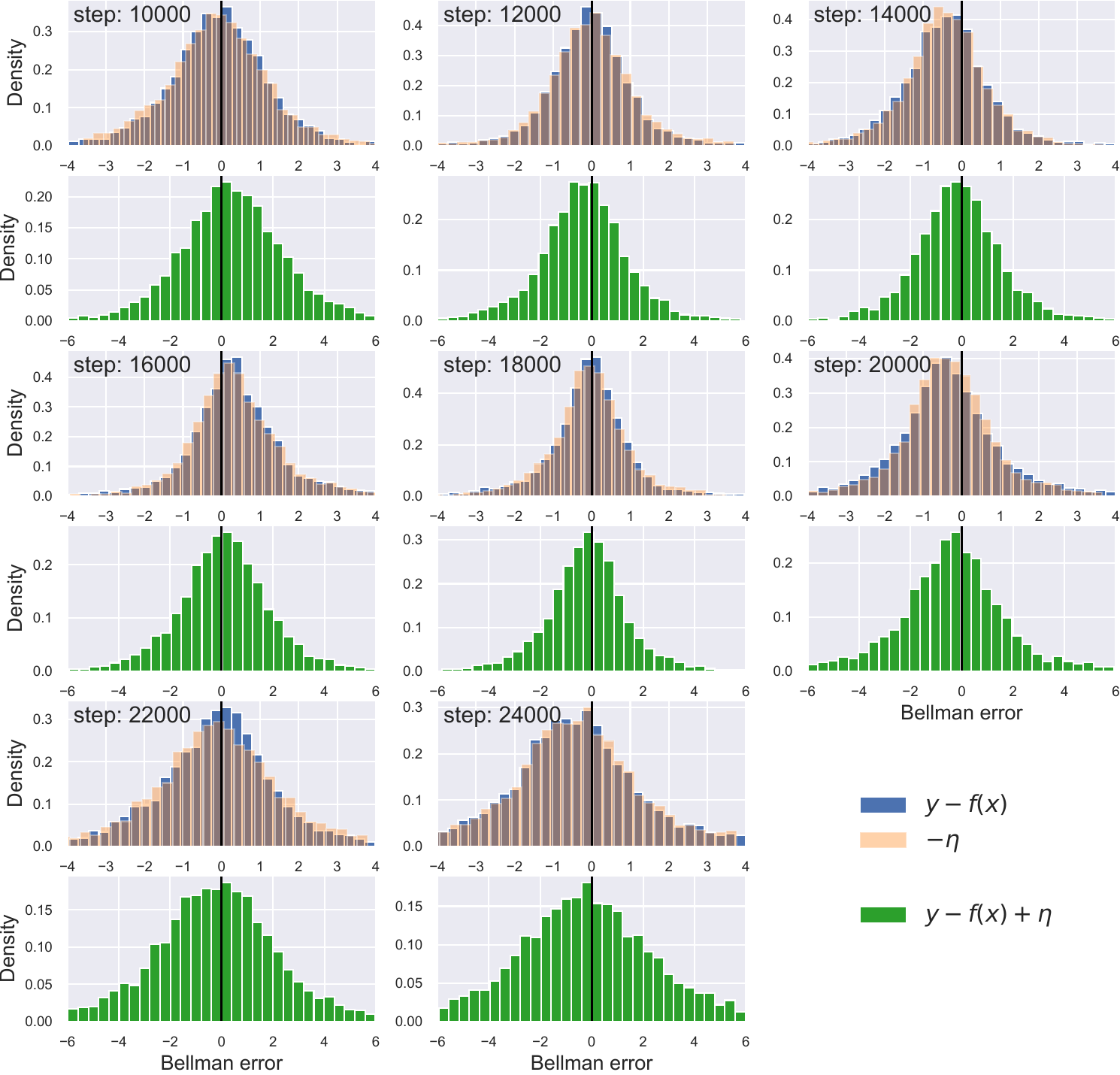}
    \caption*{(a) Hopper-v2}
\end{figure}
\begin{figure}[H]
    \centering
    \includegraphics[width=0.65\columnwidth]{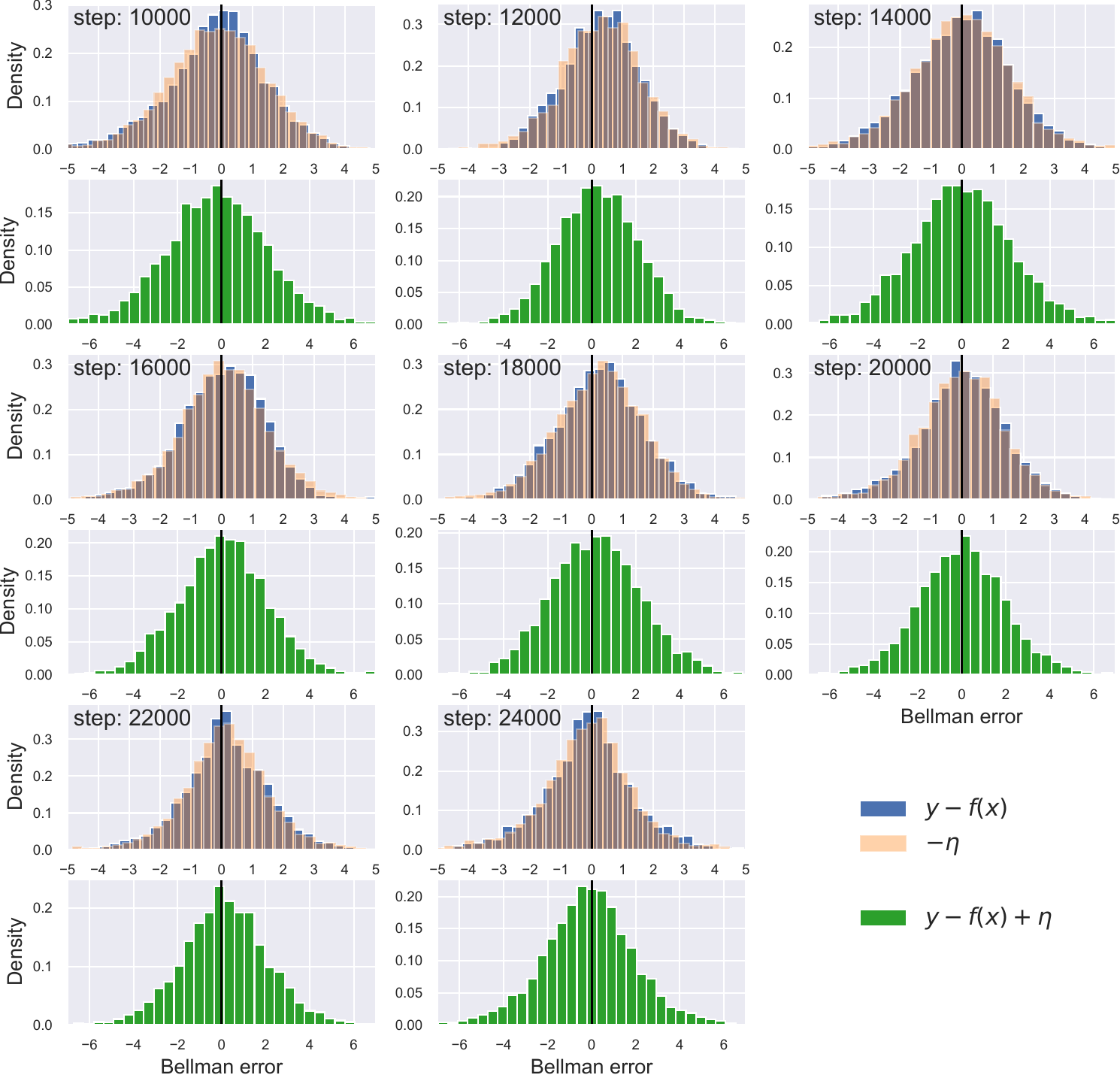}
    \caption*{(d) Ant-v2}
\end{figure}
\begin{figure}[H]
    \centering
    \includegraphics[width=0.65\columnwidth]{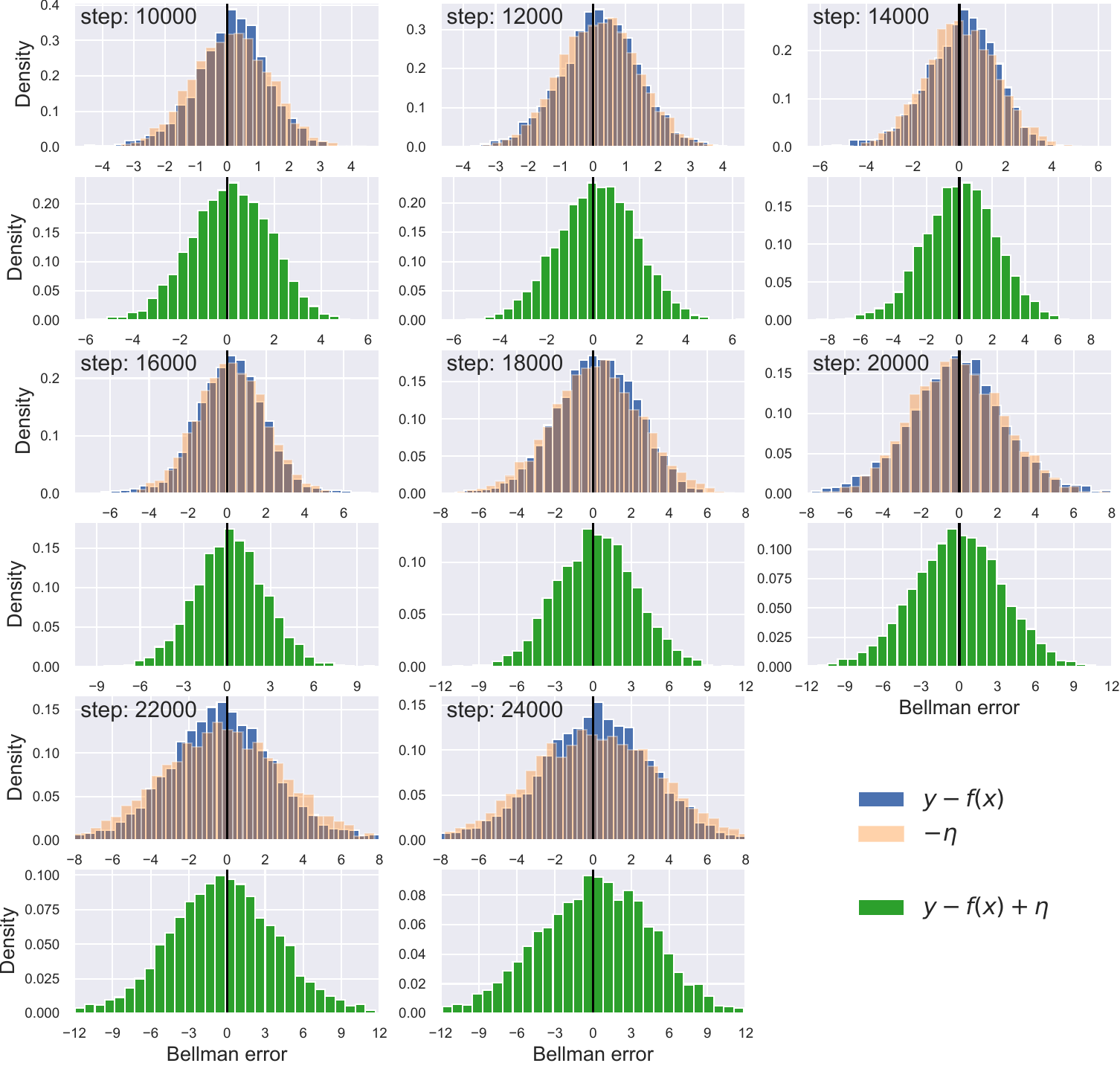}
    \caption*{(c) HalfCheetah-v2}
\end{figure}

\begin{figure}[H]
    \vspace{-8 mm}
  \caption{For each step, the top figure illustrates the density of pre-corrected Bellman error (blue) and negative values of noise (orange) used for correction. The bottom figure shows the post-corrected error (green), which is the sum of pre-corrected error and noise.}
  \label{fig:error}
\end{figure}

\end{document}